\documentclass[10pt,twocolumn,letterpaper]{article}

\usepackage[pagenumbers]{cvpr} % To force page numbers, e.g. for an arXiv version
\usepackage{graphicx}
\usepackage{amsmath}
\usepackage{amssymb}
\usepackage{booktabs}
\usepackage{bm}

\def\swsix{0.164\linewidth}
\def\swseven{0.14\linewidth}

\def\swten{0.1\linewidth}

\usepackage[pagebackref,breaklinks,colorlinks]{hyperref}

\makeatletter
\let\NAT@parse\undefined
\makeatother
\usepackage{hyperref}  %hyperref still needs to be put at the end!

\usepackage[capitalize]{cleveref}
\crefname{section}{Sec.}{Secs.}
\Crefname{section}{Section}{Sections}
\Crefname{table}{Table}{Tables}
\crefname{table}{Table}{Tabs.}

\def\cvprPaperID{3387} % *** Enter the CVPR Paper ID here
\def\confName{CVPR}
\def\confYear{2022}

\usepackage{float}

\newcommand\blfootnote[1]{%
\begingroup
\renewcommand\thefootnote{}\footnote{#1}%
\addtocounter{footnote}{-1}%
\endgroup
}

\title{RestoreFormer: High-Quality Blind Face Restoration \\ from Undegraded Key-Value Pairs}

\author{Zhouxia Wang$^1$\\
\and
Jiawei Zhang$^2$\\
\and
Runjian Chen$^1$\\
\and
Wenping Wang$^1$\\
\and
Ping Luo$^1$\thanks{This work is supported by the General Research Fund of HK No.27208720 and 17212120.}  \\
\and
\small $^1$ The University of Hong Kong, $^2$ SenseTime Research \\
\and
}

\begin{document}

\twocolumn[{%
\renewcommand\twocolumn[1][]{#1}%
\maketitle

\renewcommand{\tabcolsep}{.5pt}
\begin{figure}[H]
\hsize=\textwidth
\vspace{-0.35in}
\begin{minipage}{\textwidth}

\begin{center}
\begin{tabular}{ccccccc}
   \vspace{-1.0mm}
   \includegraphics[width=\swseven]{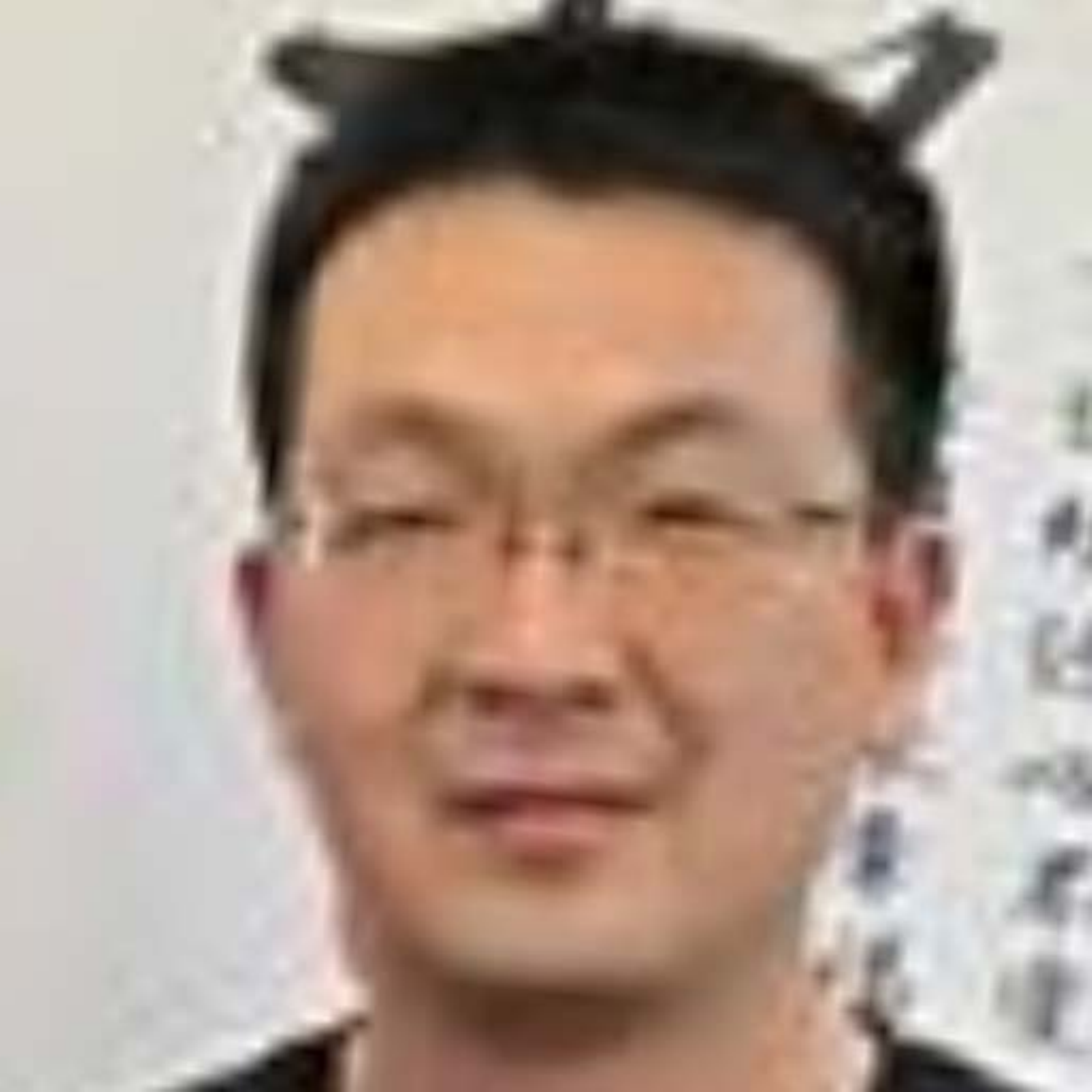}&
   \includegraphics[width=\swseven]{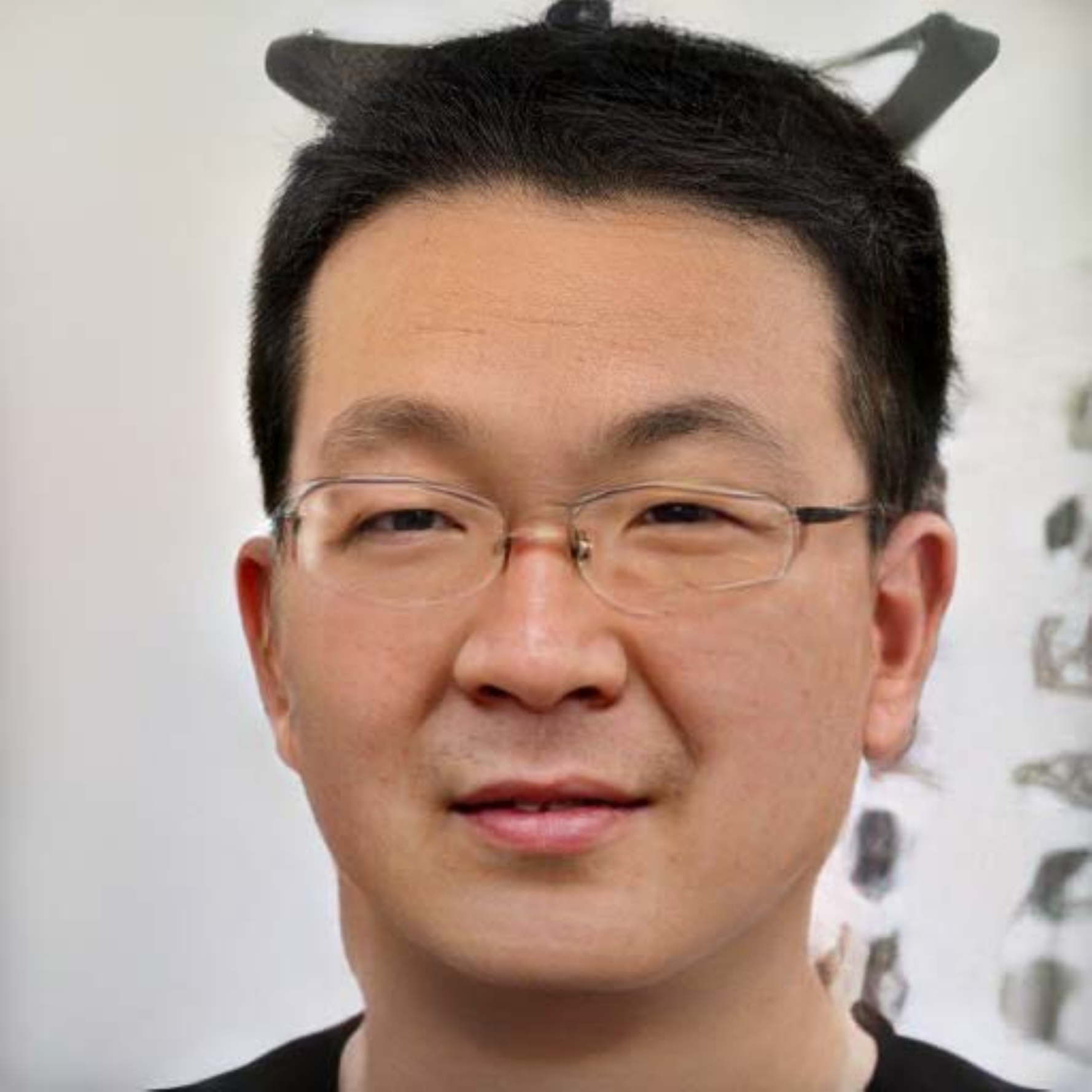}&
   \includegraphics[width=\swseven]{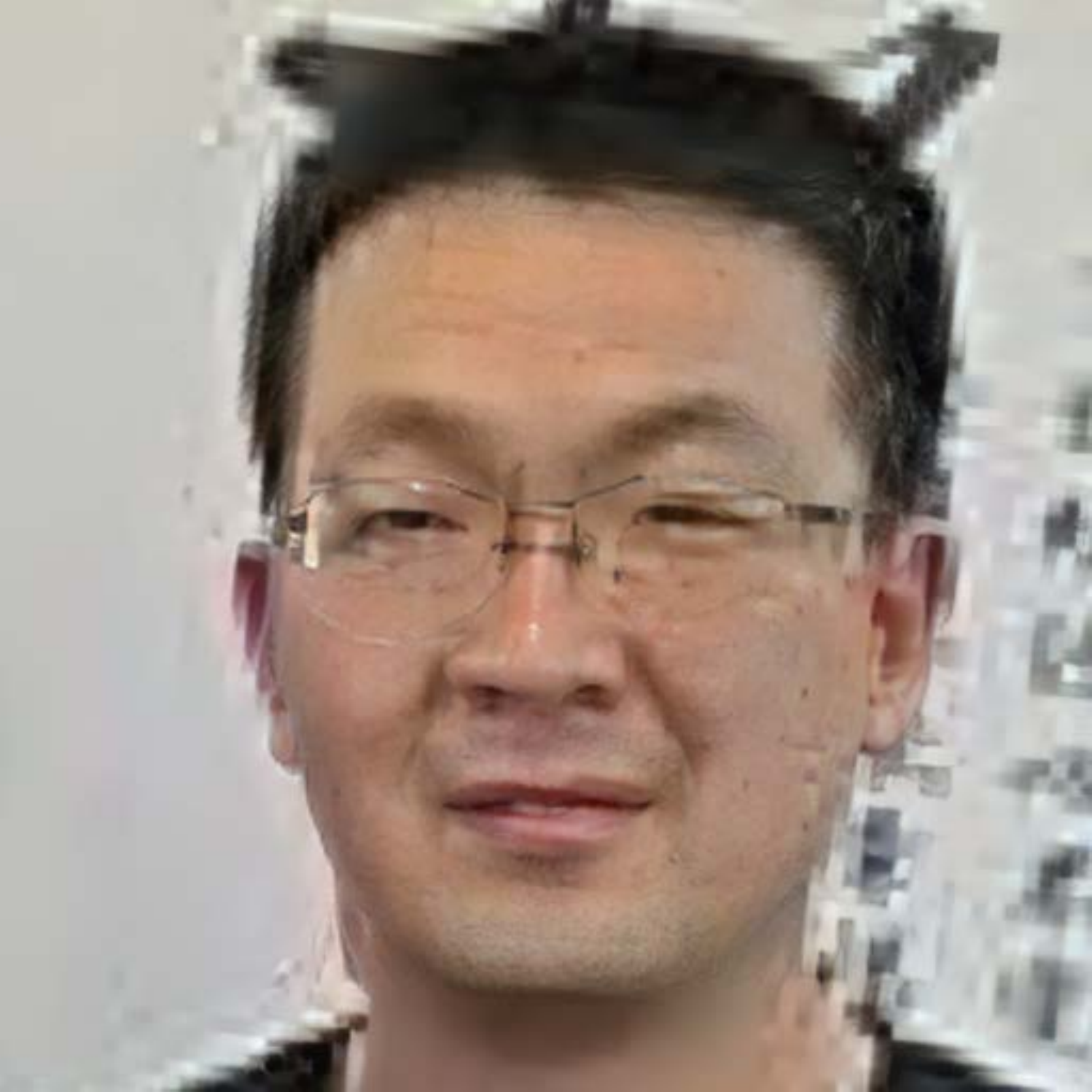}&
   \includegraphics[width=\swseven]{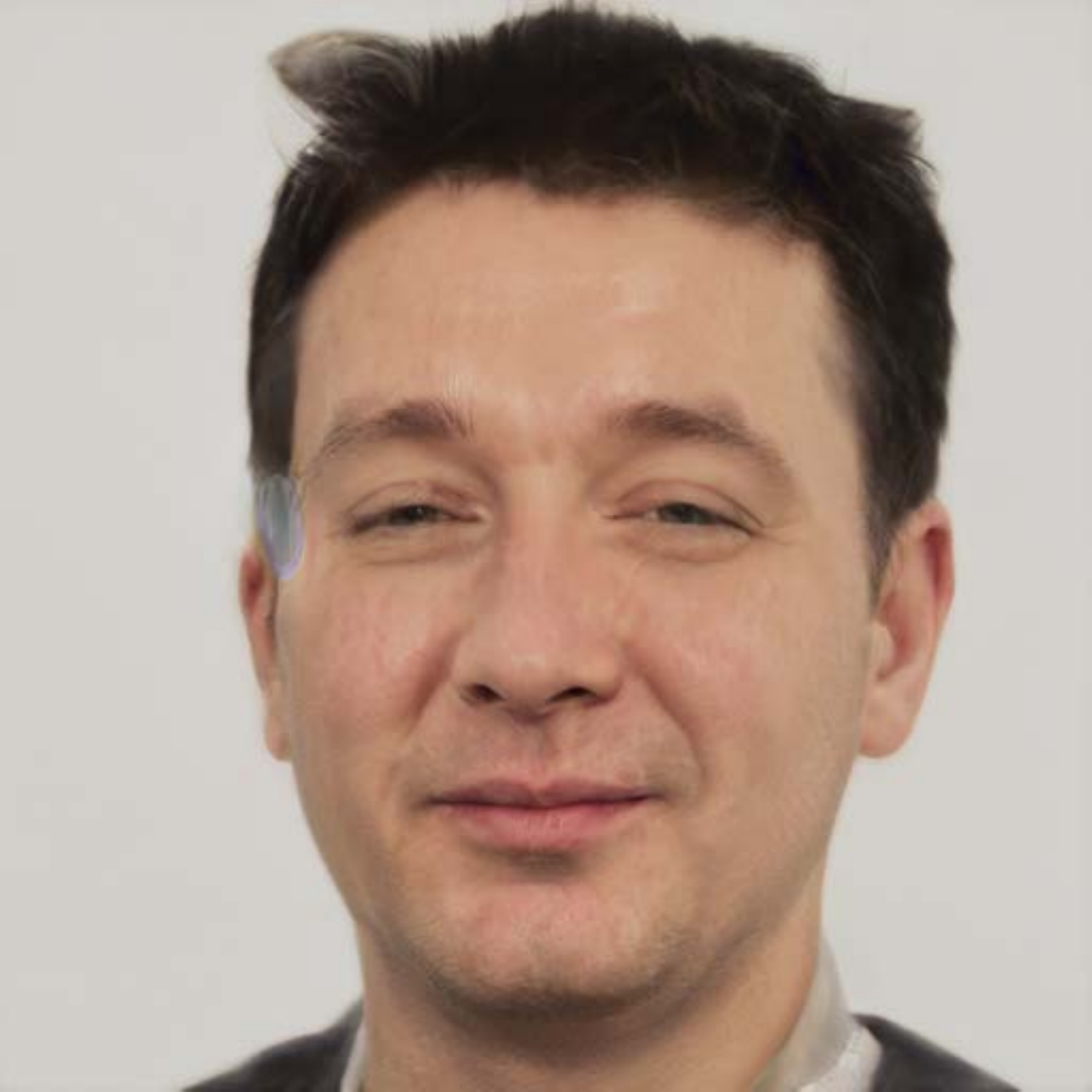}&
   \includegraphics[width=\swseven]{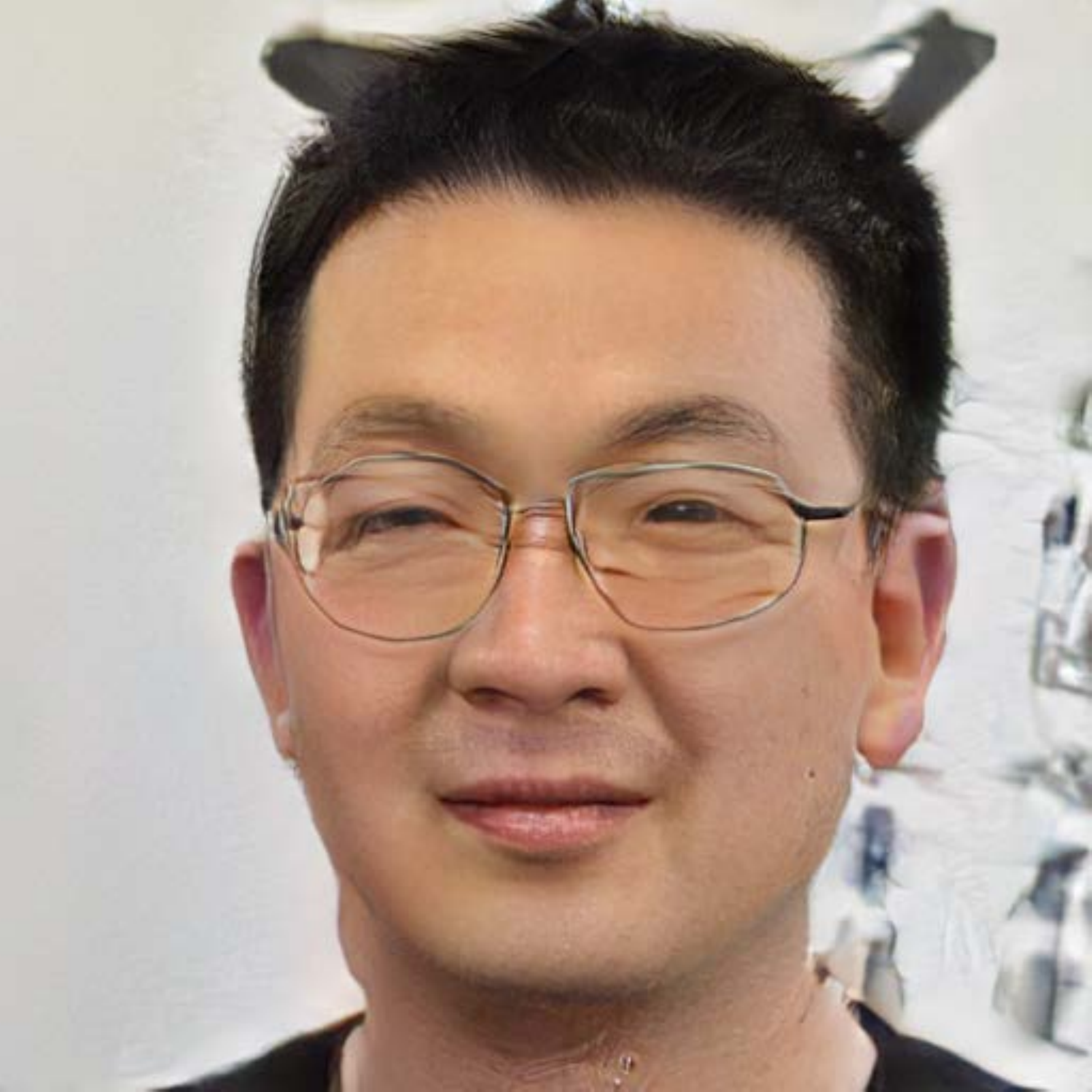}&
   \includegraphics[width=\swseven]{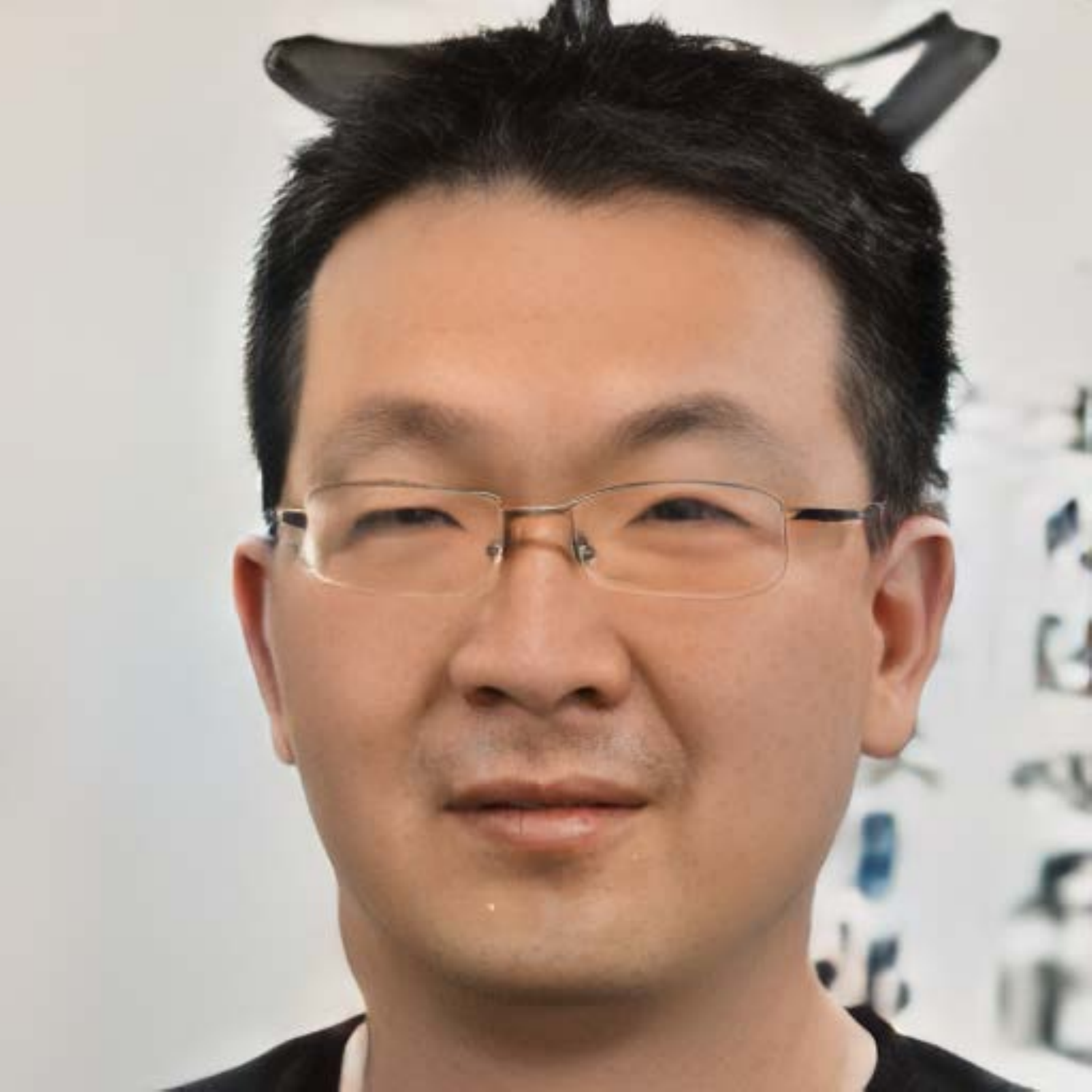}&
   \includegraphics[width=\swseven]{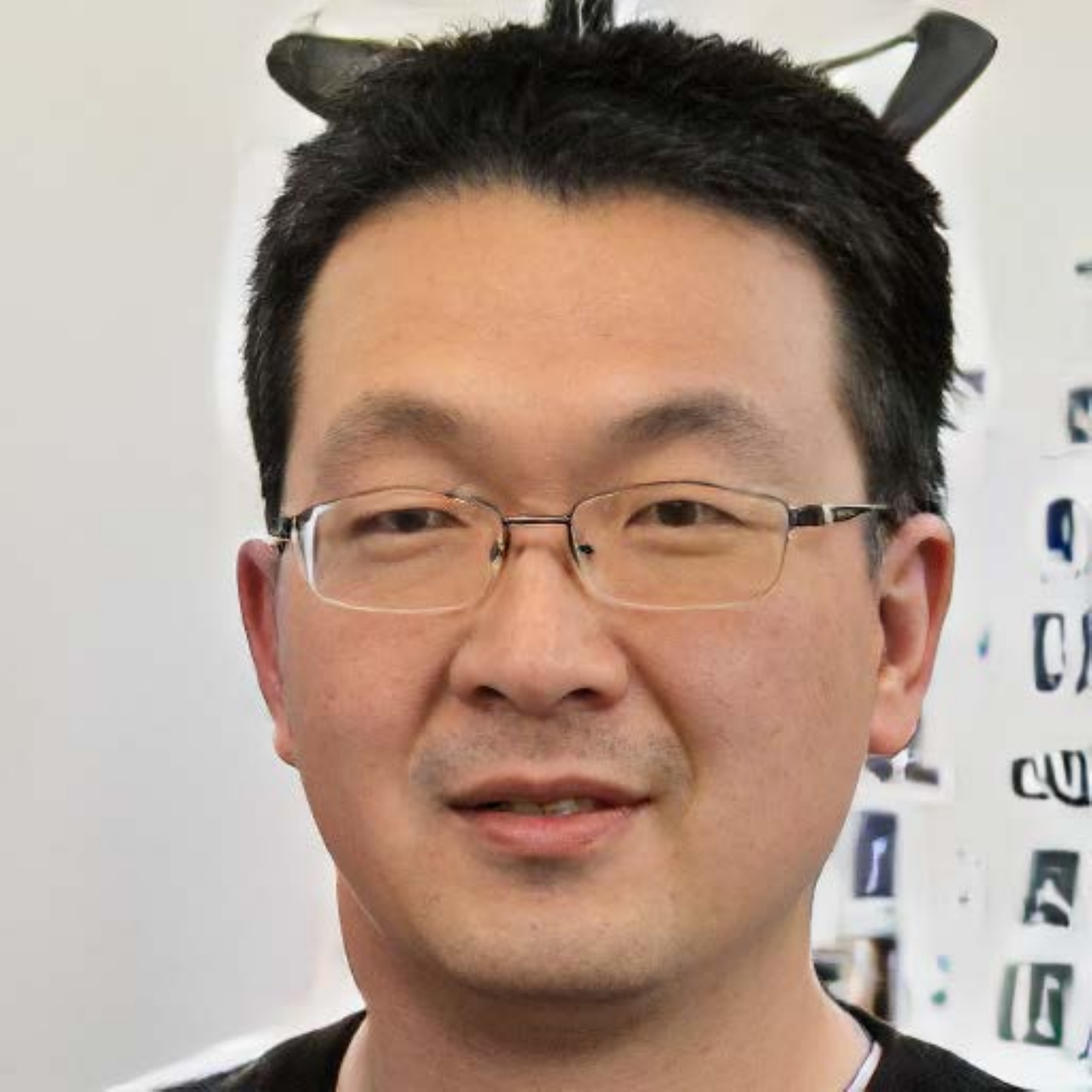} \\
%   \vspace{-1.0mm}
%   \includegraphics[width=\swseven]{figures/figure1/pdf/272_00130_04_org.pdf}&
%   \includegraphics[width=\swseven]{figures/figure1/pdf/272_00130_04_dfd.pdf}&
%   \includegraphics[width=\swseven]{figures/figure1/pdf/272_00130_04_bop.pdf}&
%   \includegraphics[width=\swseven]{figures/figure1/pdf/272_00130_04_pulse.pdf}&
%   \includegraphics[width=\swseven]{figures/figure1/pdf/272_00130_04_pfsrgan.pdf}&
%   \includegraphics[width=\swseven]{figures/figure1/pdf/272_00130_04_gfp_color_align.pdf}&
%   \includegraphics[width=\swseven]{figures/figure1/pdf/272_00130_04_ours.pdf} \\
      \vspace{-1.0mm}
    \includegraphics[width=\swseven]{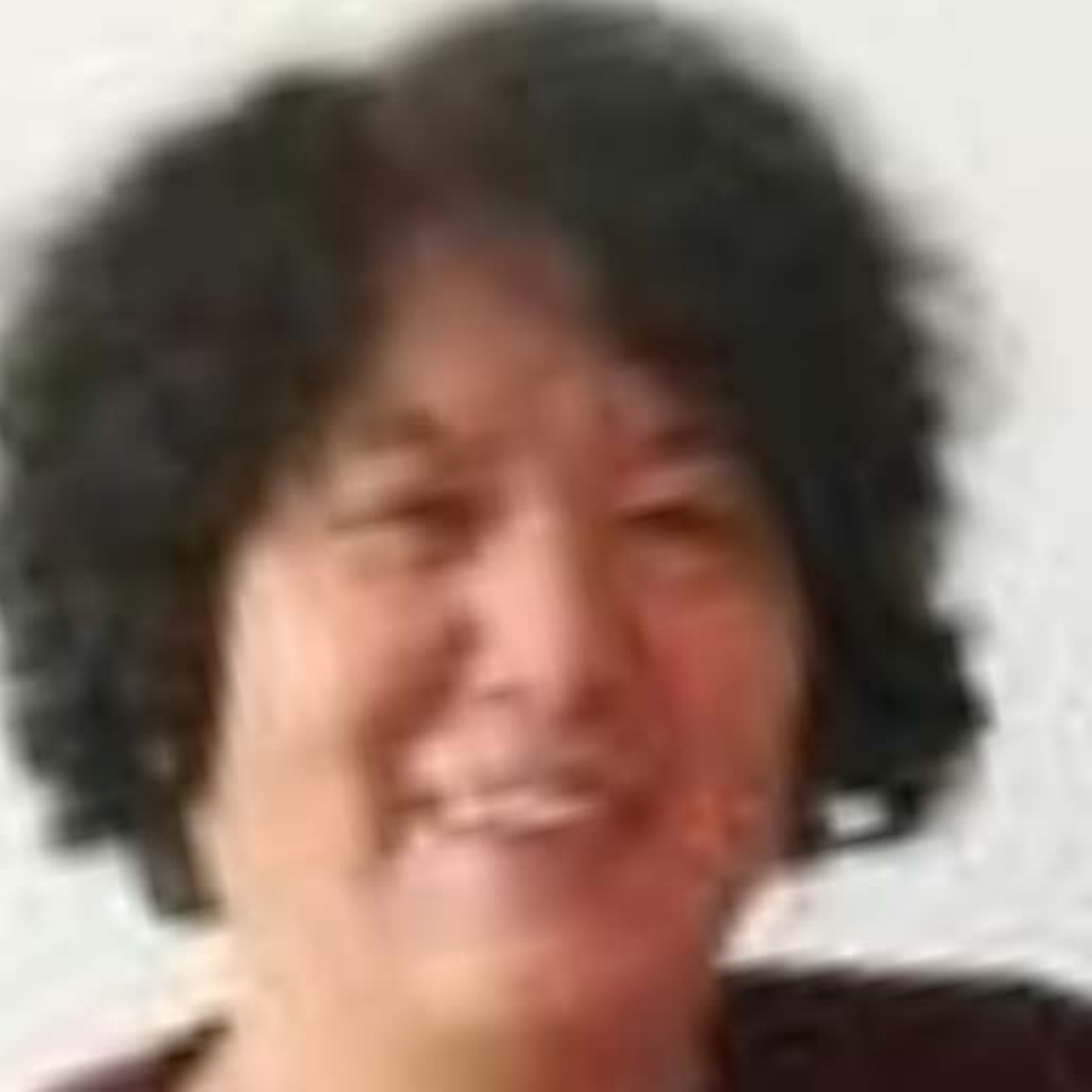}&
    \includegraphics[width=\swseven]{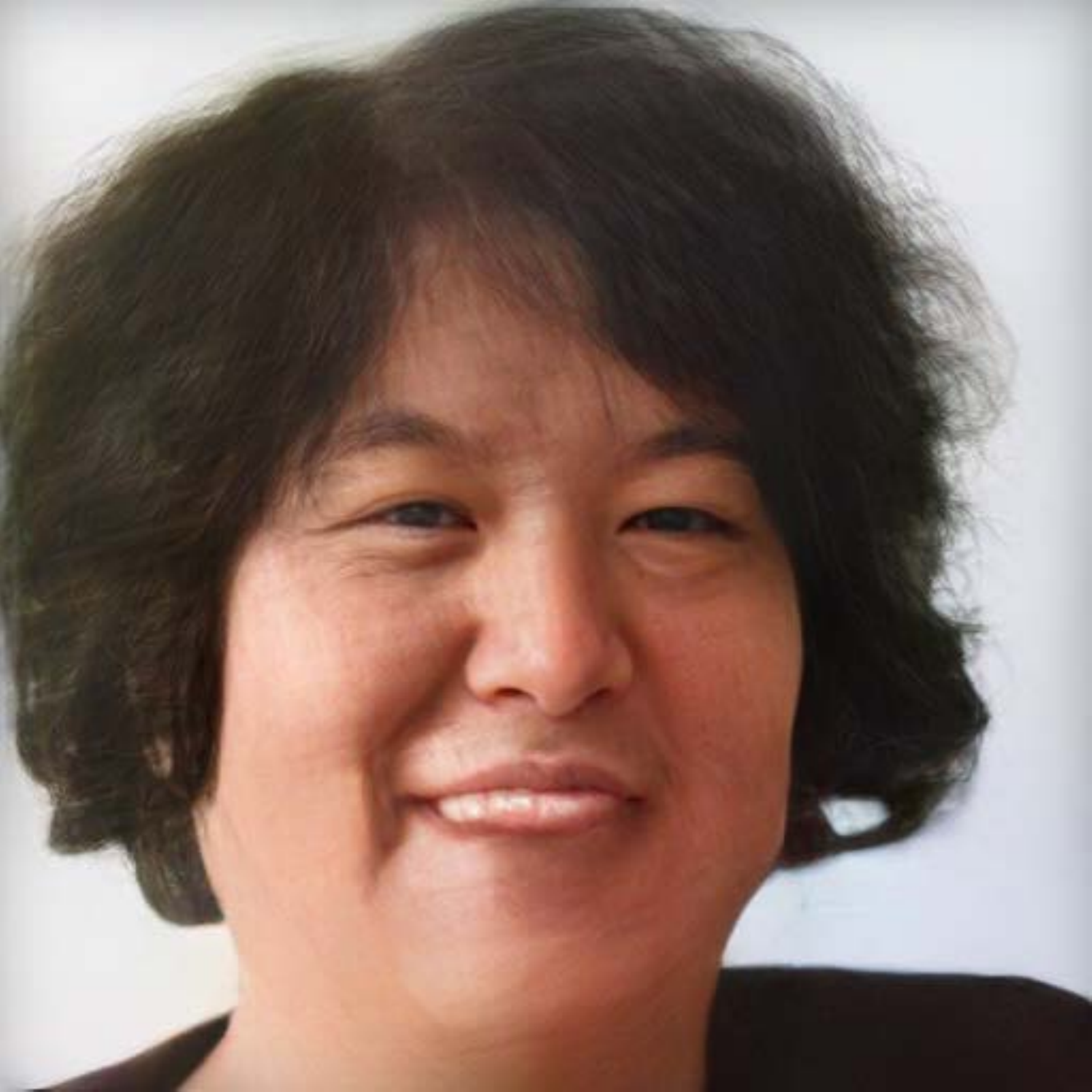}&
    \includegraphics[width=\swseven]{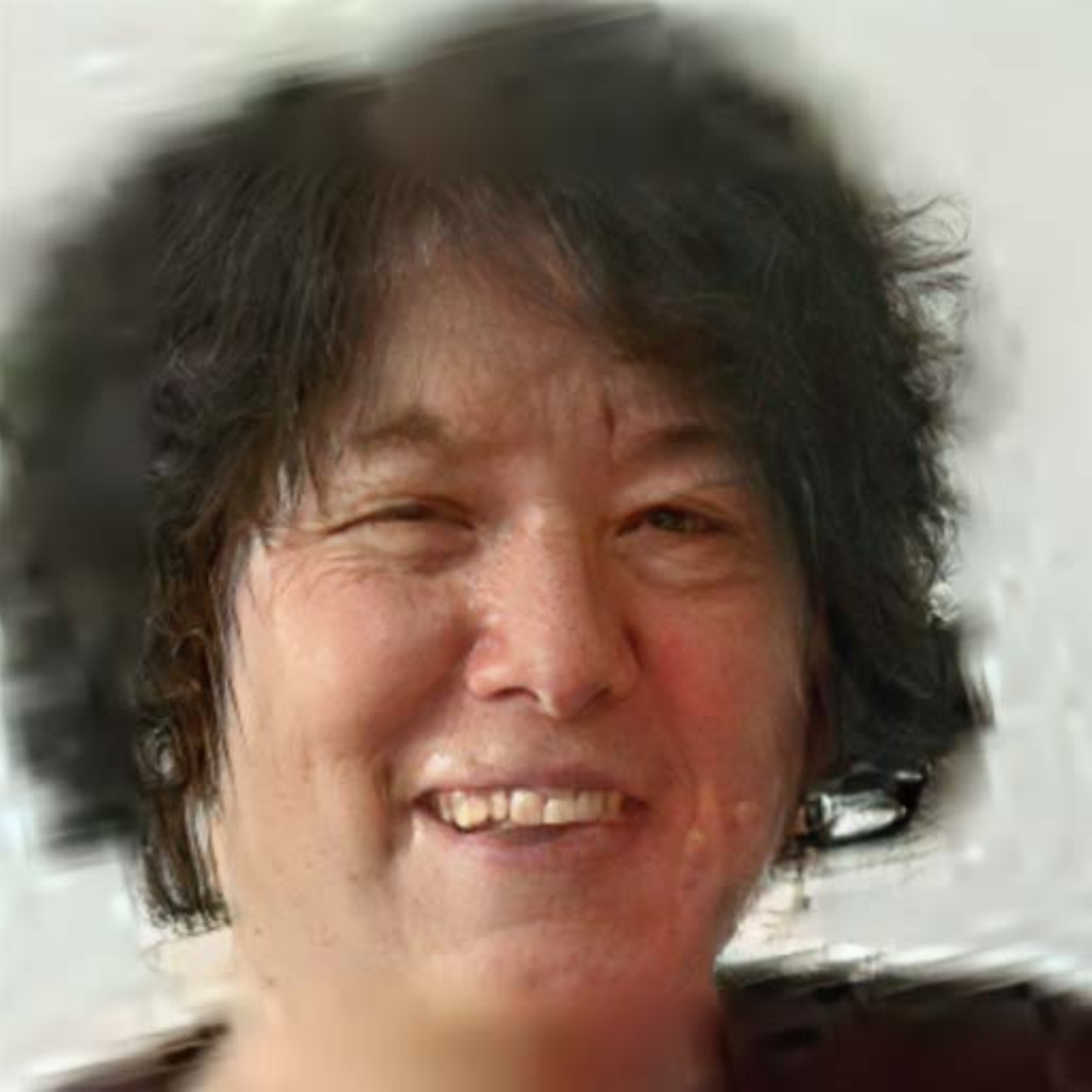}&
    \includegraphics[width=\swseven]{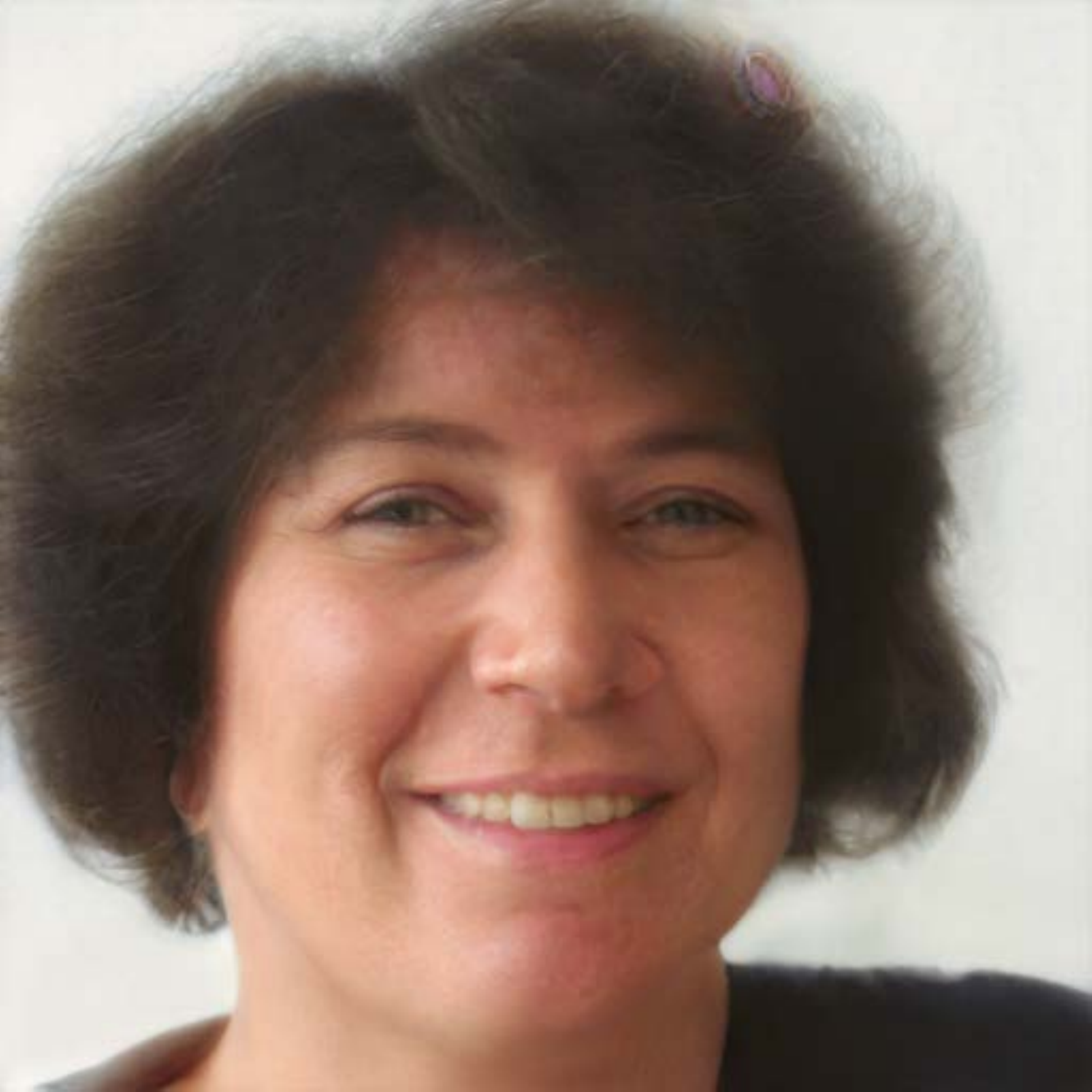}&
    \includegraphics[width=\swseven]{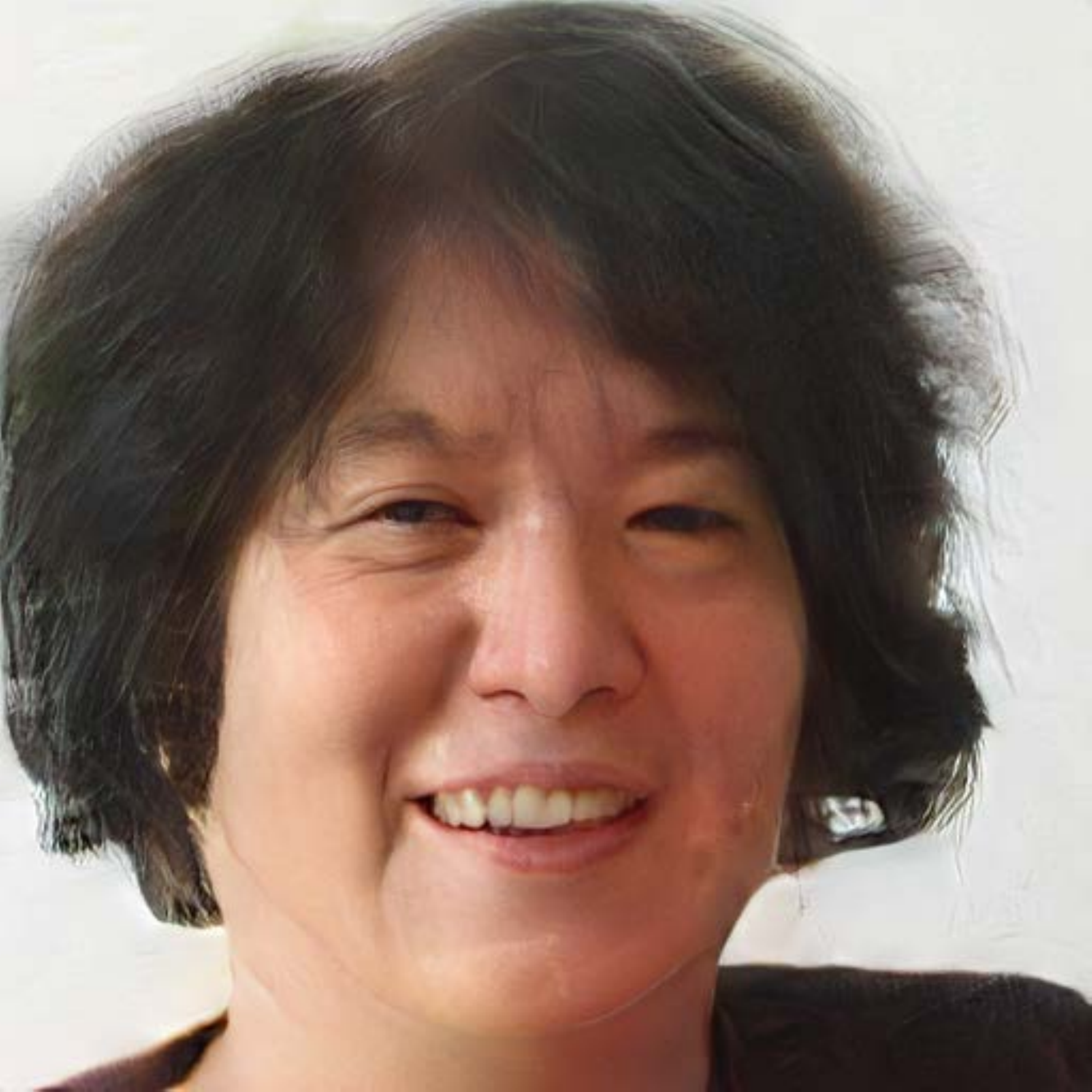}&
    \includegraphics[width=\swseven]{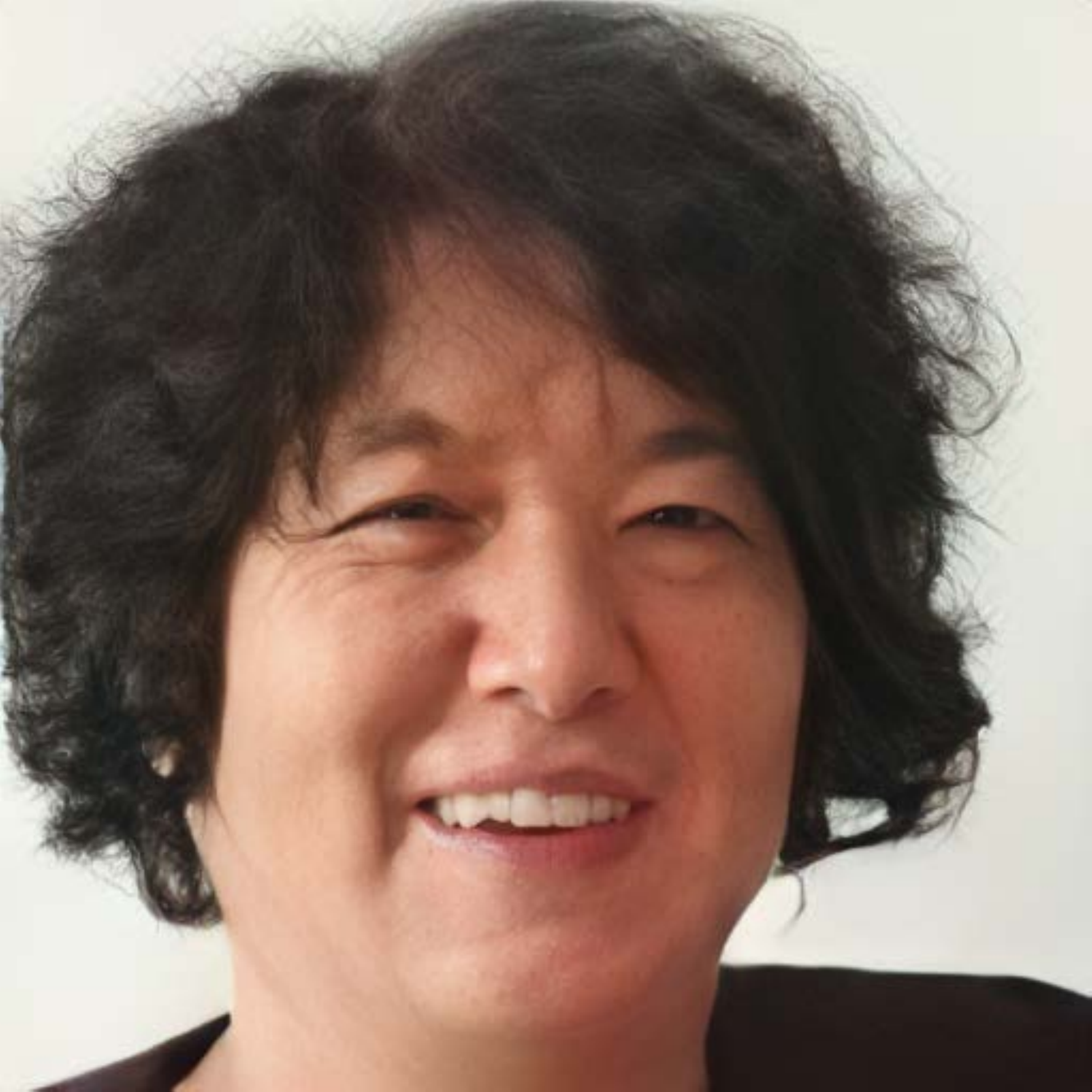}&
    \includegraphics[width=\swseven]{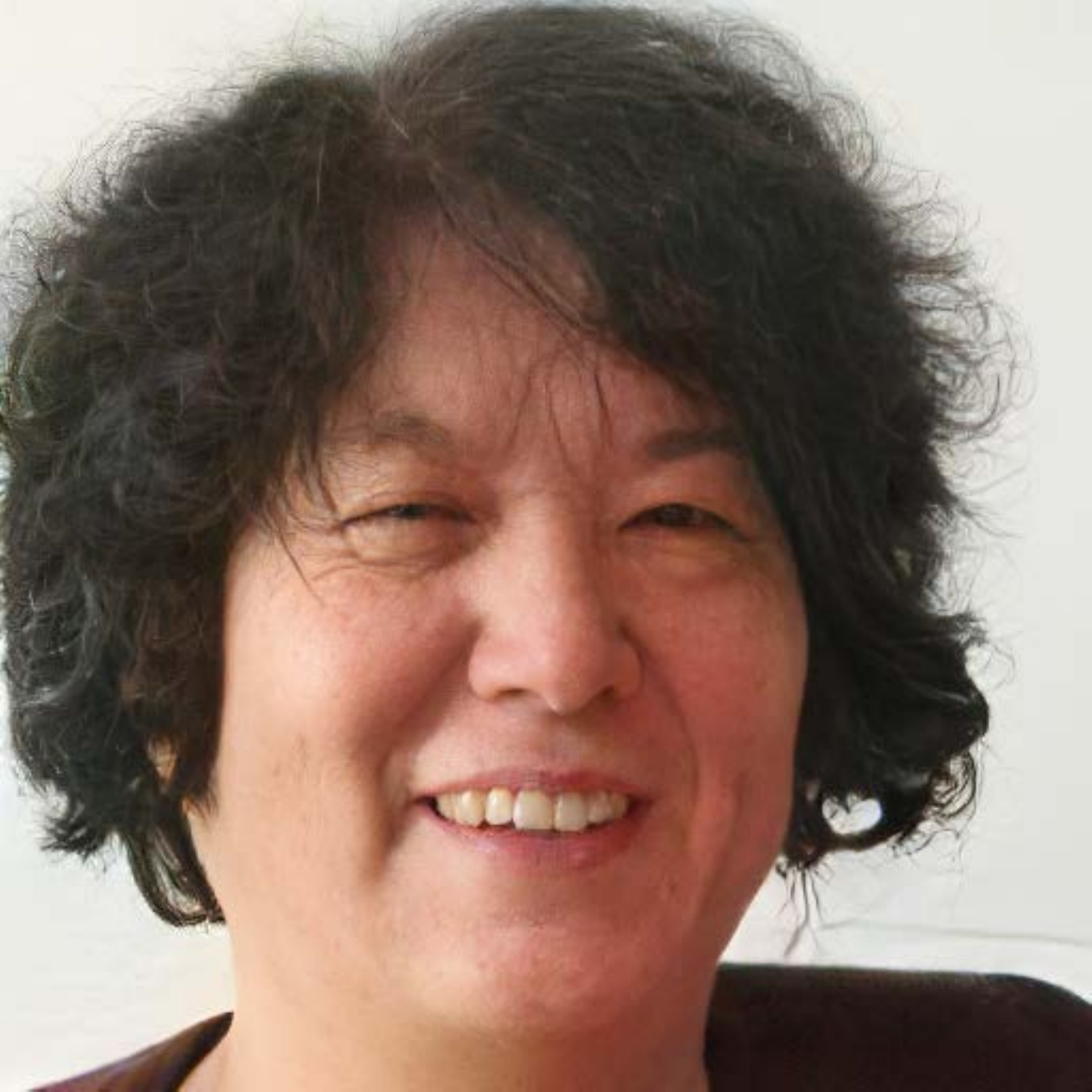} \\
   \vspace{-1.0mm}
   \includegraphics[width=\swseven]{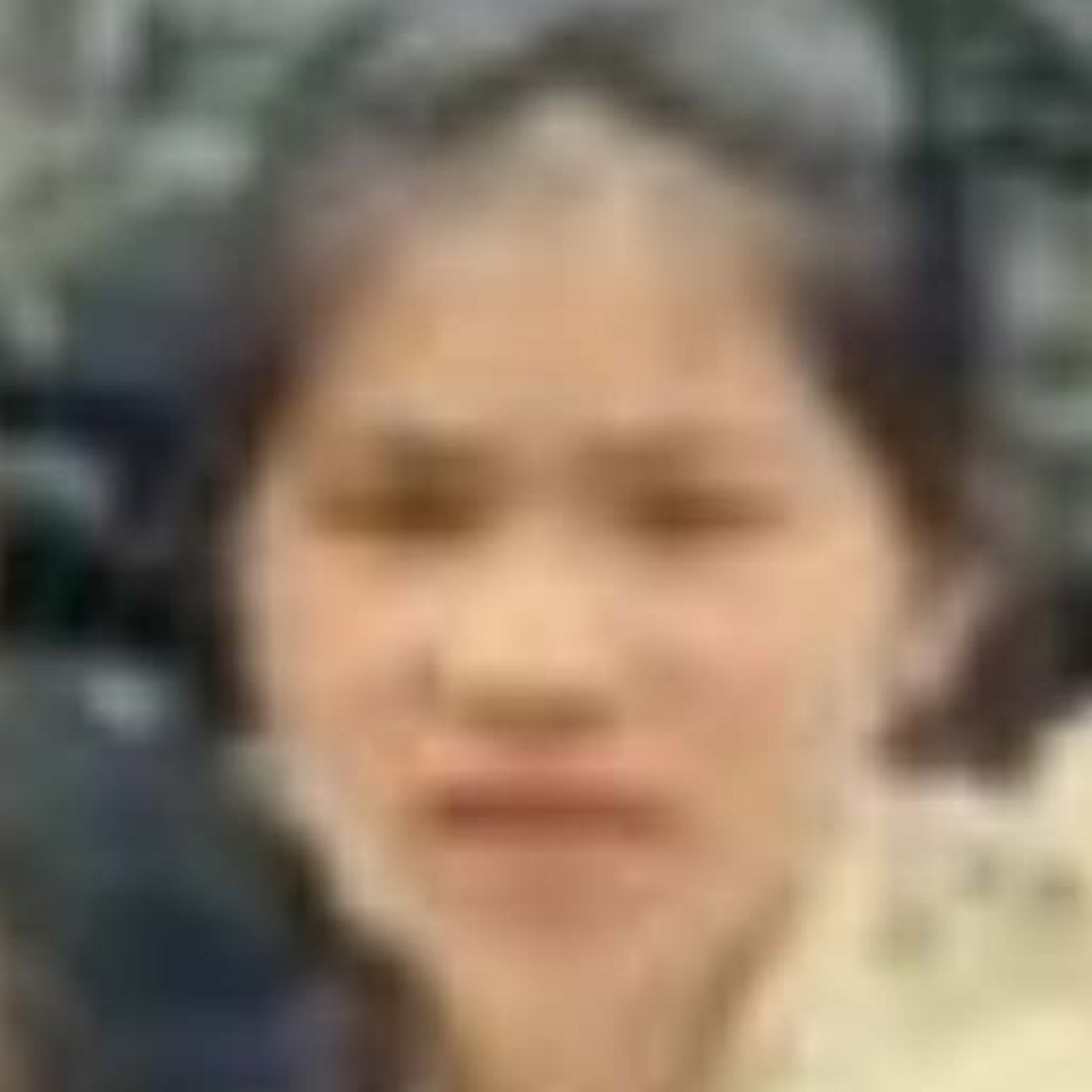}&
   \includegraphics[width=\swseven]{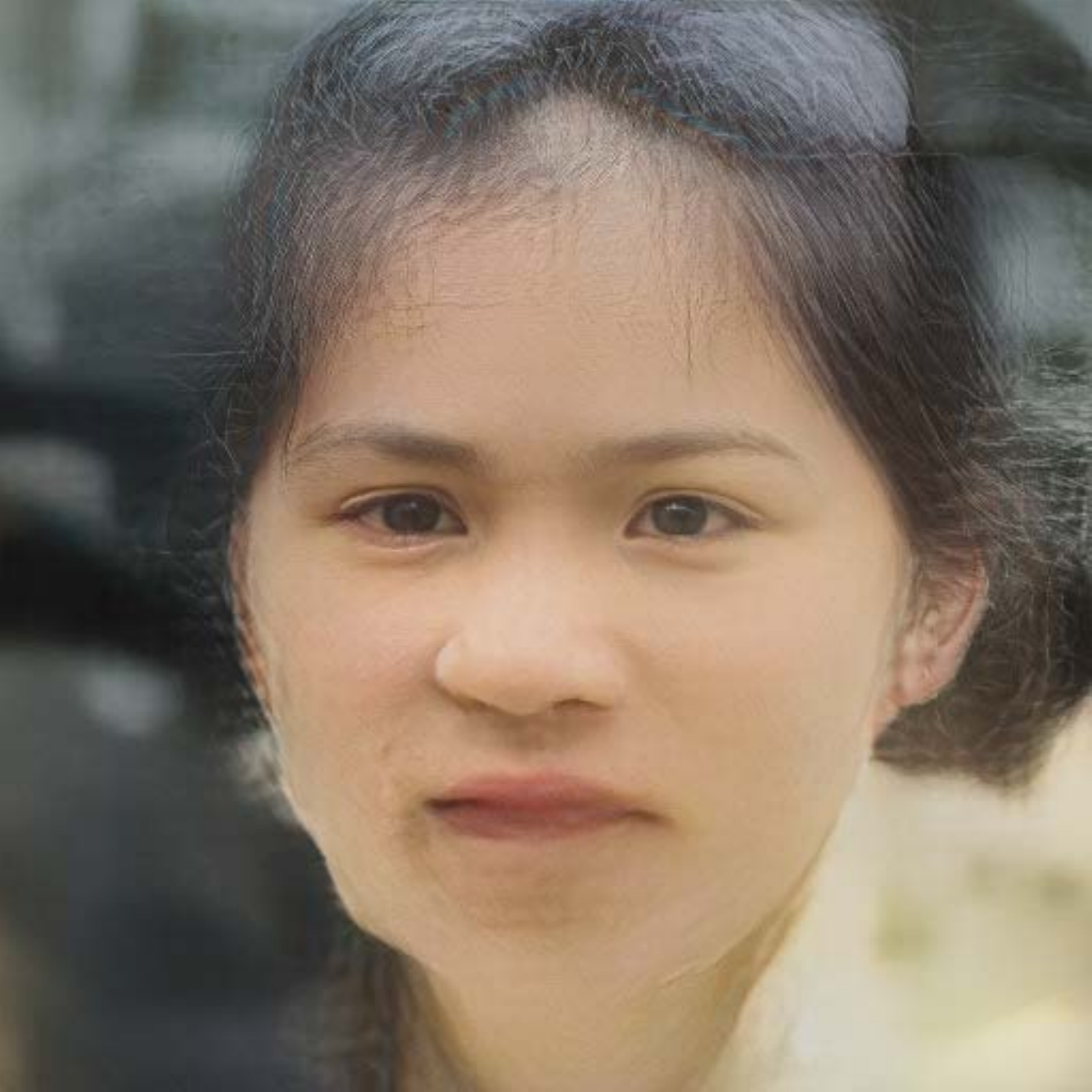}&
   \includegraphics[width=\swseven]{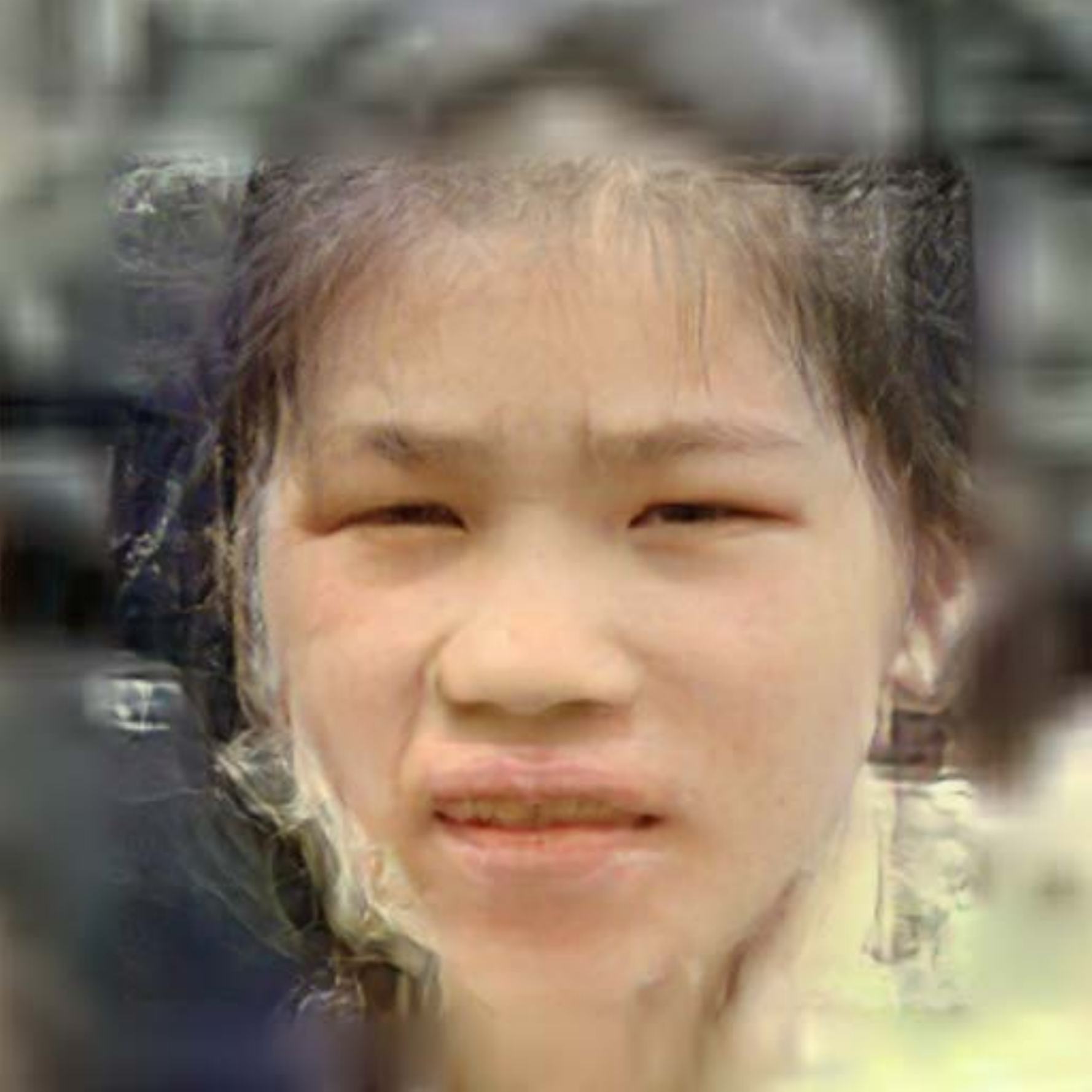}&
   \includegraphics[width=\swseven]{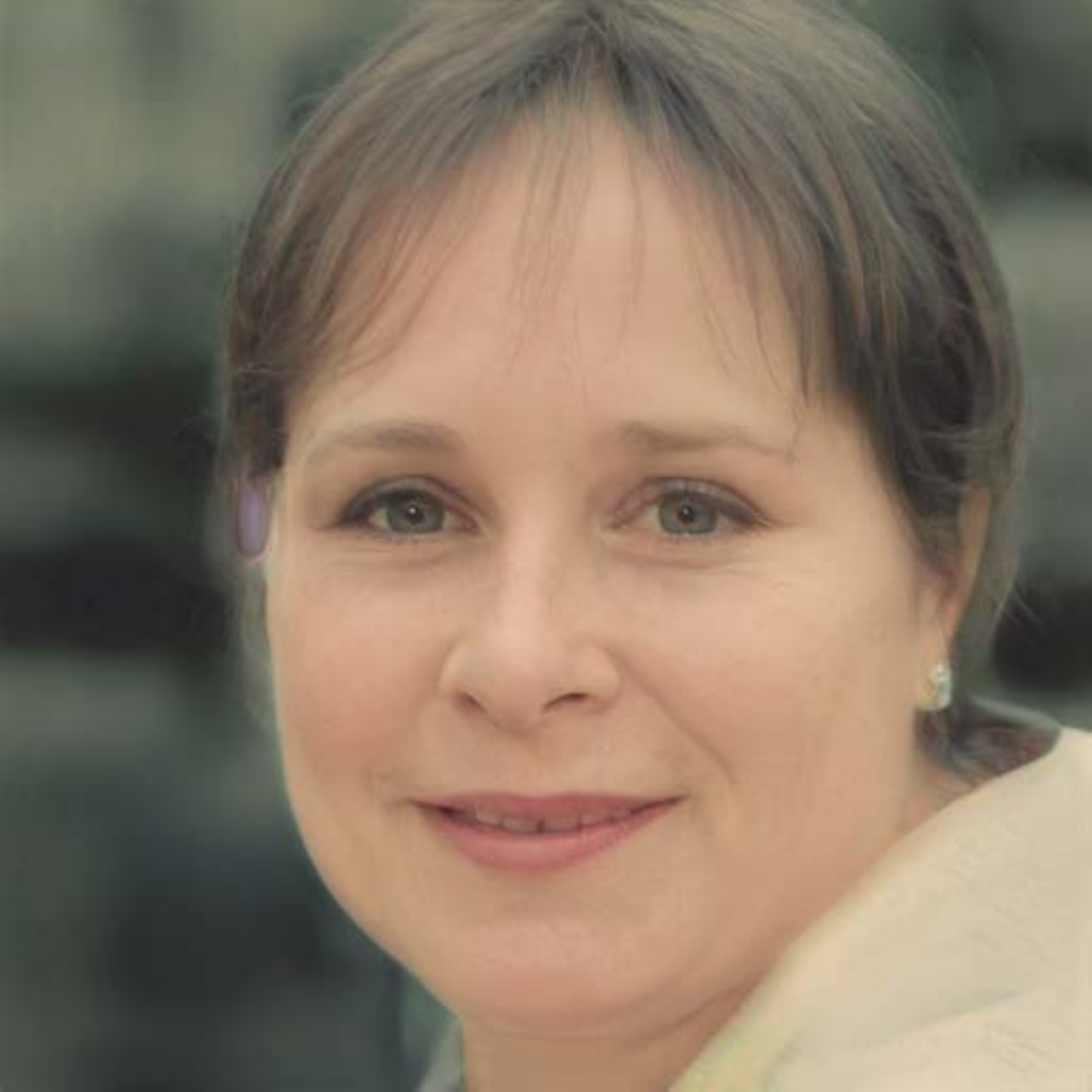}&
   \includegraphics[width=\swseven]{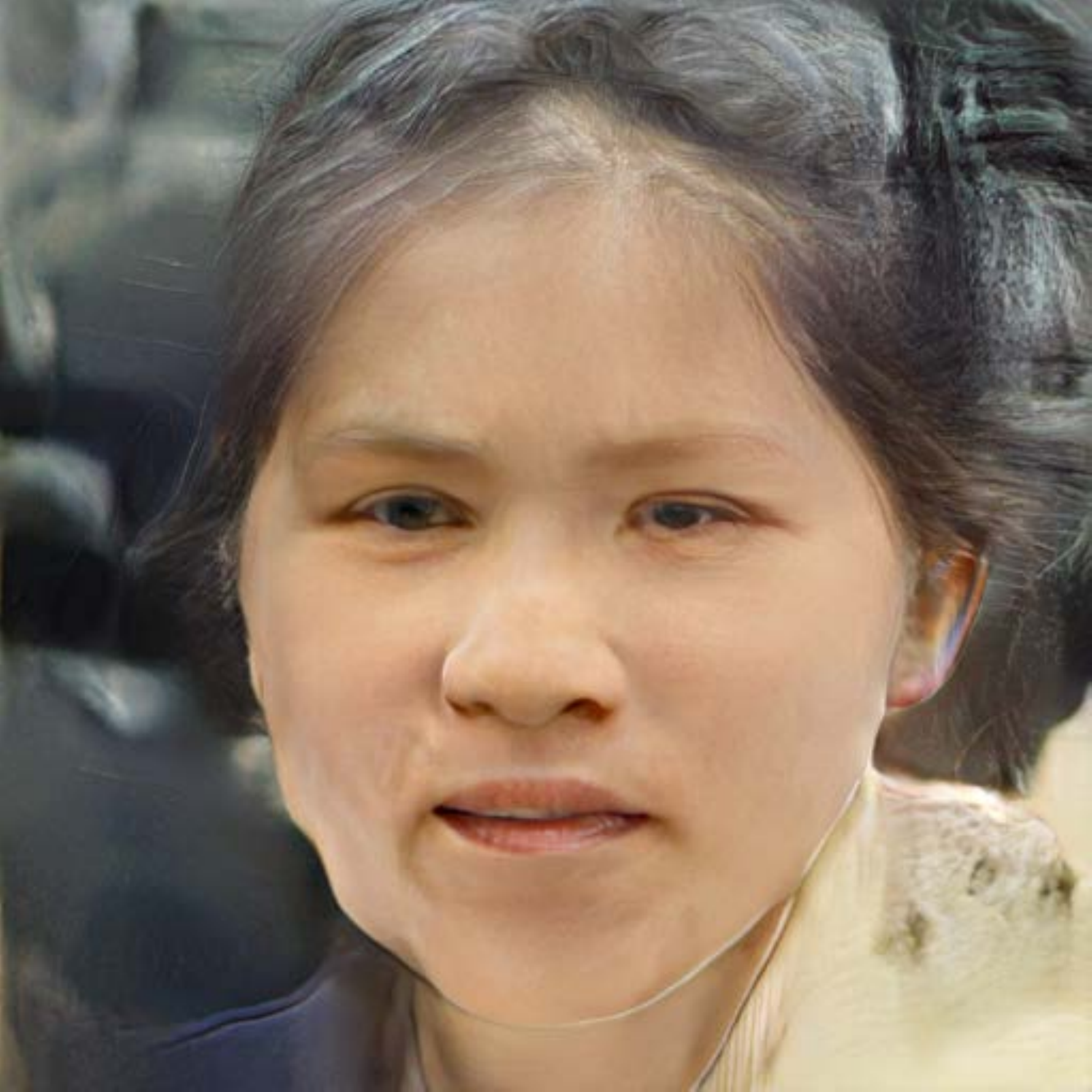}&
   \includegraphics[width=\swseven]{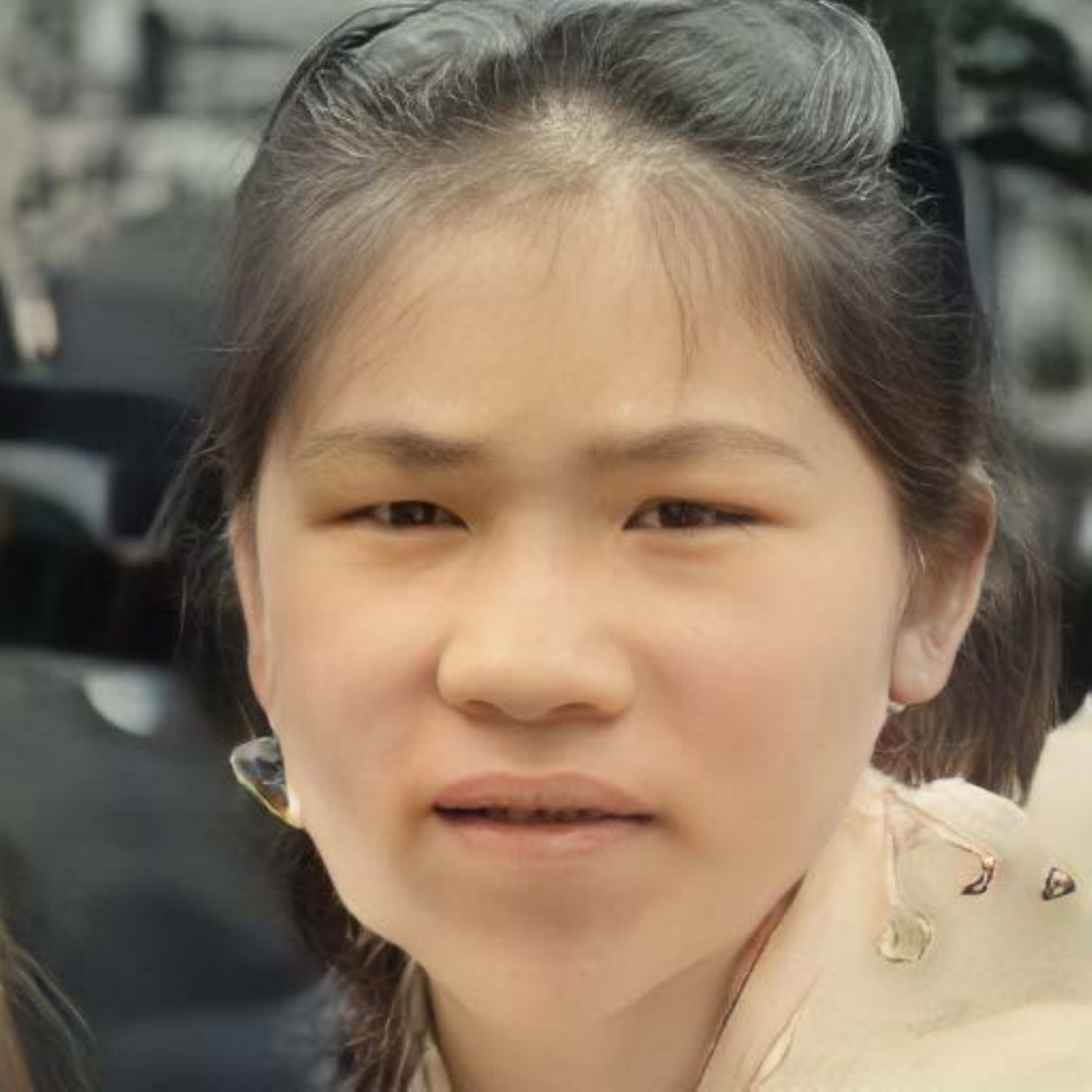}&
   \includegraphics[width=\swseven]{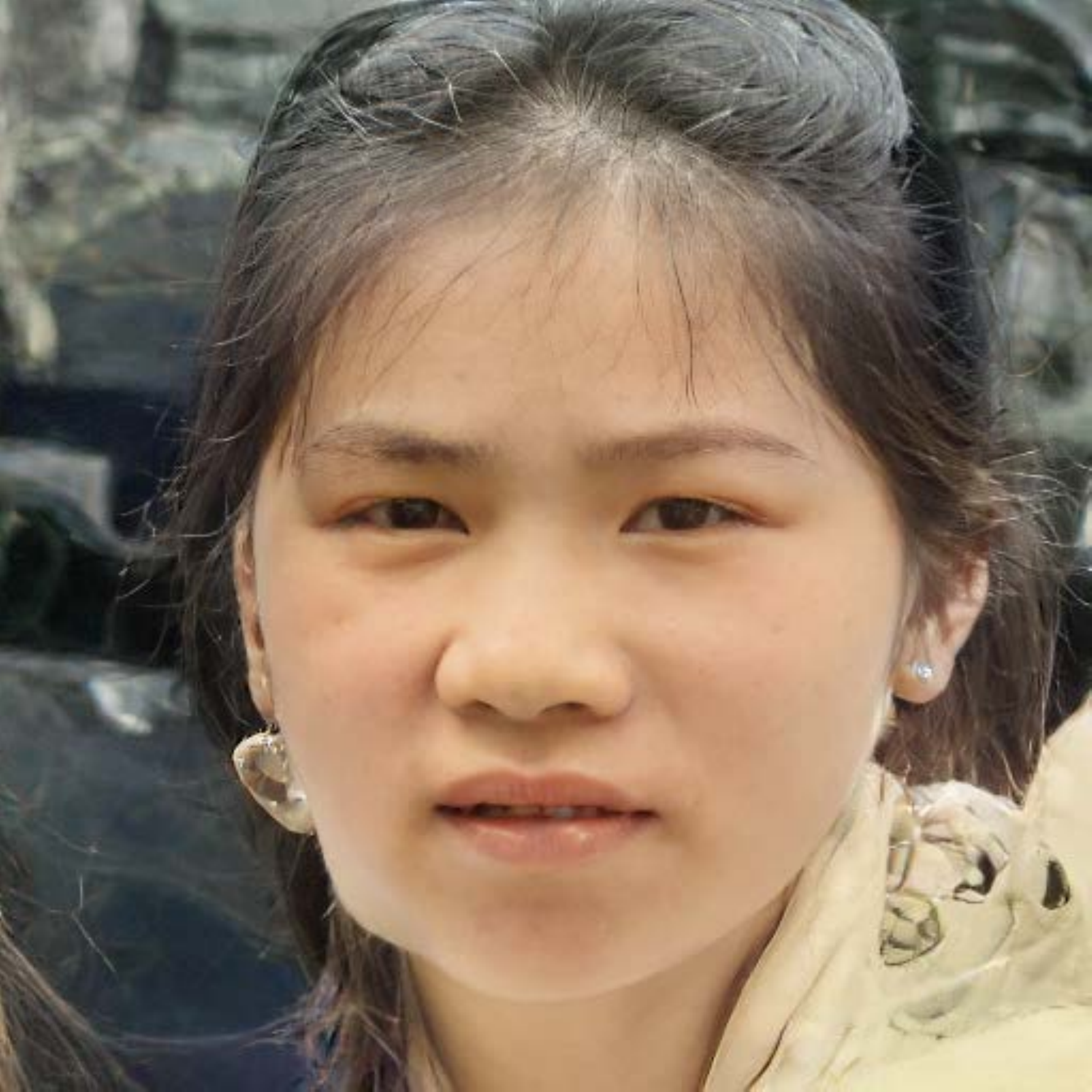} \\
   \label{fig:results}
   Input & DFDNet~\cite{li2020blind}  & Wan~\etal~\cite{wan2020bringing} & PULSE~\cite{menon2020pulse}, & PSFRGAN~\cite{chen2021progressive} & GFP-GAN~\cite{wang2021towards} & \textbf{RestoreFormer} \\
   \textit{real-world} & \textit{ECCV 20} & \textit{CVPR 20} & \textit{CVPR 20} & \textit{CVPR 21}  & \textit{CVPR 21} & \textit{\textbf{Ours}}

\end{tabular}
\end{center}
\vspace{-6mm}
\caption{Comparisons with state-of-the-art face restoration methods on some degraded real-world images. 
%From left to right, they are results of: Input, DFDNet~\cite{li2020blind}, Wan~\etal~\cite{wan2020bringing}, PULSE~\cite{menon2020pulse}, PSFRGAN~\cite{chen2021progressive}, GFP-GAN~\cite{wang2021towards}, and Ours. 
%
The restored results of our RestoreFormer contain more texture details and complete structures which make our results more natural and authentic.}
\label{fig:teaser}
\end{minipage}

\end{figure}%
}]

\blfootnote{*This work is supported by the General Research Fund of HK No.27208720 and 17212120.}

% \vspace{-2.75in}
%\vspace{9.75in}
%%%%%%%%% ABSTRACT
\begin{abstract}
Blind face restoration is to recover a high-quality face image from unknown degradations. 
As face image contains abundant contextual information, we propose a method, RestoreFormer, which explores fully-spatial attentions to model contextual information and surpasses existing works that use local operators. 
%Since different regions of a face image may suffer from different extents of degradations, using facial contextual information is the key to restoring a high-quality image. 
%
%This motivates us to propose a novel method, RestoreFormer, which explores fully-spatial attentions to model contextual information and surpasses existing works that use local convolutions. 
%
RestoreFormer has several benefits compared to prior arts. 
First, unlike the conventional multi-head self-attention in previous Vision Transformers (ViTs), RestoreFormer incorporates a multi-head cross-attention layer to learn fully-spatial interactions between corrupted queries and high-quality key-value pairs. 
Second, the key-value pairs in ResotreFormer are sampled from a reconstruction-oriented high-quality dictionary, whose elements are rich in high-quality facial features specifically aimed for face reconstruction, leading to superior restoration results. 
Third, RestoreFormer outperforms advanced state-of-the-art methods on one synthetic dataset and three real-world datasets, as well as produces images with better visual quality. 
Code is available at https://github.com/wzhouxiff/RestoreFormer.git.

\end{abstract}

%%%%%%%%% BODY TEXT
\section{Introduction}
Blind face restoration aims at restoring a high-quality face from a degraded one that has suffered from complex and diverse degradations, such as down-sampling, blur, noise, compression artifact, \emph{etc}\onedot.
Since the degradations are unknown in the real world, restoration is a challenging task.

Although there are some works~\cite{cao2017attention,huang2017wavelet,xu2017learning} tending to restore high-quality face only based on the information in the degraded one,
most of the existing works have demonstrated that priors play a critical role in blind face restoration.
% the considerable importance of priors, such as 
%
These priors include geometric priors~\cite{chen2021progressive,chen2018fsrnet,kim2019progressive,shen2018deep,yu2018face,yu2018super,zhu2016deep}, references~\cite{dogan2019exemplar,li2020enhanced,li2018learning,li2020blind}, and generative priors~\cite{gu2020image,menon2020pulse,wan2020bringing,wang2021towards}.
Geometric priors can be landmarks~\cite{chen2018fsrnet,kim2019progressive}, facial parsing maps~\cite{chen2021progressive,shen2018deep}, or facial component heatmaps~\cite{yu2018face}.
%, and most of them are predicted from the low-quality faces with an existed predictor.
%
They are considered to be helpful to reconstruct the facial structure.
% although most of them are generated from the degraded faces and their quality is constrained by the quality of the degraded faces.
However, since most of them are estimated from the corrupted faces, their performance is restricted by the quality of the corrupted inputs.
Reference priors are from high-quality exemplars~\cite{dogan2019exemplar,li2020enhanced,li2018learning} or facial component dictionaries~\cite{li2020blind}.
Whereas, the high-resolution exemplars with the same identity of the degraded image are not always accessible and the existing dictionaries-based methods only consider facial components, \emph{e.g}\onedot eyes, mouth, and nose.
%, the reference-based methods still cannot reach a high-performance restoration.%
%At the very beginning, the high-quality references are supposed to be in the same identity with their corresponding degraded faces~\cite{dogan2019exemplar,li2020enhanced,li2018learning}.
%
%For releasing this strict and impractical constraint, Li~\etal~\cite{li2020blind} proposes a component dictionary that consists of amounts of high-quality facial component features and takes it as the unified reference for all the restorations of the corrupted faces.
%
Generative priors encapsulated in a well-trained high-quality face generator are also adopted in blind face restoration.
By exploring an appropriate latent vector from the latent space of a generator~\cite{gu2020image,menon2020pulse} or straightly projecting the degraded face into the latent space~\cite{wan2020bringing,wang2021towards}, their generators are possible to generate a high-quality face with realness.

In these prior-based works, there are two sources of information: the degraded face with identity information and the priors with high-quality facial details.
For restoring faces with realness and fidelity, it is important to fuse these two kinds of information.
% how to fully use these two kinds of information become a critical problem.
%
Most of the existing arts simply combine them by concatenation~\cite{dogan2019exemplar,li2020enhanced,li2018learning}.
Also, there exist works~\cite{chen2021progressive,li2020blind,wang2021towards} proposing to fuse these two kinds of information by Spatial Feature Transformer (SFT)~\cite{wang2018recovering}.% with local convolution operators.
%
%As different areas of a face image may confront different extends of degradations, locally fusing the degraded face and its priors tends to neglect the facial context and end up with sub-optimal restored performance.
%
However, SFT fuses the information pixel-wisely which neglects the abundant facial context and ends up with sub-optimal restored results.
Therefore, we propose a RestoreFormer, which aims for exploring fully-spatial attentions to globally model contextual information and finally transforms the feature from the degraded face into another one close to the ground-truth face feature according to its corresponding high-quality facial priors.
Different from existing ViTs works~\cite{carion2020end,chen2021pre,dosovitskiy2020image,zhu2020deformable} that tend to implement fully-spatial attentions with multi-head self-attention, our RestoreFormer proposes a multi-head cross-attention layer.
Specifically, it takes the features of a corrupted face as queries while their key-value pairs are from high-quality facial priors.
% takes the high-quality facial priors as key-value pairs.
%
By globally and spatially incorporating the corrupted facial features with their corresponding high-quality priors, the proposed method can simultaneously restore a face with realness and fidelity.

Besides, the high-quality dictionary (denoted as HQ Dictionary) proposed in this paper is a reconstruction-oriented one.
It is learned from plenty of undegraded faces by a high-quality face generation network motivated by the idea of vector quantization~\cite{oord2017neural}.
Therefore, it is rich in high-quality facial details that are learned for face restoration.
Compared to the previous Component Dictionaries proposed by Li~\etal~\cite{li2020blind}, whose elements are features of face components generated from amounts of high-quality faces with an off-line approach, our HQ Dictionary has two advantages: 
(1) HQ Dictionary owns rich and diverse details specifically aimed for high-quality face reconstruction,
% since it is learnt from a network that is towards high-quality face reconstruction
while the priors generated with an off-line recognition-oriented model, such as VGG~\cite{simonyan2014very}, may not have such abilities.
(2) HQ Dictionary involves all the areas of a face while the Component Dictionaries~\cite{li2020blind} only provide priors for eyes, nose, and mouth which restrict the ability for face restoration.

In conclusion, our main contributions are as follows: 
% 1) We propose a RestoreFormer to learn fully-spatial interactions between corrupted queries and high-quality key-value pairs which can restore a high-quality face with realness and fidelity. 2) We learn a new HQ Dictionary as priors in RestoreFormer. Its reconstruction-oriented property plays a critical role in face restoration. 3) Extensive experiments show that our RestoreFormer outperforms advanced state-of-the-art methods on both synthetic and real-world datasets, as well as restores faces with better visual quality.

\begin{itemize}
    % \vspace{-1mm}
    \item We propose a RestoreFormer to learn fully-spatial interactions between corrupted queries and high-quality key-value pairs which can attain a high-quality face with realness and fidelity from a degraded face.
    % \vspace{-1mm}
    \item We learn a new HQ Dictionary as priors in RestoreFormer. Its reconstruction-oriented property plays a critical role in face restoration.
    % \vspace{-1mm}
    \item Extensive experiments show that our RestoreFormer outperforms advanced state-of-the-art methods on both synthetic and real-world datasets, as well as restores faces with better visual quality.
\end{itemize}

\section{Related Works}
\textbf{Blind Face Restoration}
Blind face restoration aims at restoring high-quality faces from complex and unknown degradations.
Previous works have shown that additional priors play a critical role in this task and they can be coarsely categorized into three types: 
%down-sampling, blur, noise, compression artifact, and so on.
%
% While deep neural network (DNN) has shown its superiority over all kinds of computer vision tasks, researchers also attempt to apply DNN for blind face restoration and got outstanding performance~\cite{cao2017attention,huang2017wavelet,xu2017learning}.
% Although the deep neural network has shown its superiority on blind face restoration~\cite{cao2017attention,huang2017wavelet,xu2017learning}, most of the previous works show that with the help of additional priors, 
% %
% However, since degradations in this task are complex and unpredictable, previous works find that additional priors are very helpful for the restoration of facial structure or details.
% 
%
% These priors coarsely include 
geometric priors~\cite{chen2021progressive,chen2018fsrnet,kim2019progressive,li2019recovering,shen2018deep,yu2018face,yu2018super,zhu2016deep}, references~\cite{dogan2019exemplar,li2020enhanced,li2018learning,li2020blind}, and generative priors~\cite{gu2020image,menon2020pulse,wan2020bringing,wang2021towards}.

The methods based on geometric priors tend to progressively restore faces with landmark heatmaps~\cite{chen2018fsrnet, kim2019progressive} or facial component heatmaps~\cite{chen2021progressive,shen2018deep}.
%
% Geometric priors used for blind face restoration include landmark~\cite{chen2018fsrnet,kim2019progressive}, facial parsing maps~\cite{chen2021progressive,shen2018deep}, and facial component heatmaps~\cite{yu2018face}.
%
% Especially, progressively restoring faces with landmark heatmaps~\cite{chen2018fsrnet, kim2019progressive} or facial component heatmaps~\cite{chen2021progressive,shen2018deep} are the widely-used methods in most of the previous works based on geometric priors.
%
% For instance, Chen~\etal~\cite{chen2018fsrnet} attain landmark heatmaps and facial parsing maps from a coarse super-resolved face for further finetuning the restored face;
%
% Kim~\etal~\cite{kim2019progressive} progressively restore faces with extra landmark heatmaps.
% Kim ~\etal ~\cite{kim2019progressive} used extra landmark heatmaps to guide the restoration of face progressively;
%
% Both Shen~\etal~\cite{shen2018deep} and Chen~\etal~\cite{chen2021progressive} design a neural network that takes the degraded face and its corresponding facial component heatmaps as input and attains a high-quality face progressively.
%
Since these geometric priors are mainly generated from low-quality faces, the corrupted face limits the performance of restoration.
%
% Besides, most of these methods fuse the corrupted face and its geometric prior by locally concatenating, which have neglected the rich facial context in the face.
%
On the other hand, reference-based works need the references to be in the same identity with the degraded face, which is not always accessible~\cite{dogan2019exemplar,li2018learning,li2020enhanced}.
% set a strict assumption that the high-quality references are supposed to be in the same identity with the degraded face, no matter there is only one reference~\cite{dogan2019exemplar,li2018learning} or multiple diversity exemplars~\cite{li2020enhanced}.
%
% The assumption is unpractical, especially for the severely corrupted face whose identity cannot be recognized.
% and their similar fusion method leads to the same shortage as above geometric-based works.
%
Although Li~\etal~\cite{li2020blind} alleviate this constraint by collecting component dictionaries consisting of high-quality facial component features as general references, 
% The work proposed by Li~\etal~\cite{li2020blind} alleviates this constraint. They collect component dictionaries consisting of high-quality facial component features and take them as general references for blind face restoration.
% %
% However, unlike our reconstruction-oriented HQ Dictionary that is rich in high-quality facial details aimed for face reconstruction, 
the facial details in these component dictionaries are limited since they are extracted with an off-line recognition-oriented model and only focus on some facial components.
%
% %
% It attains an improvement compared to its previous works. However, since its dictionary only focuses on certain facial components (eyes, nose, and mouth) and is generated off-line, there is still a big room for further improvement.
%
Besides, some works tend to exploit the generative priors encapsulated in a high-quality face generation model for blind face restoration.
They implement it by exploring a latent vector with an expensive target-specific optimization~\cite{menon2020pulse} or projecting the degraded face into the latent space directly\cite{wan2020bringing,wang2021towards}.
% Generative priors are a kind of information encapsulated in a well-trained high-quality face generation model.
%
%
% Menon~\etal~\cite{menon2020pulse} explores a latent vector of a GAN model with an expensive target-specific optimization while Wan~\etal~\cite{wan2020bringing} and Wang~\etal~\cite{wang2021towards} straightly project the degraded face into the latent space.
%
%Since they~\cite{menon2020pulse,wan2020bringing} almost leave the identity information including in the corrupted face aside, their restored faces lack fidelity.
As they~\cite{menon2020pulse,wan2020bringing} fail to consider the identity information during training, their restored lack fidelity.
% and that is the reason why their results tend to be faithful but lack fidelity.
%
Although Wang~\etal~\cite{wang2021towards} combine their generative priors with the degraded face with a spatial feature transformer layer, the locally combining method ignores the rich facial context in the face image
%
%Instead, our proposed RestoreFormer is designed with a multi-head cross-attention layer for learning fully-spatial interactions between corrupted queries and high-quality key-value pairs which are sampled from a reconstruction-oriented high-quality dictionary that has appealing benefits compared to the previous prior dictionaries. 

\textbf{Vision Transformer}
Transformer is a kind of deep neural network originally used in natural language processing field~\cite{brown2020language,devlin2018bert,vaswani2017attention}.
Due to its competitive representation ability, it begins to be applied to computer vision tasks, such as recognition~\cite{dosovitskiy2020image}, detection~\cite{carion2020end,zhu2020deformable}, and segmentation~\cite{wang2021max}.
The low-level vision tasks also get benefits from it in \cite{chen2021pre,esser2021taming,parmar2018image,yang2020learning,zhao2021improved,zhusketch}.
Chen~\etal~\cite{chen2021pre} exploits the advantage of the transformer on large scale pre-training to construct a complex model covered several image processing tasks, such as denoise, deraining, and super-resolution.
Esser~\etal~\cite{esser2021taming} apply the transformer to generate a high-resolution image by predicting a sequence of codebook-indices of their encoders, which makes full use of the strong representative capacity of the transformer within an acceptable computational resource.
In \cite{zhusketch}, Zhu~\etal adopt the transformer to obtain the global structure of the face which is helpful for photo-sketch synthesis.
%
%These works mainly focus on the attention within the input image, while our RestoreFormer tends to model the interaction between the degraded input and its high-quality priors with a multi-head cross-attention layer. Extra delicate priors play an important role in blind face restoration.

\section{Methodology}
This section introduces the proposed RestoreFormer for restoring high-quality faces from unknown degradations with an HQ Dictionary consisting of reconstruction-oriented high-quality priors.
The whole pipeline is shown in Figure~\ref{fig:framework}~(c).
An encoder $\mathbf{E}_d$ is first deployed to extract representation $\bm{Z}_d$ of the degraded face $\bm{I}_d$ and its nearest high-quality priors $\bm{Z}_p$ are fetched from the HQ Dictionary $\mathbb{D}$.
% $\mathbb{D}$ by minimizing the distance between the priors in $\mathbb{D}$ and feature vectors in $\bm{Z}_d$.
%
Then two consecutive transformers implemented with multi-head cross-attention (denoted as MHCA) are utilized to fuse the features of degraded images and priors.
Finally a decoder $\mathbf{D}_d$ is applied on the fused representation $\bm{Z}'_f$ to restore a high-quality face $\bm{\hat{I}}_d$.
Details of each step will be presented in Sec.~\ref{subsec: RestoreFormer}.

To obtain the HQ Dictionary $\mathbb{D}$, we incorporate the idea of vector quantization~\cite{oord2017neural} and propose a high-quality face generation network to learn $\mathbb{D}$ from plenty of undegraded faces.
Compared to previous works~\cite{li2020blind} whose component dictionaries are extracted with an off-line recognition model VGG~\cite{simonyan2014very}, the priors in $\mathbb{D}$ are reconstruction-oriented and can provide rich facial details for the restoration of degraded faces.
The specific procedure of getting the reconstruction-oriented HQ Dictionary will be introduced in Sec.~\ref{subsec: HQ dict}.

\subsection{RestoreFormer}
\label{subsec: RestoreFormer}
Even though facial image contains abundant global contextual information, \emph{e.g}\onedot eyes and teeth, the existing arts~\cite{chen2021progressive,li2020blind,wang2021towards} only apply local operators for blind face restoration.
Recently, ViT (Vision Transformer)~\cite{vaswani2017attention} is proposed to consider the contextual information in images.
However, most of the ViT-based methods~\cite{carion2020end,chen2021pre,dosovitskiy2020image,zhu2020deformable} only consider one source of information, \emph{i.e}\onedot the degraded face in our task, by multi-head self-attention (namely MHSA) and it cannot be directly applied into face restoration which needs to combine the information from degraded image and priors.
%
%Since different regions of a face tend to suffer from different extents of degradations, both rich facial context and extra facial priors are critical for blind face restoration.
%
%On the contrary to previous works~\cite{chen2021progressive,li2020blind,wang2021towards} that only locally fuse the degraded information and its corresponding priors, the proposed RestoreFormer consider the rich facial context and seek to fully-spatially incorporate these two sources of information with transformers \cite{vaswani2017attention}.
%
%However, most of ViT (Vision Transformer)~\cite{carion2020end,chen2021pre,dosovitskiy2020image,zhu2020deformable} only consider one source of information by multi-head self-attention mechanism (namely MHSA-TF), making them infeasible in image restoration.
%
Thus, we propose transformers with the multi-head cross-attention mechanism (MHCA) to fully-spatially fuse two sources of information to restore face with realness and fidelity.% for taking both facial context and extra priors into consideration.
% benefiting the realness and fidelity of our final restored faces.
%
% Considering the rich facial context ignored by the previous works~\cite{chen2021progressive,li2020blind,wang2021towards} that mainly locally fuse the degraded information and its corresponding priors, the proposed RestoreFormer tend to fully-spatially incorporate these two sources of information with transformers implemented with multi-head cross-attention mechanism (MHCA-TF).
% %
% Also different from the multi-head self-attention mechanism (namely MHSA-TF) in most of the previous ViTs~\cite{carion2020end,chen2021pre,dosovitskiy2020image,zhu2020deformable} that mainly focus on searching globally contents in one source of information, our MHCA-TF aims for fully-spatially fusing two sources of information (the features of degraded face and its priors) for benefiting the realness and fidelity of our final restored faces.
%
%We have some delicate designs in MHCA-TF towards blind face restoration.
In this subsection, we first explain the MHCA by comparing it with MHSA and then give a detailed description of RestoreFormer built upon MHCA.
%
%For presenting a comprehensive understanding of our MHCA-TF, we first explain the MHCA-TF by comparing it with MHSA-TF before giving a specific introduction of RestoreFormer.

\renewcommand{\tabcolsep}{5pt}
\begin{figure}
\begin{center}
% \begin{tabular}{cc}
% \includegraphics[width=0.43\linewidth]{figures/fig2_a1.pdf} & \includegraphics[width=0.43\linewidth]{figures/fig2_a2.pdf} \\
% \textbf{MHSA-TF} & \textbf{MHCA-TF} \\
% \multicolumn{2}{c}{(a) \textbf{MHSA-TF} and \textbf{MHCA-TF}} \\
% % \hdashrule[0.5ex]{4cm}{1pt}{1pt} \\
% \multicolumn{2}{c}{\includegraphics[width=0.92\linewidth]{figures/fig2_b.pdf}}\\
% \multicolumn{2}{c}{(b) \textbf{RestoreFormer}} \\
\begin{tabular}{cccc}
\includegraphics[width=0.27\linewidth]{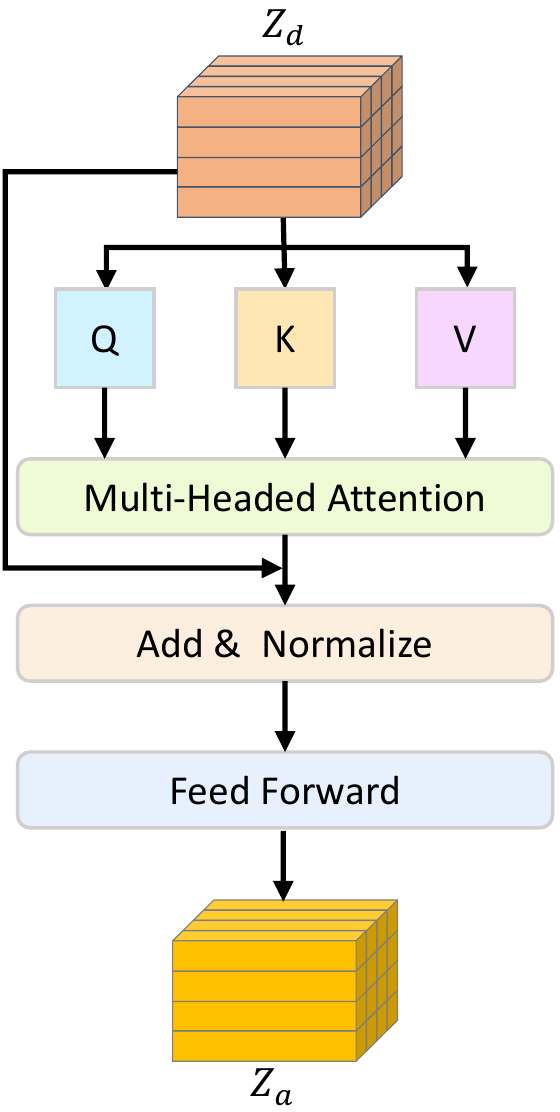} & \includegraphics[width=0.27\linewidth]{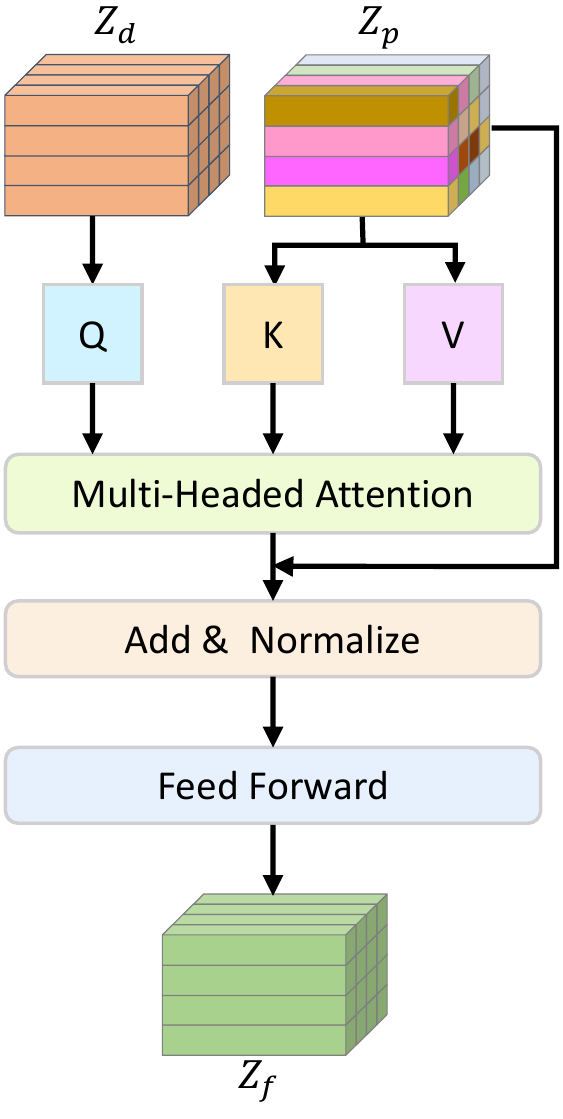} & \multicolumn{2}{c}{\includegraphics[width=0.37\linewidth]{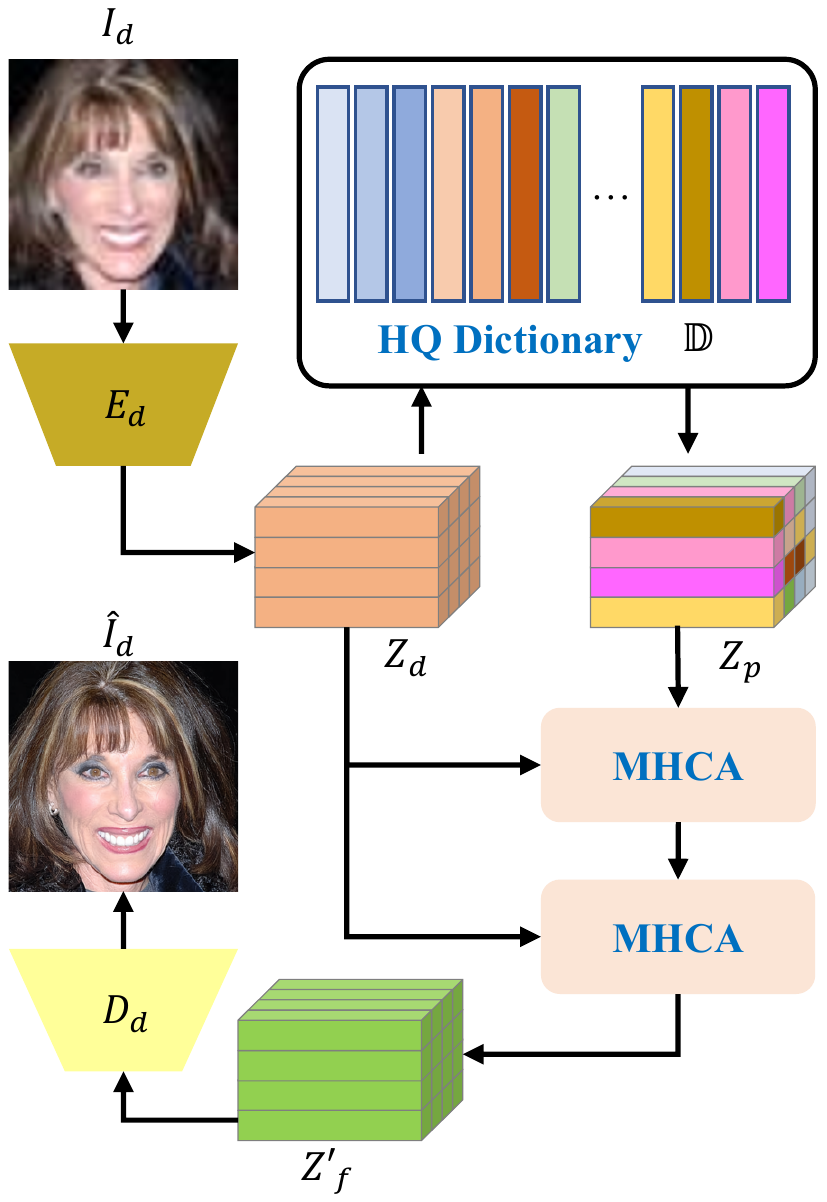}} \\
\small(a) \textbf{MHSA} & \small(b) \textbf{MHCA} & \multicolumn{2}{c}{\small(c) \textbf{RestoreFormer}} \\
% \multicolumn{2}{c}{(a) \textbf{MHSA-TF} and \textbf{MHCA-TF}} & \multicolumn{2}{c}{(b) \textbf{RestoreFormer}}
% \begin{tabular}{cc}
% \includegraphics[width=0.4\linewidth,scale=0.2]{figures/fig3_a2.pdf} & \includegraphics[width=0.4\linewidth]{figures/fig3_a1.pdf} \\
% \multicolumn{2}{c}{\includegraphics[width=0.92\linewidth]{figures/fig2_b.pdf}}
\end{tabular}
\end{center}
\vspace{-6mm}
\caption{\textbf{Framework of RestoreFormer}.
(a) MHSA is a transformer with multi-head self-attention used in most of previous ViTs~\cite{carion2020end,chen2021pre,dosovitskiy2020image,zhu2020deformable}. Its queries, keys, and values are from the degraded information $\bm{Z}_d$.
(b) MHCA is a transformer with a multi-head cross-attention used in the proposed RestoreFormer. 
It is designed to spatially fuse both degraded information $\bm{Z}_d$ and its corresponding high-quality priors $\bm{Z}_p$ by taking $\bm{Z}_d$ as queries while $\bm{Z}_p$ as key-value pairs. 
(c) is the whole pipeline of RestoreFormer. 
An encoder $\mathbf{E}_d$ is first deployed to extract representation $\bm{Z}_d$ of the degraded face $\bm{I}_d$ and its nearest high-quality priors $\bm{Z}_p$ are fetched from the HQ Dictionary $\mathbb{D}$.
% $\mathbb{D}$ by minimizing the distance between the priors in $\mathbb{D}$ and feature vectors in $\bm{Z}_d$.
%
Then two MHCAs are utilized to fuse the degraded features $\bm{Z}_d$ and priors $\bm{Z}_p$.
Finally, a decoder $\mathbf{D}_d$ is applied on the fused representation $\bm{Z}_f'$ to restore a high-quality face $\bm{\hat{I}}_d$. 
\textbf{The detailed structures of RestoreFormer are in the supplemental materials.}
% both the facial context and rich facial context and extra high-quality facial priors are helpful for blind face restoration. 
% %
% Therefore, the MHCA-TF is designed to take the degraded face information $\bm{Z}_d$ as queries while taking the facial priors $\bm{Z}_p$ as keys and values. 
%
% First, a degraded face $\mathbf{I_D}$ is being sent into an encode network $\mathbf{E_L}$ for attaining its feature $\mathbf{z_d}$.
% %
% Then, $\mathbf{z_d}$ spatially searches its most closed facial priors from an HQ Dictionary whose elements are high-quality facial information learning from plenty of undegraded faces and forms its corresponding priors $\mathbf{z_p}$.
% %
% Different from the traditional multi-head self-attention in previous ViTs, the transformer in our RestoreFormer takes the degraded features $\mathbf{z_d}$ as queries while regards the undegraded facial priors $\mathbf{z_p}$ as keys and values for learning fully spatial interactions between these two sources of information.
% %
% After two transformers, we attain fused facial feature $\mathbf{z_{f2}}$.
% %
% By sending $\mathbf{z_{f2}}$ into the face generator $\mathbf{D_L}$, we finally achieve a high-quality face.
}
\label{fig:framework}
\end{figure}

\vspace{5mm}
\noindent\textbf{MHSA.} 
As Figure~\ref{fig:framework}~(a) shown, MHSA used in most of the previous ViTs~\cite{carion2020end,chen2021pre,dosovitskiy2020image,zhu2020deformable} tends to globally attend contents from $\bm{Z}_d\in\mathbb{R}^{H'\times W'\times C}$ ($H',W'$ are spatial size of the feature map while $C$ is the number of channels) which is extracted from the degraded input in our task.
And the queries $\bm{Q}$, keys $\bm{K}$, and values $\bm{V}$ can be represented as:
%
%Followings are the formulas:
\begin{equation}
\small{\bm{Q}=\bm{Z}_d\bm{W}_{q}+\bm{b}_{q}\ ,\  
\bm{K}=\bm{Z}_d\bm{W}_{k}+\bm{b}_{k}\ ,\   
\bm{V}=\bm{Z}_d\bm{W}_{v}+\bm{b}_{v},}
\end{equation}
where $\bm{W}_{q/k/v}\in\mathbb{R}^{C \times C}$ and  $\bm{b}_{q/k/v}\in\mathbb{R}^{C}$ are learnable parameters.
% and $C_{out}$ is the number of channels of $\bm{Q}$, $\bm{K}$, and $\bm{V}$.

%
For getting more powerful representations, multi-head attention~\cite{vaswani2017attention} is adopted on $\bm{Q}$, $\bm{K}$, and $\bm{V}$.
First, $\bm{Q}$, $\bm{K}$, and $\bm{V}$ are separated into $N_h$ blocks along the channel dimension to obtain $\{\bm{Q}_1, \bm{Q}_2, \dots, \bm{Q}_{N_h}\}$, $\{\bm{K}_1, \bm{K}_2, \dots, \bm{K}_{N_h}\}$, and $\{\bm{V}_1, \bm{V}_2, \dots, \bm{V}_{N_h}\}$.
For each block, it has $C_h=\frac{C}{N_h}$ channels. 
Their attention maps can be represented as:
\begin{equation}
    \bm{Z}_{i} = \operatorname{softmax}(\frac{\bm{Q}_i\bm{K}_i^\intercal}{\sqrt{C_h}})\bm{V}_i, i=1,2,\dots,N_h
    \label{eq:softmax}
\end{equation}
and the final output of multi-head attention is the concatenation of $\bm{Z}_{i}$:
\begin{equation}
   \bm{Z}_{mh} = \operatornamewithlimits{concat}_{i=1,...,N_h}\bm{Z}_{i}.
    \label{eq:concat}
\end{equation}
%$
Similar to \cite{vaswani2017attention},
$\bm{Z}_{mh}$ is regarded as residual.
$\bm{Z}_{mh}$ and $\bm{Z}_d$ are added before sending the summation into a normalization layer and a feed forward network sequentially as:
\begin{equation}
    \bm{Z}_{a} = \operatorname{FFN}(\operatorname{LN}(\bm{Z}_{mh} + \bm{Z}_d)),
    \label{eq:shortcut}
\end{equation}
where $\operatorname{LN}$ is the layer normalization, $\operatorname{FFN}$ is the feed-forward network composed by two convolution layers, and $\bm{Z}_{a}$ is the finally globally attended feature map.

\noindent\textbf{MHCA.} 
Different from MHSA, our MHCA aims for spatially fusing the information from the degraded face and its corresponding priors that can respectively provide identity information and high-quality facial details for face restoration.
Therefore, as Figure~\ref{fig:framework} (b) shown, our MHCA takes features $\bm{Z}_d$ from the degraded face, as queries $\bm{Q}$, while the keys $\bm{K}$ and values $\bm{V}$ are from its high-quality facial priors $\bm{Z}_p\in\mathbb{R}^{H'\times W'\times C}$:
\begin{equation}
\small{
\bm{Q}=\bm{Z}_d\bm{W}_{q}+\bm{b}_{q}\ ,\ 
\bm{K}=\bm{Z}_p\bm{W}_{k}+\bm{b}_{k}\ ,\ 
\bm{V}=\bm{Z}_p\bm{W}_{v}+\bm{b}_{v},}
\end{equation}
Following multi-head attention in MHSA according to Eq.~\ref{eq:softmax} and Eq.~\ref{eq:concat}, $\bm{Z}_{mh}$ in MHCA can be estimated similarly.
To generate features with more face details, $\bm{Z}_{mh}$ is added by $\bm{Z}_{p}$ before LN and FNN to get the final fused features $\bm{Z}_{f}$:
%However, since in blind face restoration, high-quality facial priors play a more critical role in the final restored face, the short-cut happens between $\bm{Z}_{mh}$ and $\bm{Z}_p$, rather than between $\bm{Z}_{mh}$ and $\bm{Z}_d$.
%
%Therefore, the final fused output $\bm{Z}_f$ is formulated as:
\begin{equation}
    \bm{Z}_{f} = \operatorname{MHCA}(\bm{Z}_{d}, \bm{Z}_p)=\operatorname{FFN}(\operatorname{LN}(\bm{Z}_{mh} + \bm{Z}_p)).
    \label{eq:shortcut}
\end{equation}

\noindent\textbf{RestoreFormer.} 
The whole pipeline of the proposed RestoreFormer based on MHCA is shown in Figure~\ref{fig:framework}~(c).
First, a degraded image $\bm{I}_d$ is sent into an image encoder $\mathbf{E}_d$, which is composed of 12 residual blocks and 5 average poolings, to extract representations $\bm{Z}_d$.
Then we fetch priors from a reconstruction-oriented HQ Dictionary $\mathbb{D}=\{\bm{d}_m\}_{m=1}^{M} (d_m\in\mathbb{R}^{C}$). $\mathbb{D}$ consists of $M$ high-quality facial priors and the learning of the HQ Dictionary will be explained in Sec.~\ref{subsec: HQ dict}.
By finding the most similar priors of feature vectors in $\bm{Z}_d$ from $\mathbb{D}$, we get the priors $\bm{Z}_p$:
%
% The above procedure can be formulated as:
\begin{equation}
% \bm{Z}_d = \operatorname{f_{E_d}}(\bm{I}_d)\ ;\ 
\bm{Z}_p^{(i,j)} = \mathop{\arg\min}_{\bm{d}_m \in \mathbb{D}} \|\bm{Z}_d^{(i,j)}-\bm{d}_m\|_2^2,
\end{equation}
where $\bm{Z}_p^{(i,j)}$ and $\bm{Z}_d^{(i,j)}$ indicate the feature vector on the location $(i,j)$ of $\bm{Z}_q$ and $\bm{Z}_d$, respectively.
$||\cdot||_2$ is the L2-norm. 

Given $\bm{Z}_p$ and $\bm{Z}_d$, two consecutive MHCAs are applied and we can get a refined representation $\bm{Z'}_{f}$ as:
\begin{equation}
\bm{Z'}_f = \operatorname{MHCA}(\bm{Z}_d, \operatorname{MHCA}(\bm{Z}_d, \bm{Z}_p)).
\end{equation}

Finally, $\bm{Z'}_{f}$ is fed into a decoder $\mathbf{D}_d$ with 12 residual blocks and 5 nearest neighbour upsampling to recover the high-quality image $\bm{\hat{I}}_d\in\mathbb{R}^{H\times W\times 3}$.% as:
% \begin{equation}
%   \bm{\hat{I}}_d = \operatorname{f_{D_d}}(\bm{Z'}_f).
% \end{equation}

\begin{figure}
\begin{center}
\begin{tabular}{cc}
\includegraphics[width=0.4\linewidth]{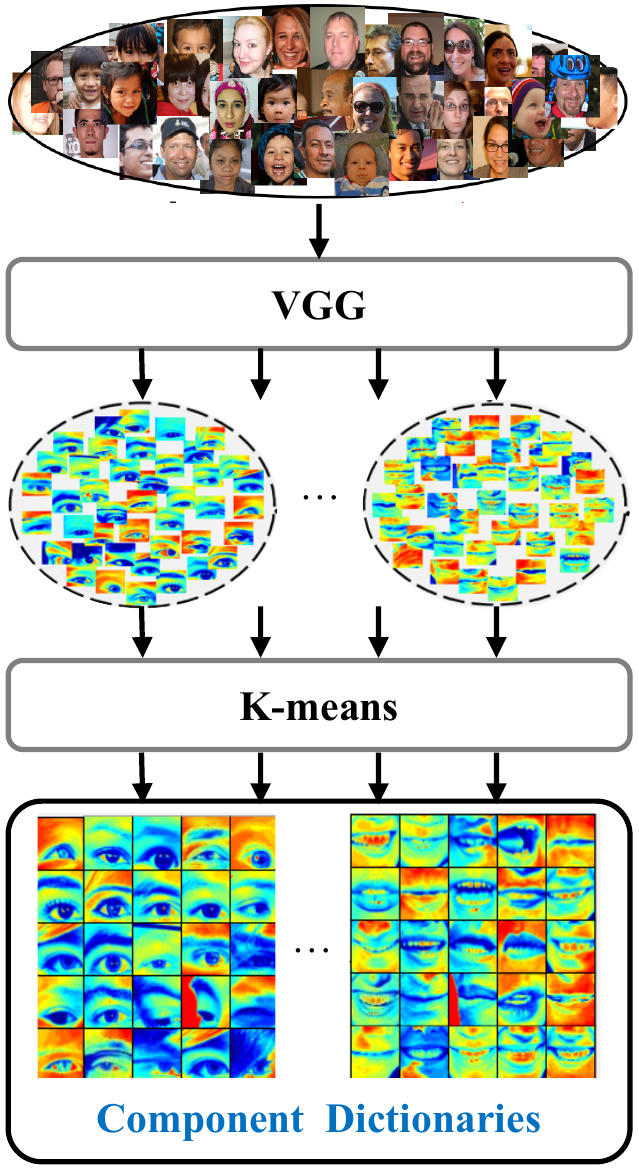} & \includegraphics[width=0.4\linewidth]{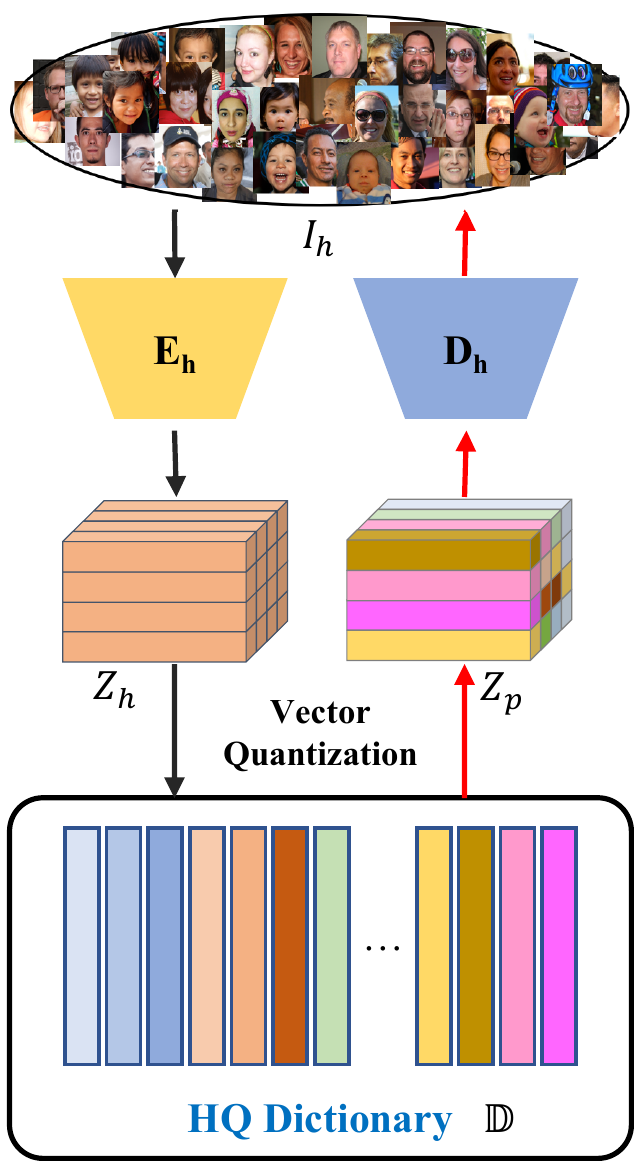} \\
(a) Component Dictionaries & (b) HQ Dictionary
\end{tabular}
\end{center}
\vspace{-0.6cm}
\caption{\textbf{Comparison of Prior Dictionary}.
(a) Component Dictionaries, proposed in DFDNet~\cite{li2020blind}, are off-line generated by a VGG network~\cite{simonyan2014very} and clustered with K-means. They only consider eyes, nose, and mouth.
%
% It utilizes a well-trained VGG network~\cite{simonyan2014very} to extract face features from plenty of high-quality faces. 
%
% It only focuses on certain parts of these facial features, such as eyes, nose, and mouth. 
%
% It adopts K-means to generate K clusters for each component and considers them as the elements of the component dictionary. 
%
(b) HQ Dictionary, proposed in this paper, is learned by a high-quality face generation network incorporating the idea of vector quantization~\cite{oord2017neural}. 
The high-quality priors in the HQ Dictionary are reconstruction-oriented and provide more facial details for the restoration of degraded faces.
Besides, the priors in the HQ Dictionary involve all the facial regions.
%
% It also attains high-quality facial priors from amounts of undegraded faces.
%
% However, instead of off-line generation, its elements are learnt by a high-quality face generation network incorporating the idea of vector quantization~\cite{oord2017neural}.
%
% Our HQ Dictionary has two appealing benefits compared to the Component Dictionary.
% %
% (1) Since the elements in HQ Dictionary are distilling from a network that is towards high-quality face reconstruction, they own rich and diverse details for the restoration of degraded faces. 
% %
% On the other side, VGG network is initially trained for recognition where facial details will not take much attention. %
% %
% Therefore, by contrast, the Component Dictionary cannot provide more details for face restoration.
% %
% (2) Since the Component Dictionary only focuses on certain facial parts while our HQ Dictionary involves all the facial parts, our HQ Dictionary can restore a better face in the whole view.
}
\label{fig:dictionary}
\end{figure}

\vspace{5mm}
\noindent\textbf{Learning.}
To train RestoreFormer, our losses involve several aspects, including pixel-level, component-level, and image-level.
Following are the detailed discussions.

\noindent\textit{Pixel-level losses.}
In pixel-level, we adopt two widely-used losses for face restoration: $L1$ loss and perceptual loss~\cite{johnson2016perceptual,ledig2017photo}.
They are expressed as:
\begin{equation}
   \mathcal{L}_{l1} = |\bm{I_h} - \bm{\hat{I}_d}|_1\ ;\ \mathcal{L}_{per} = \|\phi(\bm{I}_h) - \phi(\bm{\hat{I}}_d)\|_2^2
\end{equation}
where $\bm{I_h}$ is the ground truth high-quality image;
$\phi$ is the pretrained VGG-19~\cite{simonyan2014very} and the feature maps are extracted from $\lbrace conv1, \dots, conv5 \rbrace$.

Besides, for accurately matching high-quality priors from the HQ Dictionary, we force the extracted features $\bm{Z}_d$ to approach their selected priors $\bm{Z}_p$.
That is:
\begin{equation}
   \mathcal{L}_{p} = \|\bm{Z}_p - \bm{Z}_d\|_2^2.
\end{equation}

\noindent\textit{Component-level losses.}
Since eyes and mouth play an important role in the overview of a face, we also adopt a discrimination loss and feature style loss on the facial areas of eyes and mouth for further enhancing their restored quality.
Following~\cite{wang2021towards}, we only focus on regions $r\in\{$left eye, right eye, mouth$\}$ and the loss functions are formulated as:
% \small\small
\begin{equation}
\small{
% \begin{aligned}
\begin{split}
\mathcal{L}_{disc} &= \sum_{r}[\log D_r(R_{r}(\bm{I_h})) + \log (1 - D_r(R_{r}(\bm{\hat{I}}_d)))], \\
\mathcal{L}_{style} &= \sum_{r}\|\operatorname{Gram}(\varphi(R_{r}(\bm{I_h})) - \operatorname{Gram}(\varphi(R_{r}(\bm{\hat{I}}_d))\|_2^2,
\end{split}
% \end{aligned}
}
\end{equation}
where $R_{r}(\cdot)$ is ROI align~\cite{he2017mask} and 
$\varphi$ denotes the multi-resolution features of discriminator $D_r$ trained on region $r$.
$\operatorname{Gram}$ denotes the Gram matrix~\cite{gatys2016image} that calculates the feature correlations to measure the style difference.

\noindent\textit{Image-level losses.}
The proposed method aims for attaining faces with high realness and fidelity.
Therefore, in image-level, we adopt an adversarial loss for improving the realness of the restored face and an identity loss~\cite{wang2021towards} for keeping its fidelity as:
\begin{equation}
\begin{split}
\mathcal{L}_{adv} &= [\log D(\bm{I}_h) + \log (1-D(\bm{\hat{I}}_d))], \\
\mathcal{L}_{id} &= \|\eta(\bm{I_h}) - \eta(\bm{\hat{I}_d})\|_2^2,
\end{split}
\label{eq: per adv}
\end{equation}
where $D$ is the discriminator trained on the face image and $\eta$ denotes the identity feature extracted from a well-trained face recognition ArcFace~\cite{deng2019arcface} model.

In the end, with all the loss functions proposed above, the final loss to train RestoreFormer is:
\begin{equation}
\begin{aligned}
\mathcal{L}_{RF} &= \mathcal{L}_{l1} + \lambda_{per} \mathcal{L}_{per} + \lambda_{p} \mathcal{L}_{p} + \lambda_{disc} \mathcal{L}_{disc} \\
&+ \lambda_{style} \mathcal{L}_{style} + \lambda_{adv} \mathcal{L}_{adv} + \lambda_{id} \mathcal{L}_{id},\\
\end{aligned}
\end{equation}
where $\lambda_{\dots}$ are the weighting factors for different losses.

\subsection{HQ Dictionary}
\label{subsec: HQ dict}
In this subsection, we introduce the generation of the HQ Dictionary $\mathbb{D}=\{\bm{d}_m\}_{m=0}^{M}, d_m\in\mathbb{R}^{C}$ used in RestoreFormer.

As shown in Figure~\ref{fig:dictionary}, different from \cite{li2020blind} whose component dictionaries are generated from an off-line recognition-orientated feature extractor VGG~\cite{simonyan2014very}, we aim for getting a reconstruction-oriented high-quality dictionary that can provide richer facial details for face restoration.
Therefore, we deploy a high-quality face generation network motivated from vector quantization~\cite{oord2017neural} to learn a high-quality dictionary $\mathbb{D}$ from plenty of undegraded faces.

The framework of this face generation network is shown in Figure~\ref{fig:dictionary} (b).
First, an encoder $\mathbf{E}_h$ is used to extract the representation $\bm{Z}_h\in \mathbb{R}^{H' \times W' \times C}$ from a high-quality undegraded image $\bm{I}_h\in \mathbb{R}^{H \times W \times 3}$. %:
%we sample a high-quality image $\bm{I}_h\in \mathbb{R}^{H \times W \times 3}$ from plenty of high-quality images and use an encoder $\mathbf{E}_h$ to extract the representation $\bm{Z}_h\in \mathbb{R}^{H' \times W' \times C}$:
%
% \begin{equation}
% \bm{Z}_h = \operatorname{f_{E_h}}(\hm{I}_h).
% \end{equation}
%
Then rather than decoding $\bm{Z}_h$ with a decoder $\mathbf{D}_h$ directly, we quantize feature vectors of $\bm{Z}_h$ by their nearest elements in $\mathbb{D}$ and finally get $\bm{Z}_{p}\in \mathbb{R}^{H' \times W' \times C}$:
\begin{equation}
\bm{Z}_p^{(i,j)} = \mathop{\arg\min}_{\bm{d}_m \in \mathbb{D}} \|\bm{Z}_h^{(i,j)}-\bm{d}_m\|_2^2,
\label{eq:quantization}
\end{equation}
where $\bm{Z}_p^{(i,j)}$ and $\bm{Z}_h^{(i,j)}$ are the feature vectors on the position $(i,j)$ of $\bm{Z}_p$ and $\bm{Z}_h$, respectively. 
Taking $\bm{Z}_p$ as input, the decoder $\mathbf{D}_h$ can reconstruct a high-quality face $\bm{\hat{I}}_h \in \mathbb{R}^{H \times W \times 3}$. % as:
%
\iffalse
\begin{eqnarray}
\bm{\hat{I}_h} = \operatorname{f_{D_h}}(\bm{Z}_p).
\end{eqnarray}
\fi
%
Noted that the structures of $\mathbf{E}_h$ and $\mathbf{D}_h$ are the same with that of $\mathbf{E}_d$ and $\mathbf{D}_d$ in Sec.~\ref{subsec: RestoreFormer}.

\vspace{5mm}
\noindent\textbf{Learning.}
The elements $\bm{d}_m$ in $\mathbb{D}$ are randomly initialized by a uniform distribution.
For updating them to capture high-quality facial information, we adopt a dictionary learning algorithm, Vector Quantization (VQ)~\cite{oord2017neural}, to move $\bm{Z}_p$ towards $\bm{Z}_h$ as:
\begin{equation}
   \mathcal{L'}_{d} = \|\operatorname{sg}[\bm{Z}_h] - \bm{Z}_p\|_2^2
\end{equation}
where $sg[\cdot]$ denotes the stop-gradient operation.
Noted that since $\bm{Z}_p$ consists of the elements in $\mathbb{D}$ according to Eq.~\ref{eq:quantization}, $\mathbb{D}$ is updated through $\bm{Z}_p$.
To keep the encoder $\mathbf{E}_h$ and dictionary $\mathbb{D}$ in the same learning space, a commitment loss~\cite{oord2017neural} is also adopted:
\begin{equation}
   \mathcal{L'}_{c} = \|\bm{Z}_h - \operatorname{sg}[\bm{Z}_p]\|_2^2.
\end{equation}
%
% \textcolor{red}{
As the above two losses make $\bm{Z}_p$ close to $\bm{Z}_h$ which is extracted from high-quality undegraded image $\hm{I}_h$, $\bm{Z}_p$ contains facial detail information which can benefit face restoration.
And we consider $\mathbb{D}=\{\bm{d}_m\}_{m=0}^{M}, d_m\in\mathbb{R}^{C}$ as facial prior in RestoreFormer.
% }

%
Besides the two losses for dictionary, an $L1$ loss, a perceptual loss, and an adversarial loss are also applied to the final reconstructed result $\bm{\hat{I}}_h$ to make sure $\bm{Z}_p$ has sufficient information to restore high-quality image $\bm{I_h}$:
\begin{equation}
\begin{aligned}
\mathcal{L'}_{l1} &= \|\bm{I_h} - \bm{\hat{I}_h}\|_1\ ;\  \mathcal{L'}_{per} = \|\phi(\bm{I_h} ) - \phi(\bm{\hat{I}_h})\|_2^2 \\
\mathcal{L'}_{adv} &= [\log D(\bm{I_h}) + \log (1-D(\bm{\hat{I}_h}))].
\end{aligned}
\end{equation}
Noted that, since Eq.~\ref{eq:quantization} is non-differentiable, the gradient of $\bm{Z}_h$ is simply copied from $\bm{Z}_p$~\cite{oord2017neural}.

The final loss is:
\begin{equation}
\mathcal{L}_{Dict} = \mathcal{L'}_{l1} + \lambda_{per}\mathcal{L'}_{per} + \lambda_{adv}\mathcal{L'}_{adv} + \lambda_d\mathcal{L'}_{d} + \lambda_c \mathcal{L'}_{c},
\end{equation}
where $\lambda'_{\dots}$ are the weighting factors.

\section{Experiments and Analysis}

% \begin{itemize}
%	\item[-] Dataset: GoPro and \textbf{Blur-DVS} 
%	\item[-] Experimental Setting
%	\item[-] Analysis:
%	\begin{itemize}
%		\item[.] Compare with State-of-the-art Method ~\ref{Tab:SOFA}
%		\item[.] Ablusion Study (joint train? the number of interations? displace PhysicalDeblur with SRN? kpn? )
%		\item[.] Visualization of KPN
%	\end{itemize}                                                      
% \end{itemize}

\subsection{Datasets}
\noindent\textbf{Training Datasets.}
The HQ Dictionary is trained on the FFHQ~\cite{karras2019style} dataset. 
It contains 70000 high-quality images and all are resized to $512 \times 512$.
%
%The proposed RestoreFormer is also trained on FFHQ dataset with the degrading model proposed in ~\cite{li2020enhanced,li2018learning,wang2021towards}:
Since the proposed RestoreFormer needs degarded image and high-quality image pairs for training, we synthesize degraded images on FFHQ dataset by the degrading model proposed in ~\cite{li2020enhanced,li2018learning,wang2021towards}:
\begin{equation}
\mathbf{I_d} = \{[(\mathbf{I_h} \otimes \mathbf{k}_\sigma) \downarrow_r + \mathbf{n}_\delta]_{{JPEG}_q}\}\uparrow_r.
\end{equation}
Specifically, a high-quality image $\mathbf{I_h}$ is firstly blurred by Gaussian blur kernel $\mathbf{k}_\sigma$ whose sigma is $\sigma$.
Then, it will be bilinearly downsampled with a scale factor $r$ and added with white Gaussian noise $\mathbf{n}_\delta$ with sigma $\delta$.
Finally, a JPEG compression with quality factor $q$ will be adopted to generate the final degraded image.% $\mathbf{I_d}$.
And it will be resized to the same size of $\mathbf{I_h}$ by bilinear upsampling as the degraded input $\mathbf{I_d}$ of our network similar to existing arts~\cite{li2020enhanced,li2018learning,wang2021towards}.
In this paper, $\sigma$, $r$, $\delta$, and $q$ are randomly sampled from $\lbrace0.2:10\rbrace$, $\lbrace1:8\rbrace$, $\lbrace0:20\rbrace$, and $\lbrace60:100\rbrace$, respectively.

\noindent\textbf{Testing Datasets.}
We evaluate our method on a synthetic dataset: CelebA-Test and three real-world datasets: LFW-Test, CelebChild-Test, and WebPhoto-Test.
CelebA-Test consists of 3000 images and it is synthesized by applying the degrading described above on the testing set of CelebA-HQ images~\cite{liu2015deep}. 
For LFW-Test, it consists of the first image of each identity in the validation partition of the original LFW~\cite{huang2008labeled} and there are 1711 images.
Another two real-world datasets are collected by Wang~\etal~\cite{wang2021towards} from the Internet.
Specifically, CelebChild-Test contains 180 child faces of celebrities and WebPhoto-Test consists of 407 real life faces.

\subsection{Experimental Settings and Metrics}
\noindent\textbf{Settings.}
The size of the input image is $512 \times 512 \times 3 $ and the size of $\bm{Z}_d$ is $16\times16\times256$.
The HQ Dictionary contains $M=1024$ elements and the length of each element is 256.
% and the length of the embedding codes $m=1024$.
%
The batch size is $16$ and the weighting factors of the loss function are $\lambda_{per}=1.0$, $\lambda_{p}=0.25$, $\lambda_{disc}=1.0$, $\lambda_{style}=2000$, $\lambda_{adv}=0.8$, $\lambda_{id}=1.5$, $\lambda_{d}=1.0,$ and $\lambda_{c}=0.25$.
%
% Besides, $\beta$ in the quantized loss function is 0.25.
%

During training, HQ Dictionary is trained by Adam optimizer~\cite{kingma2014adam} and the learning rate is set to $7e^{-5}$ at the beginning.
Then, the learning rate is decayed by 10 after $6e^{5}$ iterations.
%
%In the first 30000 iterations, the discriminators are disabled and at the 600000 iterations, the learning rate is decayed by 10.
%
The dictionary is trained until $8e^{5}$ iterations.
%We stop the training when it reaches 800000 iterations.
%
We also optimize the RestoreFormer with Adam.
Since $\mathbf{E}_d$ and $\mathbf{D}_d$ in RestoreFormer are initialized by $\mathbf{E}_h$ and $\mathbf{D}_h$ for dictionary learning, the learning rate of RestoreFormer is set to $7e^{-6}$ and trained by $6e^{4}$ iterations.
%Since we respectively pretrain $\mathbf{E}_d$ and $\mathbf{D}_d$ with the final parameters of $\mathbf{E}_h$ and $\mathbf{D}_h$, the learning rate of RestoreFormer is set to $7e^{-6}$ and the whole training lasts 60000 iterations.

\noindent\textbf{Metrics.}
Our evaluation is based on both the realness and fidelity of the restored faces.
To measure the realness, except a widely-used non-reference metric FID~\cite{heusel2017gans}, we also deploy a user study for further evaluating the visual performance of the restored results from the perspective of humans.
As for the facial fidelity, we adopt two pixel-wise metrics: PSNR and SSIM and a perceptual metric: LPIPS~\cite{zhang2018unreasonable}.
Since identity recognition is a more straight and convincing approach for evaluating the fidelity of faces, we introduce an identity distance (denoted as IDD) that is implemented by measuring the distance of the features extracted from ArcFace~\cite{deng2019arcface} with angle.
%
%The fidelity-related metrics can only be applied to the CelebA-Test dataset which requires ground truth (GT), while the realness-related metrics can be used in both the synthetic and real-world datasets.
%
% Besides, excepting PSNR and SSIM, which are the larger the better, the other metrics are the smaller the better.

\renewcommand{\tabcolsep}{.5pt}
\begin{figure*}[t]
\begin{center}
\begin{tabular}{ccccccc}
   \vspace{-1.0mm}
   \includegraphics[width=\swseven]{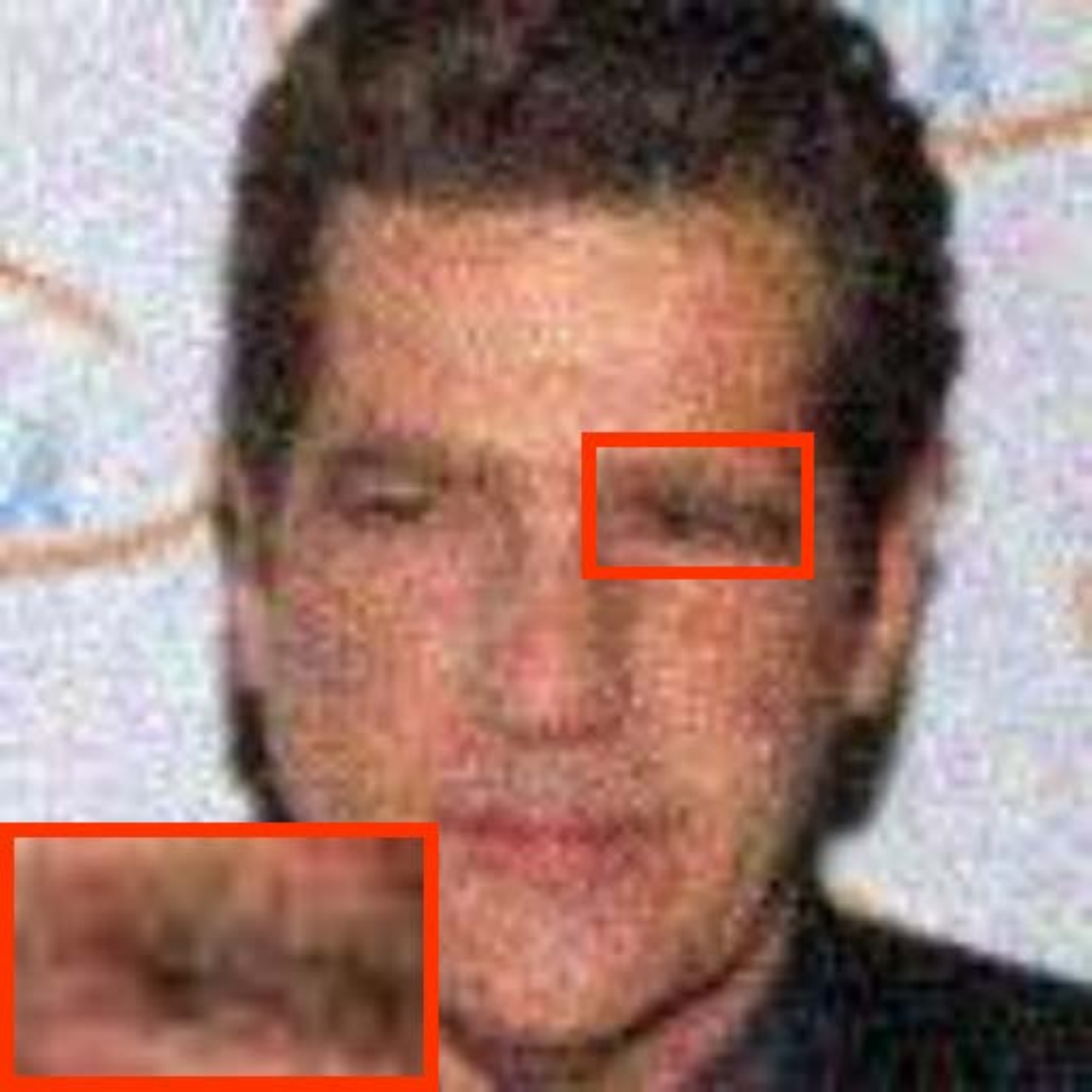}&
   \includegraphics[width=\swseven]{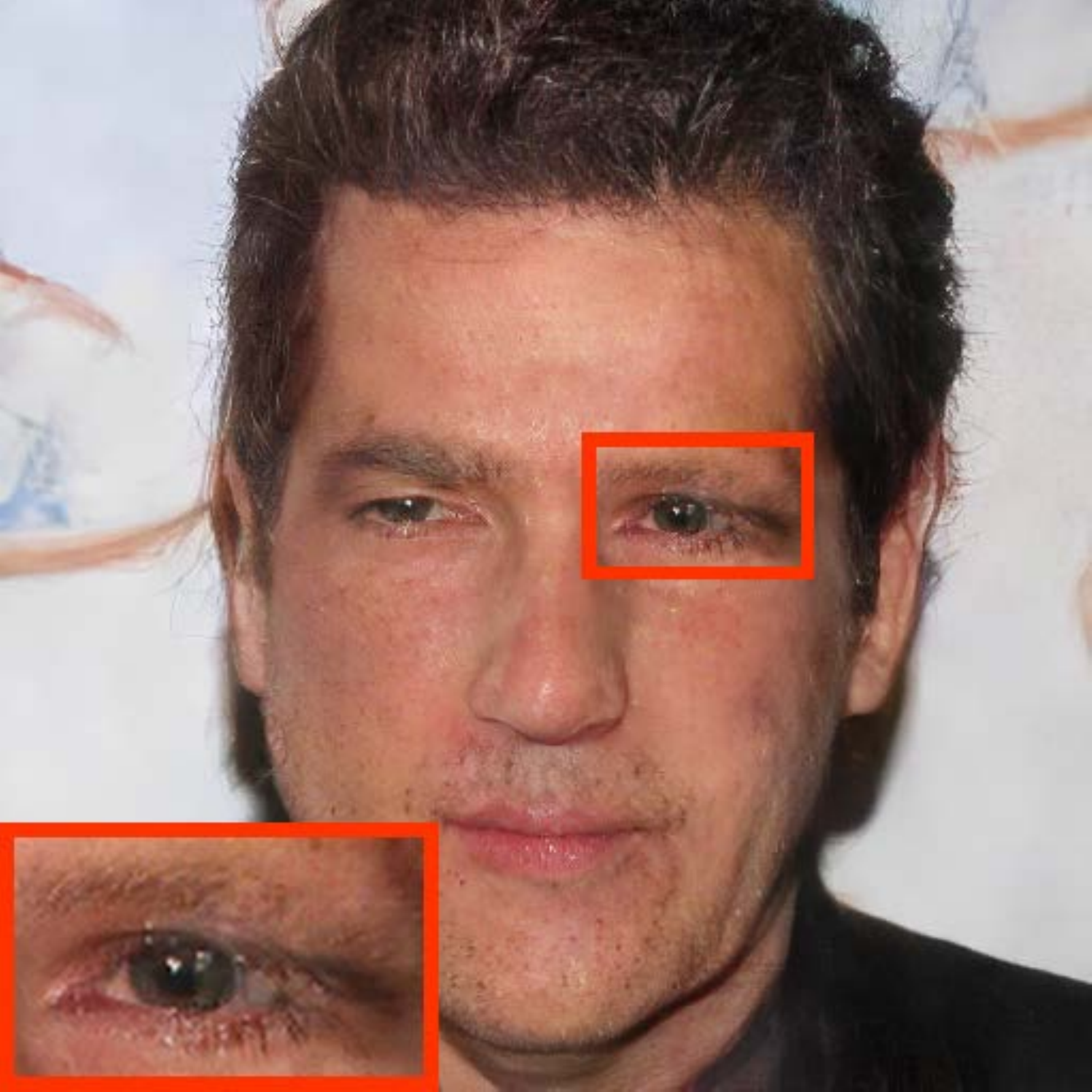}&
   \includegraphics[width=\swseven]{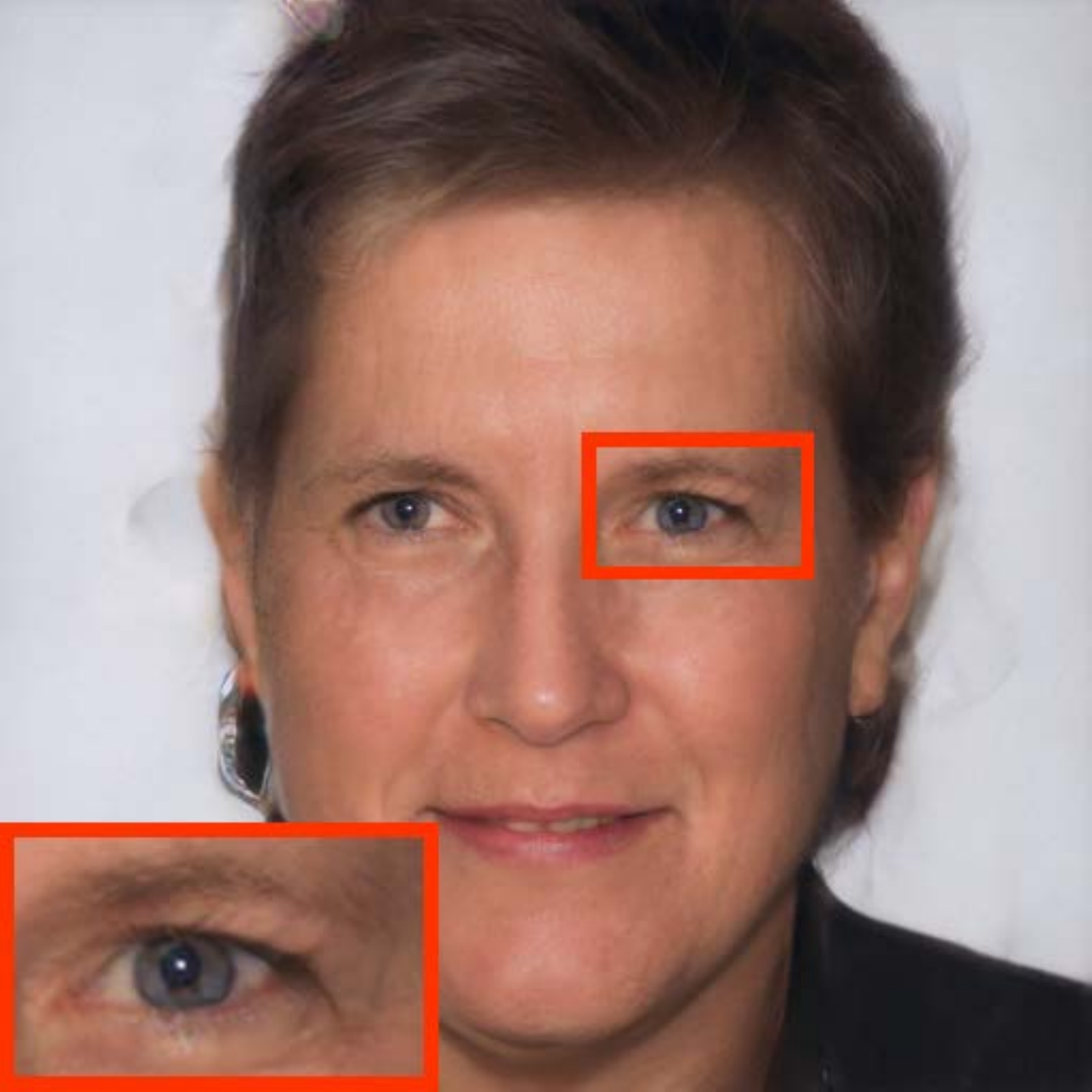}&
   \includegraphics[width=\swseven]{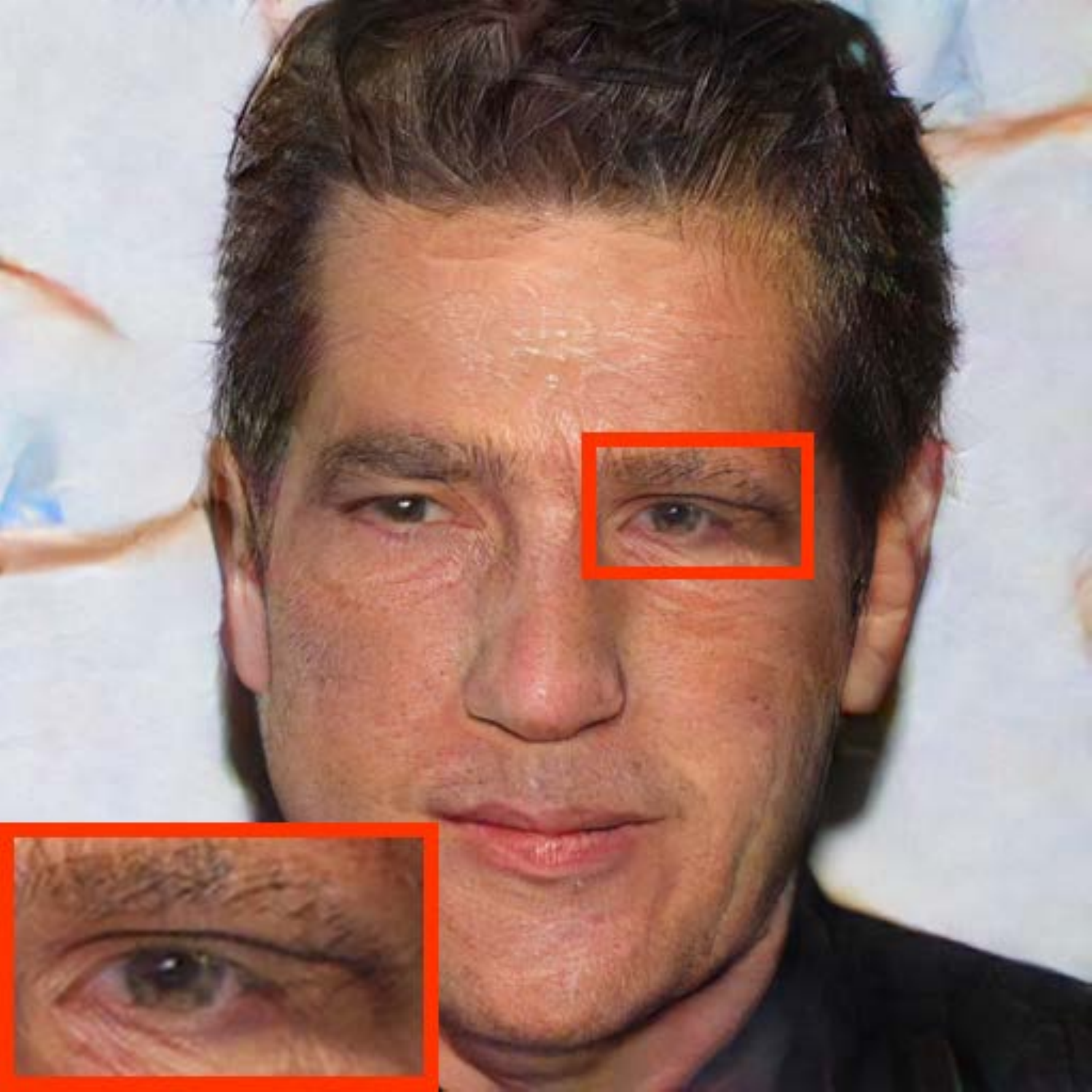}&
   \includegraphics[width=\swseven]{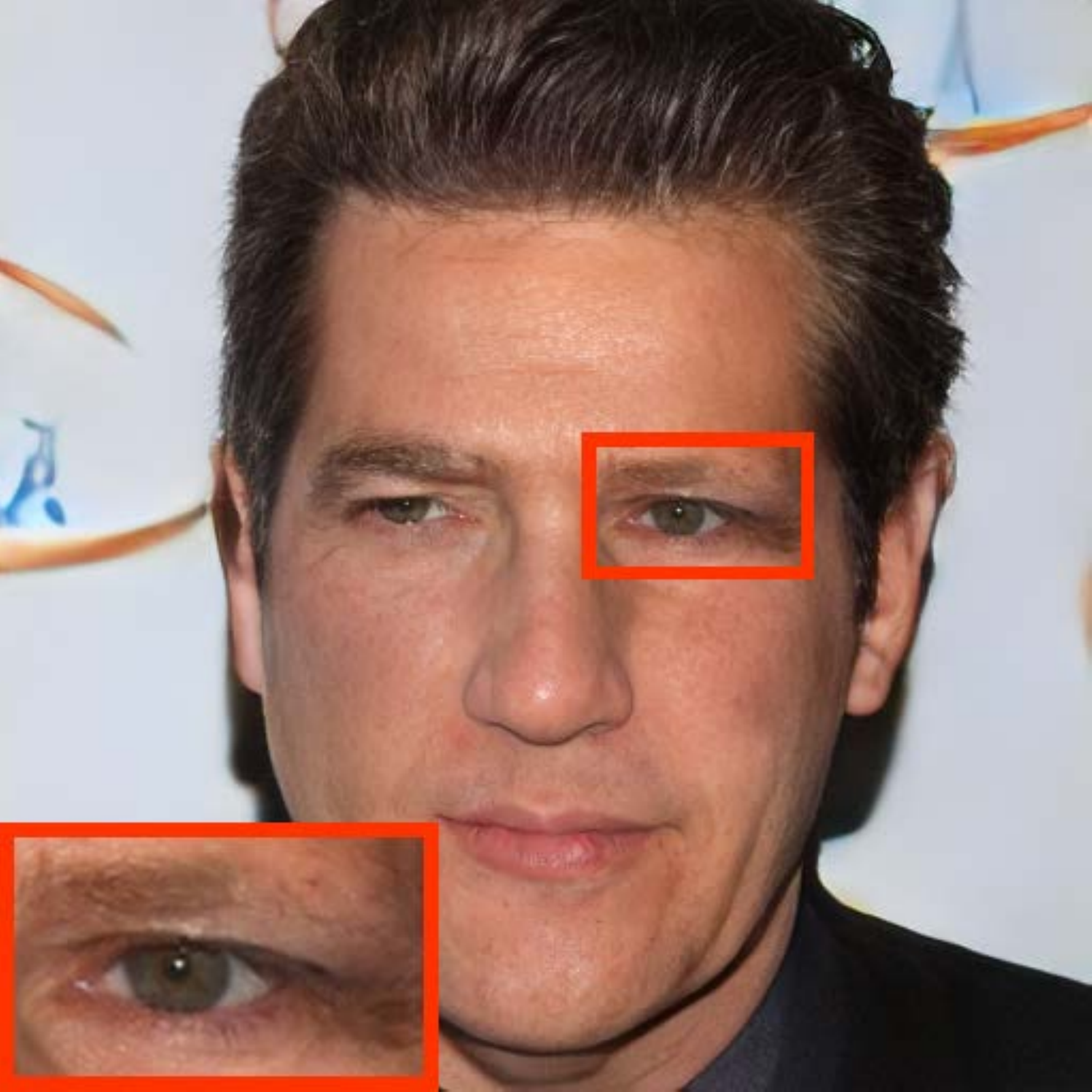}&
   \includegraphics[width=\swseven]{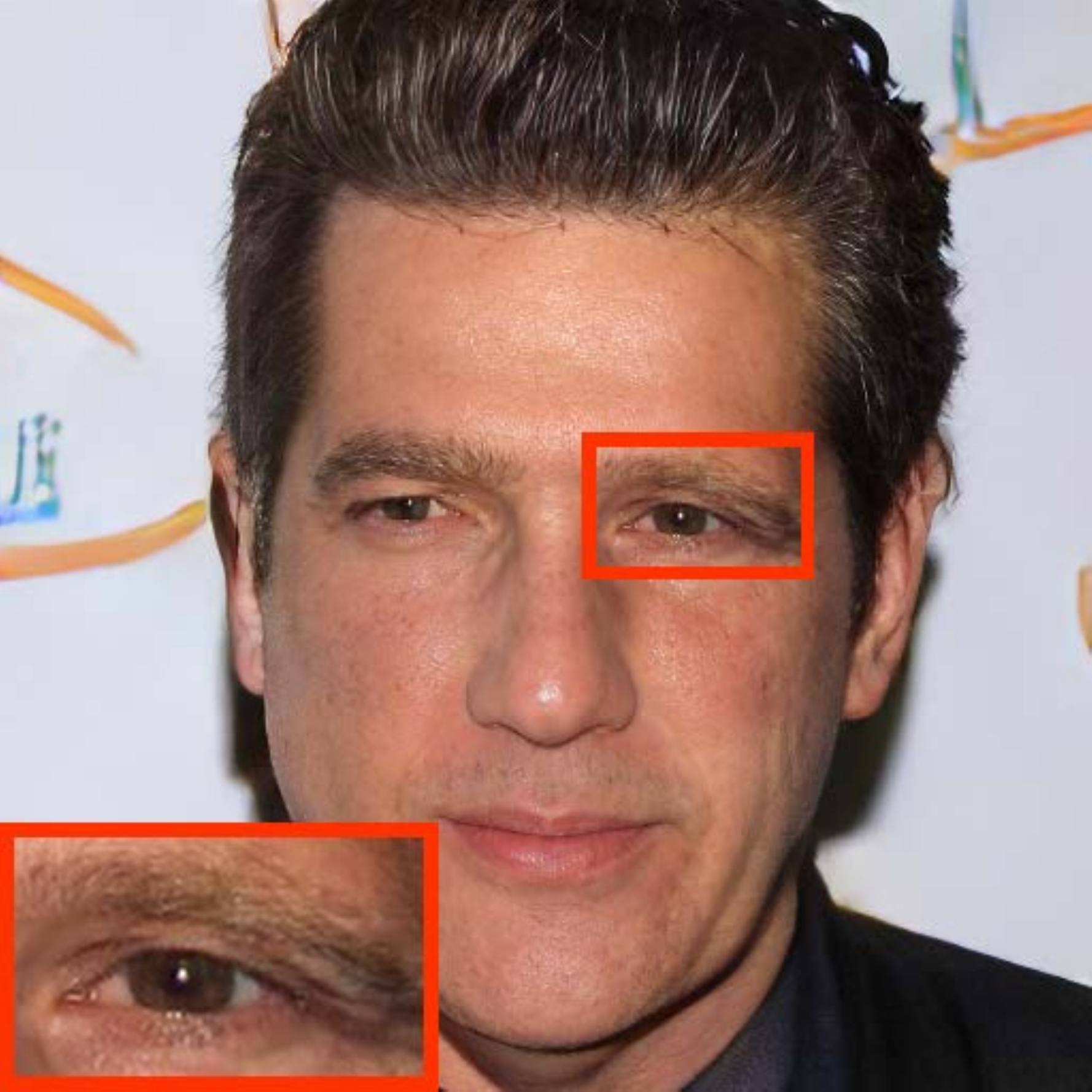} &
   \includegraphics[width=\swseven]{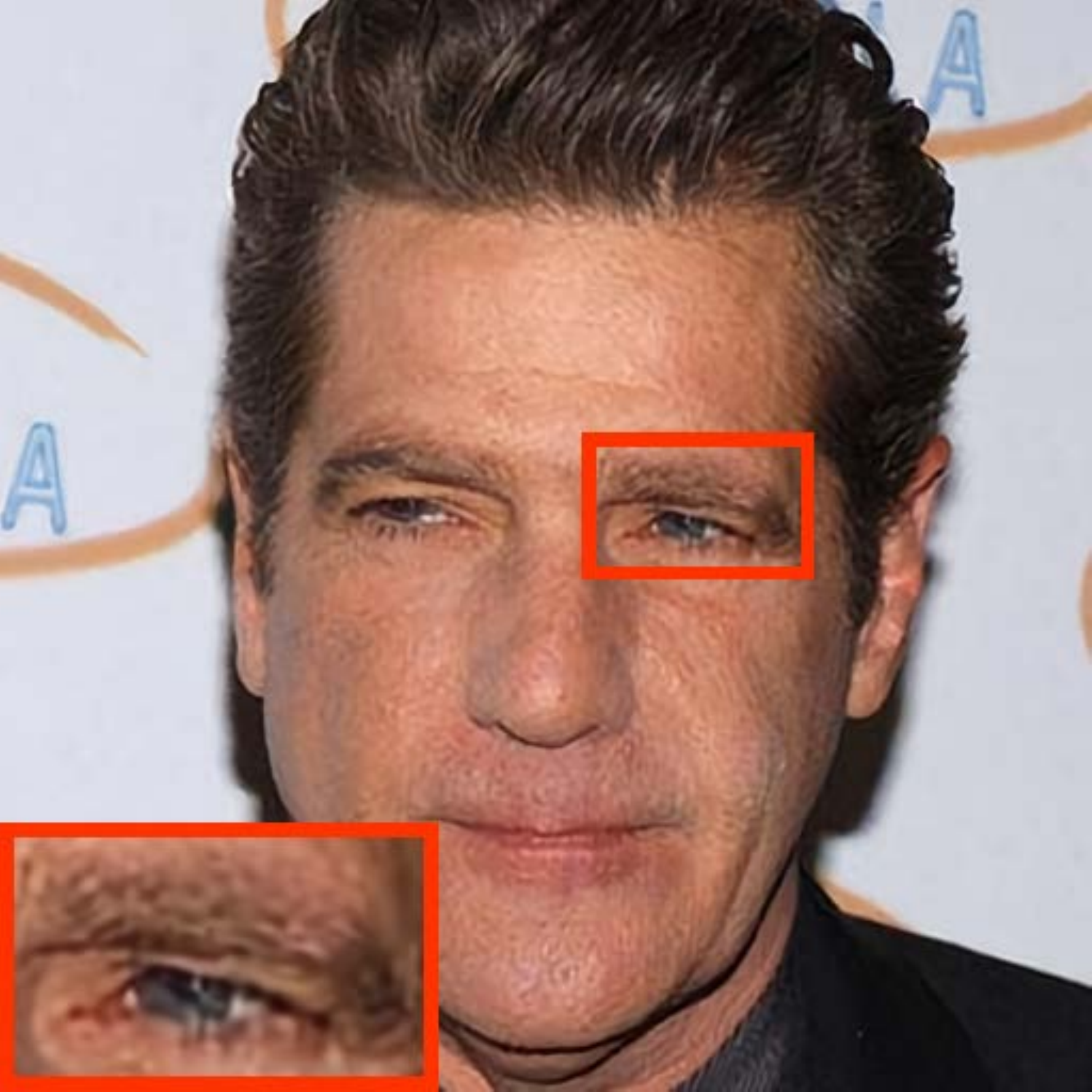} \\
   % \vspace{-1.0mm}
   % \includegraphics[width=\sweight]{figures/exp/celeba1/059_validation_1052_org.pdf}&
   % \includegraphics[width=\sweight]{figures/exp/celeba1/059_validation_1052_dfd.pdf}&
   % \includegraphics[width=\sweight]{figures/exp/celeba1/059_validation_1052_bop.pdf}&
   % \includegraphics[width=\sweight]{figures/exp/celeba1/059_validation_1052_pulse.pdf}&
   % \includegraphics[width=\sweight]{figures/exp/celeba1/059_validation_1052_pfsrgan.pdf}&
   % \includegraphics[width=\sweight]{figures/exp/celeba1/059_validation_1052_gfp.pdf}&
   % \includegraphics[width=\sweight]{figures/exp/celeba1/059_validation_1052_ours.pdf} &
   % \includegraphics[width=\sweight]{figures/exp/celeba1/059_validation_1052_gt.pdf} \\
   \vspace{-1.0mm}
   \includegraphics[width=\swseven]{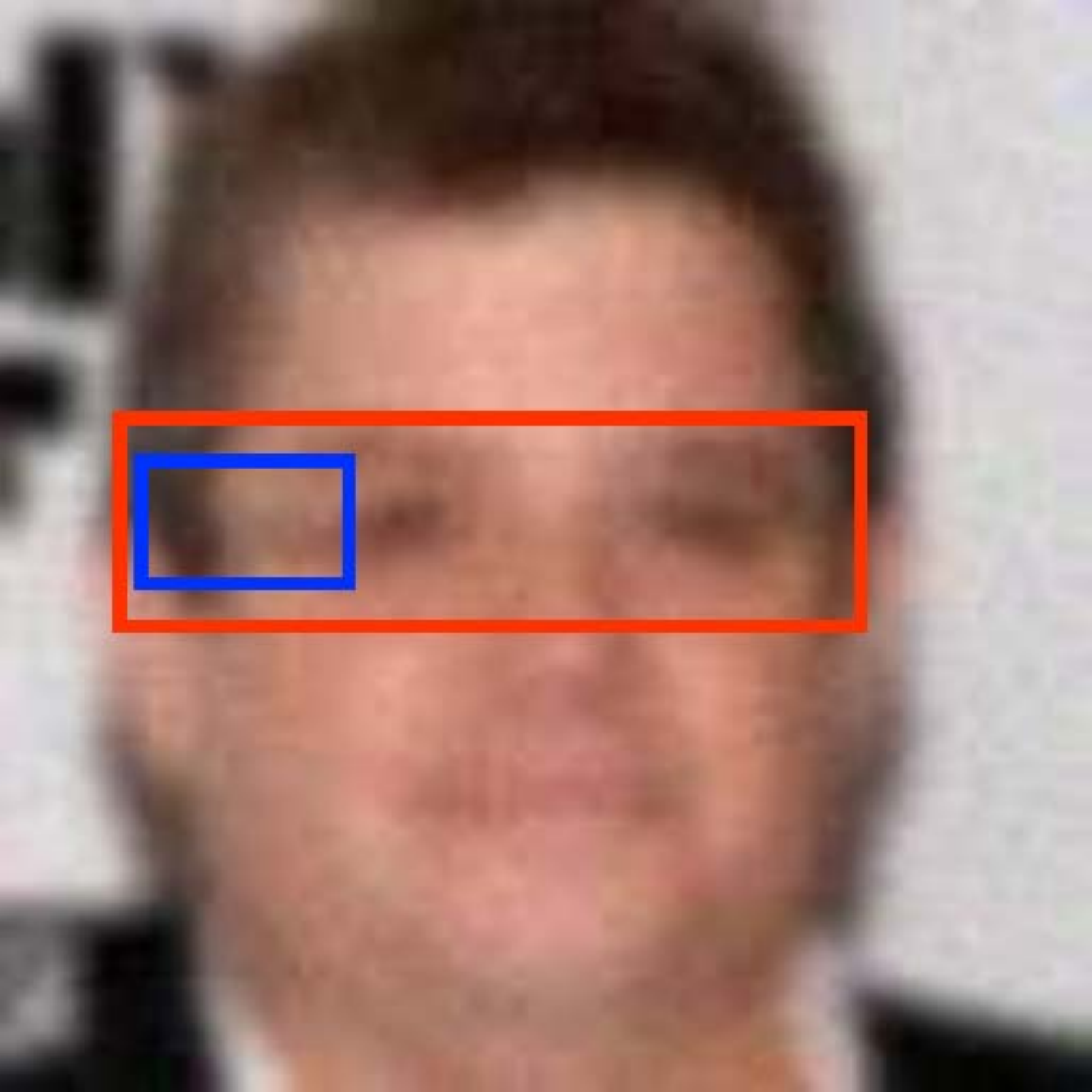}&
   \includegraphics[width=\swseven]{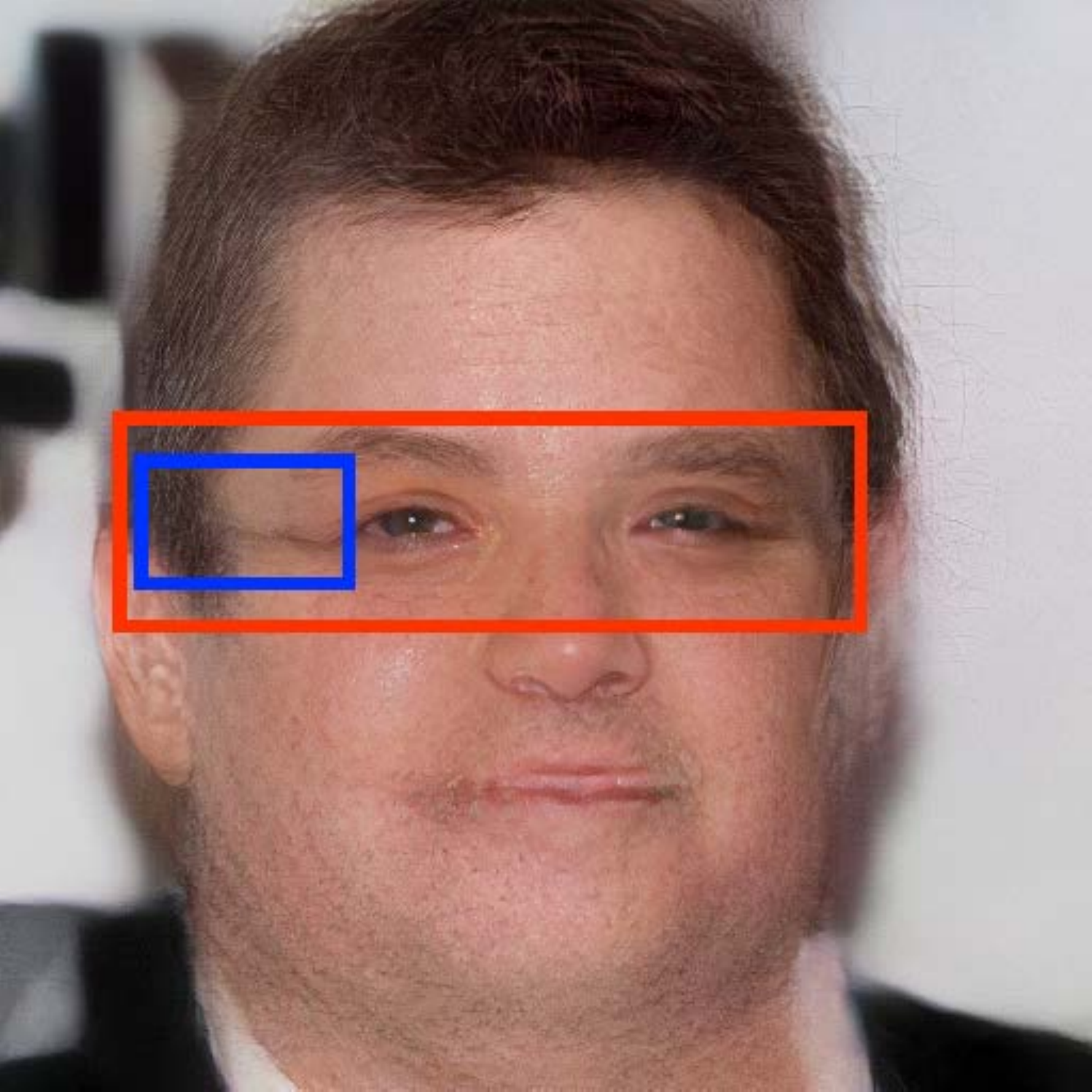}&
   \includegraphics[width=\swseven]{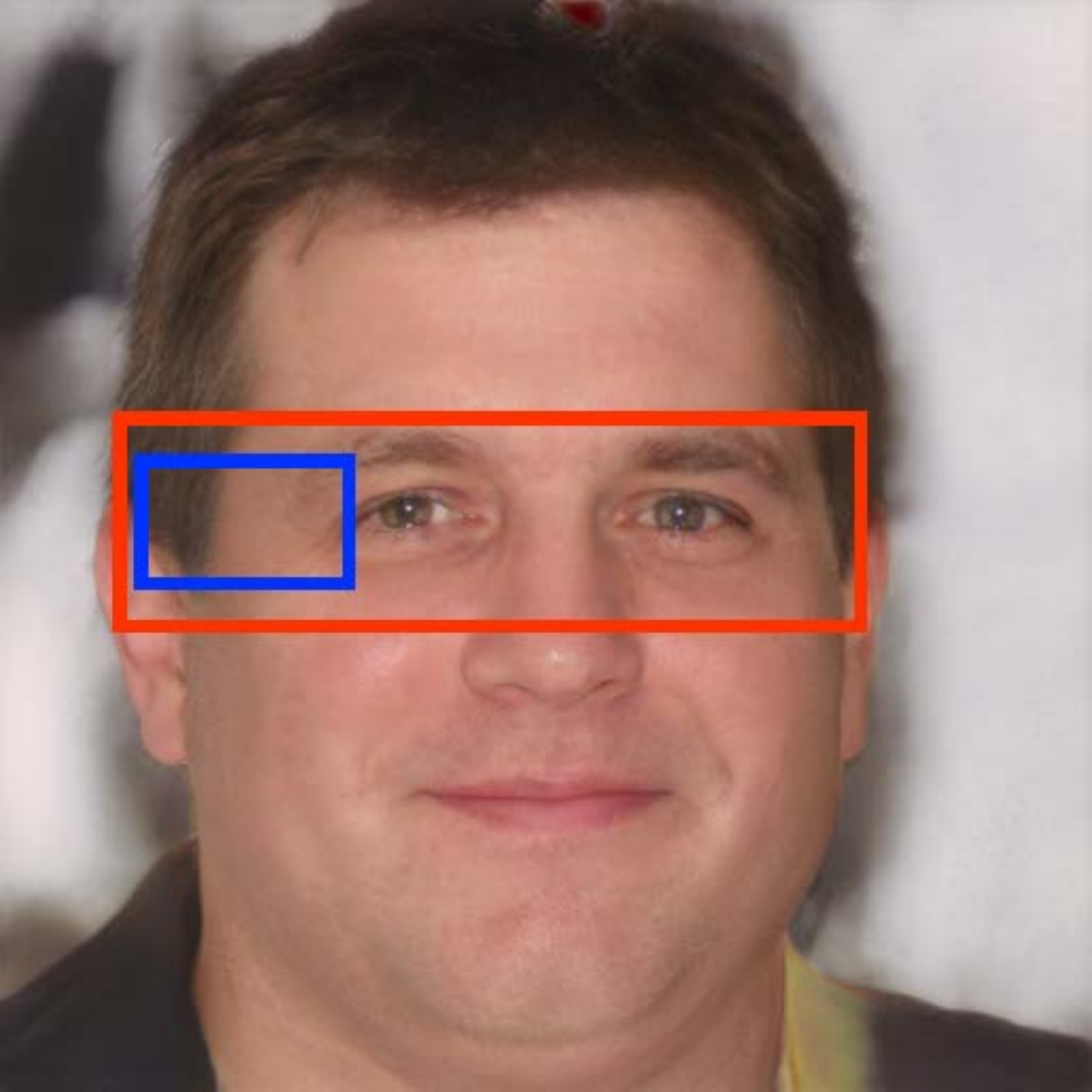}&
   \includegraphics[width=\swseven]{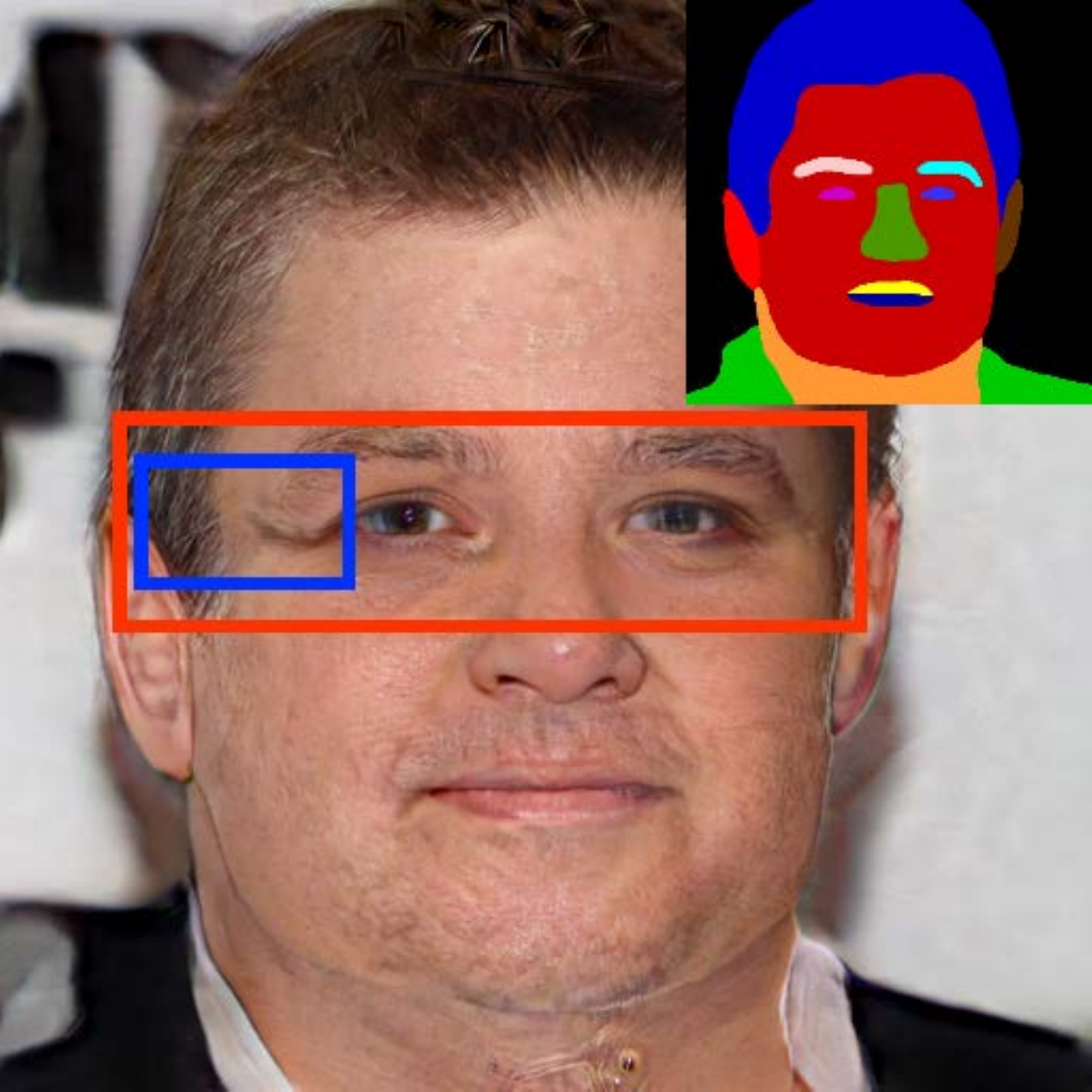}&
   \includegraphics[width=\swseven]{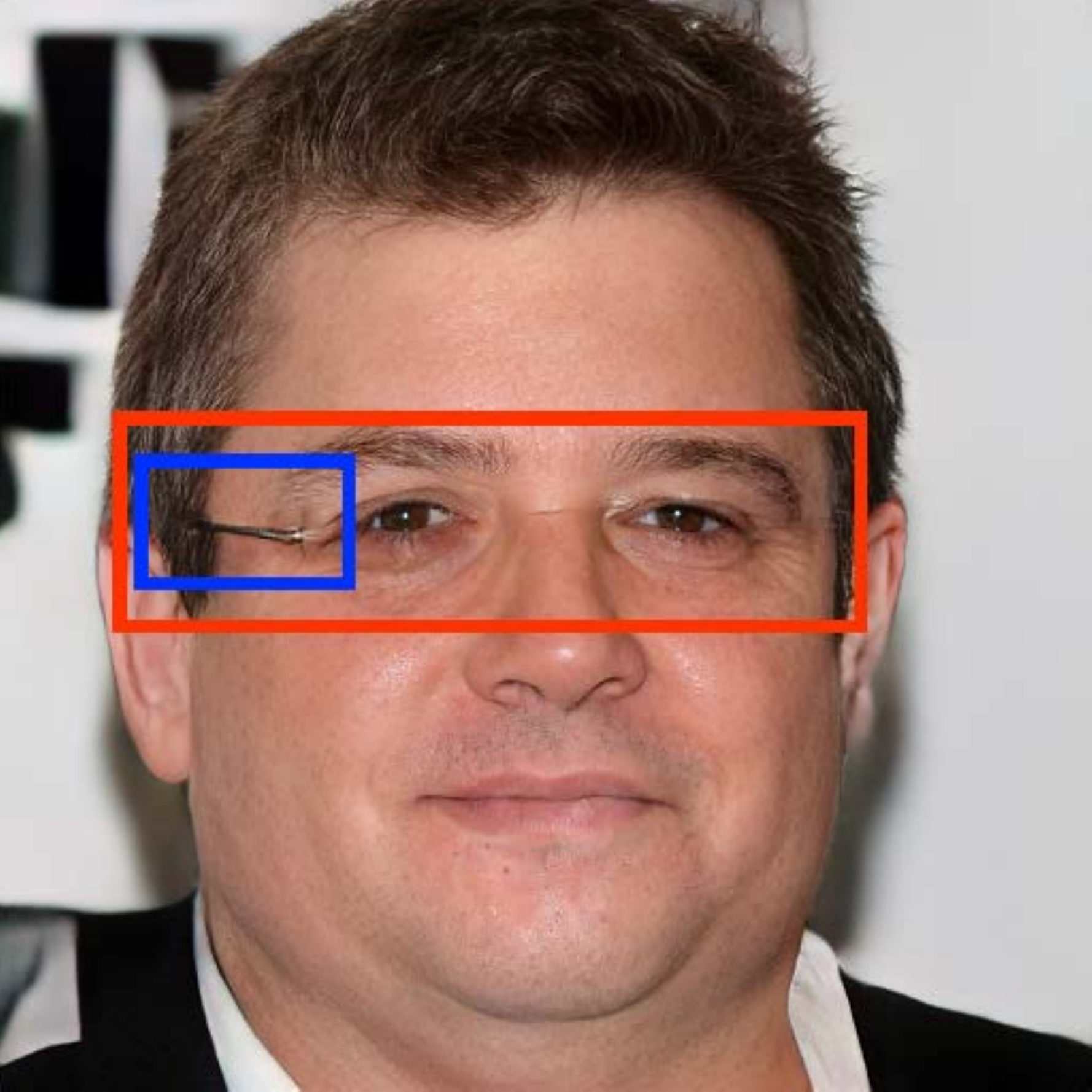}&
   \includegraphics[width=\swseven]{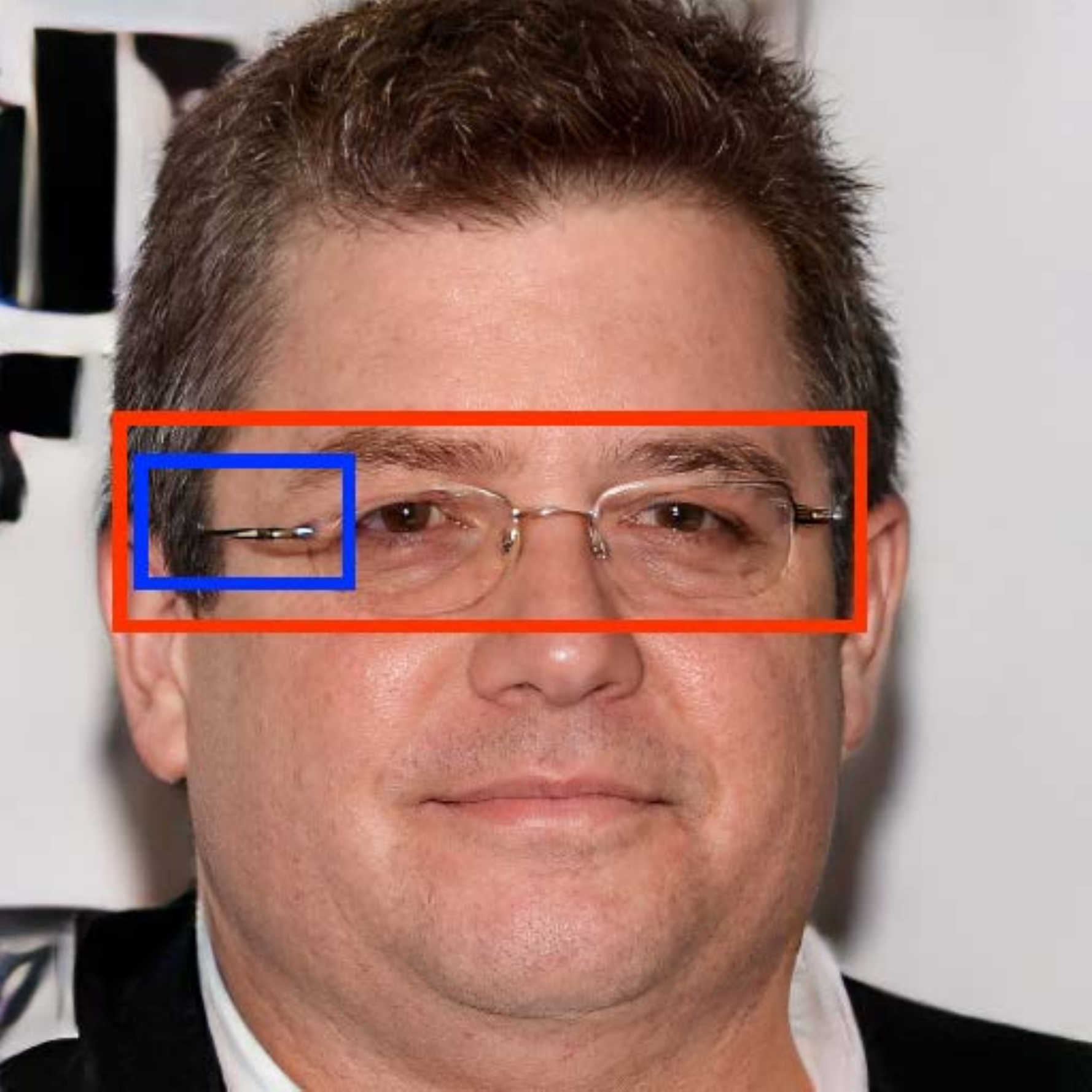} &
   \includegraphics[width=\swseven]{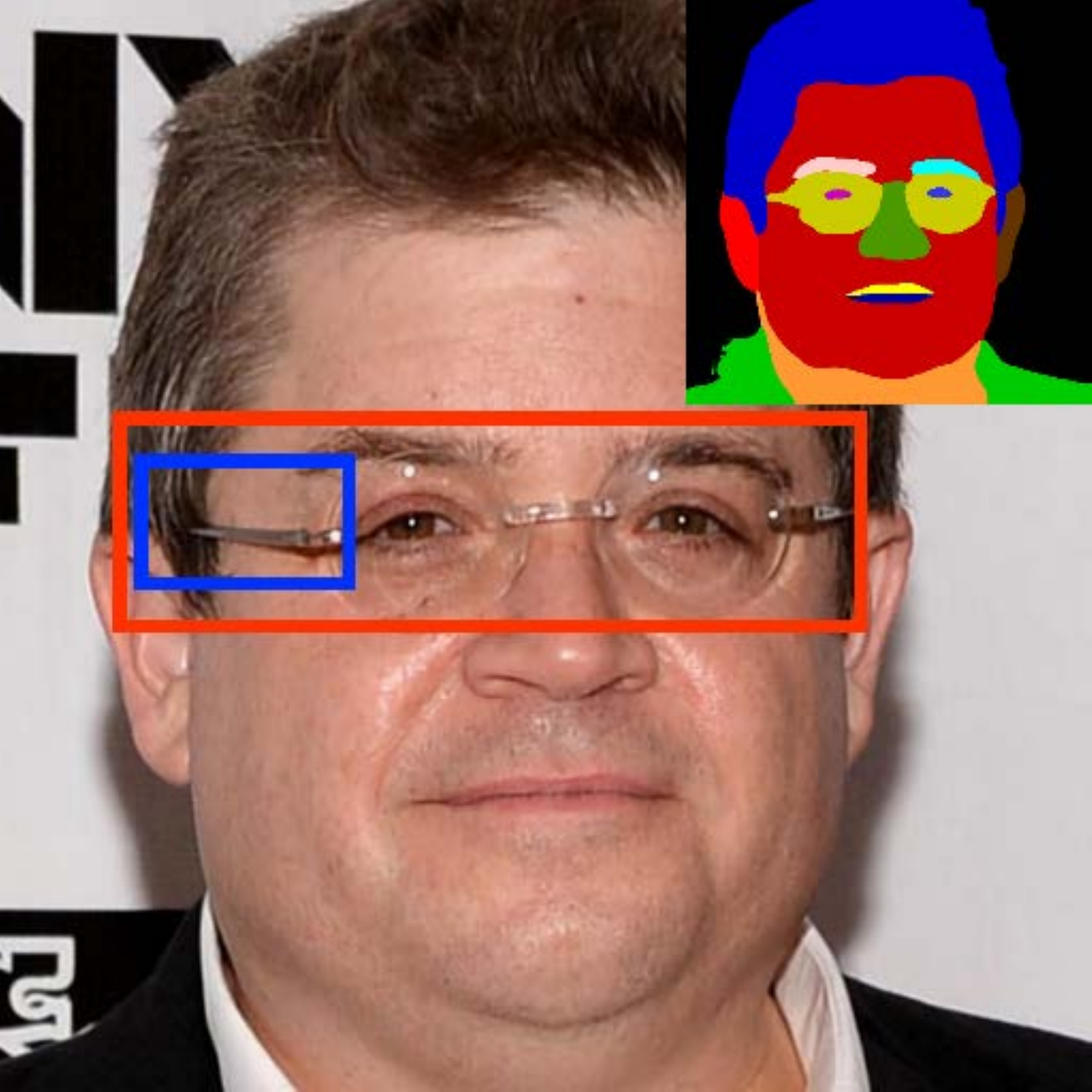} \\
   Input & DFDNet~\cite{li2020blind}  &  PULSE~\cite{menon2020pulse}, & PSFRGAN~\cite{chen2021progressive} & GFP-GAN~\cite{wang2021towards} & \textbf{RestoreFormer} & GT
\end{tabular}
\end{center}
\vspace{-6mm}
\caption{Qualitative comparison on the \textbf{CelebA-Test}. The results of our RestoreFormer have a more realistic overview and contain more details in eyes, mouth, and hair. \textbf{Zoom in for a better view and more results are shown in supplementary materials}.}
\label{fig:celeba}
\end{figure*}%

\begin{table}
  \small
  \centering
  \begin{tabular}{c|c|c|c|c|c} % {@{}lc@{}}
    \toprule
    Methods & FID$\downarrow$ & PSNR$\uparrow$ & SSIM$\uparrow$ & LPIPS$\downarrow$ & IDD$\downarrow$ \\
    \midrule
    Input & 132.69 & \textcolor{red}{24.96} & \textcolor{blue}{0.6624} & 0.4989 & 0.9308 \\
    \midrule
    DFDNet~\cite{li2020blind} & 52.92 & 24.10 & 0.6092 & 0.4478 & 0.7581  \\
    PSFRGAN~\cite{chen2021progressive} & 43.88 & 24.45 & 0.6308 & 0.4186 & 0.7163 \\
    Wan~\etal~\cite{wan2020bringing} & 70.21 & 23.00 & 0.6189 & 0.4778 & 0.8018 \\
    % Wan~\etal~\cite{wan2020bringing} & 67.58 & 24.71 & 
    PULSE~\cite{menon2020pulse} & 67.75 & 21.61 & 0.6287 & 0.4657 & 1.2019 \\
    GFP-GAN\cite{wang2021towards} & \textcolor{blue}{42.39} & \textcolor{blue}{24.46} & \textcolor{red}{0.6684} & \textcolor{red}{0.3551} & \textcolor{blue}{0.6034} \\
    \textbf{RestoreFormer} & \textcolor{red}{41.45} & 24.42 & 0.6404 & \textcolor{blue}{0.3650} & \textcolor{red}{0.5650} \\
    \midrule
    GT & 43.43 & $\infty$ & 1 & 0 & 0 \\
    \bottomrule
  \end{tabular}
  \vspace{-0.3cm}
  \caption{Quantitative comparisons on \textbf{CelebA-Test}. Our RestoreFormer has better performance based on FID and IDD which indicates the realness and identity preserving property of our method. It also gets a comparable results on PSNR, SSIM, and LPIPS.
  }
  \label{tab:celeba}
\end{table}

\subsection{Comparison with State-of-the-art Methods}{\label{sec:sofa}}
To validate the effectiveness of our proposed method on blind face restoration, we compare its performance with several state-of-the-art face restoration methods, including DFDNet~\cite{li2020blind}, 
PSFRGAN~\cite{chen2021progressive}, Wan~\etal~\cite{wan2020bringing}, PULSE~\cite{menon2020pulse}, and GFP-GAN~\cite{wang2021towards}.
These methods cover different types of priors, \emph{e.g}\onedot reference (DFDNet), geometric priors (PSFRGAN), and generative priors (Wan~\etal, PULSE, and GFP-GAN).
%
%And all of them are performed on both synthetic dataset and real-world datasets.

\noindent\textbf{Synthetic Dataset.}
We first compare our RestoreFormer with other methods on CelebA-Test.
%, which has the same degradation synthetic process as training.
%
The quantitative results of each method are shown in Table~\ref{tab:celeba}.
Our RestoreFormer has a better performance based on FID and IDD which indicates its restored faces are closer to the real face and have a more similar identity with their ground truth at the same time.
It also has comparable results on the pixel-wise and perceptual metrics: PSNR, SSIM, and LPIPS, although they have been proven not that consistent with the subjective evaluation of human beings~\cite{blau20182018,ledig2017photo}.
%
%This quantitative performance is consistent with the qualitative results.
%
As to the visual results, PULSE~\cite{menon2020pulse} can generate visually pleasant results in Figure~\ref{fig:celeba}.
However, it cannot preserve the human identity compared with RestoreFormer.
%
%Unlike the priors in RestoreFormer can properly capture high-quality facial information and be robust to the degraded input, the heatmap in PSFGAN~\cite{chen2021progressive} deteriorates significantly when the input face degrading too much which will lead to the restored face containing artifacts in Figure~\ref{fig:celeba}.
%
Even though the left eyebrow and eyeglasses can be detected by DFDNet~\cite{li2020blind} and GFP-GAN~\cite{wang2021towards} in the first and second row (blue box) of Figure~\ref{fig:celeba}, they are only partially reconstructed.
This may be because only local information is considered when fusing degraded information and priors.
With the help of MHCA, RestoreFormer can reconstruct the eyebrow and eyeglasses better in Figure~\ref{fig:celeba}.
Eyeglasses also cannot be restored by PSFGAN~\cite{chen2021progressive} in Figure~\ref{fig:celeba}.
This is because its estimated heatmap (upper-right corner of PSFGAN~\cite{chen2021progressive}), from the degraded input, is inaccurate.

\begin{table}
  \small
  \centering
  \begin{tabular}{c|c|c|c} % {@{}lc@{}}
    \toprule
    Methods & \textbf{LFW-Test} & \textbf{CelebChild-Test} & \textbf{WebPhoto-Test} \\    \midrule 
    Input & 137.56 &  144.42 &  170.11  \\
    \midrule
    DFDNet~\cite{li2020blind} & 62.57 & 111.55 & 100.68  \\
    PSFRGAN~\cite{chen2021progressive} & 53.92 & 106.61 & \textcolor{blue}{84.98} \\
    Wan~\etal~\cite{wan2020bringing} & 73.19 & 115.70 & 100.40 \\
    PULSE~\cite{menon2020pulse} & 64.86 & \textcolor{blue}{102.74} & 86.45\\
    GFP-GAN~\cite{wang2021towards} & \textcolor{blue}{49.96} & 111.78 & 87.35 \\
    \textbf{RestoreFormer} & \textcolor{red}{47.75} & \textcolor{red}{101.22} & \textcolor{red}{77.33} \\
    % \textbf{RestoreFormer-uneven} & 50.65 &  & 101.36 &  & 76.31 & \\
    \bottomrule
  \end{tabular}
  \vspace{-0.3cm}
  \caption{Quantitative comparisons on three \textbf{real-world dataset} in terms of FID. RestoreFormer performs the best.}
  \label{tab:real_fid}
\end{table}

\iffalse
\begin{table}
  \small
  \centering
  \begin{tabular}{c|c|c} % {@{}lc@{}}
    \toprule
     & \textbf{LFW-Test} & \textbf{WebPhoto-Test} \\
    \midrule
     Methods & \multicolumn{2}{c}{\textbf{RestoreFormer}} \\
    \midrule
    DFDNet~\cite{li2020blind} & 15.41\%/\textbf{84.59\%} & 28.78\%\textbf{/71.22\%}  \\
    PSFRGAN\cite{chen2021progressive} & 9.96\%\textbf{/90.04\%} & 20.90\%\textbf{/79.10\%} \\
    GFP-GAN~\cite{wang2021towards} & 9.89\%\textbf{/90.11\%} & 10.40\%\textbf{/89.60\%}\\
    \bottomrule
  \end{tabular}
  \vspace{-0.3cm}
  \caption{User study results on \textbf{LFW-Test} and \textbf{WebPhoto-Test}.
  %
  For ``a/b", a is the percentage where the compared method is considered better than our RestoreFormer, and b is the percentage where our RestoreFormer is considered better than the compared method.}
  \label{tab:real_us}
\end{table}
\fi

\begin{table}
  \small
  \centering
  \begin{tabular}{c|c|c} % {@{}lc@{}}
    \toprule
     & \textbf{LFW-Test} & \textbf{WebPhoto-Test} \\
    \midrule
     Methods & \multicolumn{2}{c}{\textbf{RestoreFormer}} \\
    \midrule
    DFDNet~\cite{li2020blind} & 9.96\%\textbf{/90.04\%} & 20.90\%\textbf{/79.10\%} \\
    PSFRGAN\cite{chen2021progressive} & 9.89\%\textbf{/90.11\%} & 10.40\%\textbf{/89.60\%} \\
    GFP-GAN~\cite{wang2021towards} & 15.41\%/\textbf{84.59\%} & 28.78\%\textbf{/71.22\%}  \\
    \bottomrule
  \end{tabular}
  \vspace{-0.3cm}
  \caption{User study results on \textbf{LFW-Test} and \textbf{WebPhoto-Test}.
  For ``a/b", a is the percentage where the compared method is considered better than our RestoreFormer, and b is the percentage where our RestoreFormer is considered better than the compared method.}
  \label{tab:real_us}
\end{table}

\renewcommand{\tabcolsep}{.5pt}
\begin{figure*}
\hsize=\textwidth
\vspace{-0.3cm}
% \begin{minipage}{\textwidth}
\begin{center}
\begin{tabular}{ccccccc}
%   \multicolumn{7}{c}{\textbf{\textit{LFW-Test}}} \\
   % \vspace{-1.0mm}
   % \includegraphics[width=\swseven]{figures/exp/lfw/038_Alex_Cabrera_0001_org.pdf}&
   % \includegraphics[width=\swseven]{figures/exp/lfw/038_Alex_Cabrera_0001_dfd.pdf}&
   % \includegraphics[width=\swseven]{figures/exp/lfw/038_Alex_Cabrera_0001_bop.pdf}&
   % \includegraphics[width=\swseven]{figures/exp/lfw/038_Alex_Cabrera_0001_pulse.pdf}&
   % \includegraphics[width=\swseven]{figures/exp/lfw/038_Alex_Cabrera_0001_pfsrgan.pdf}&
   % \includegraphics[width=\swseven]{figures/exp/lfw/038_Alex_Cabrera_0001_gfp.pdf}&
   % \includegraphics[width=\swseven]{figures/exp/lfw/038_Alex_Cabrera_0001_ours.pdf} \\
   \vspace{-1.0mm}
   \includegraphics[width=\swseven]{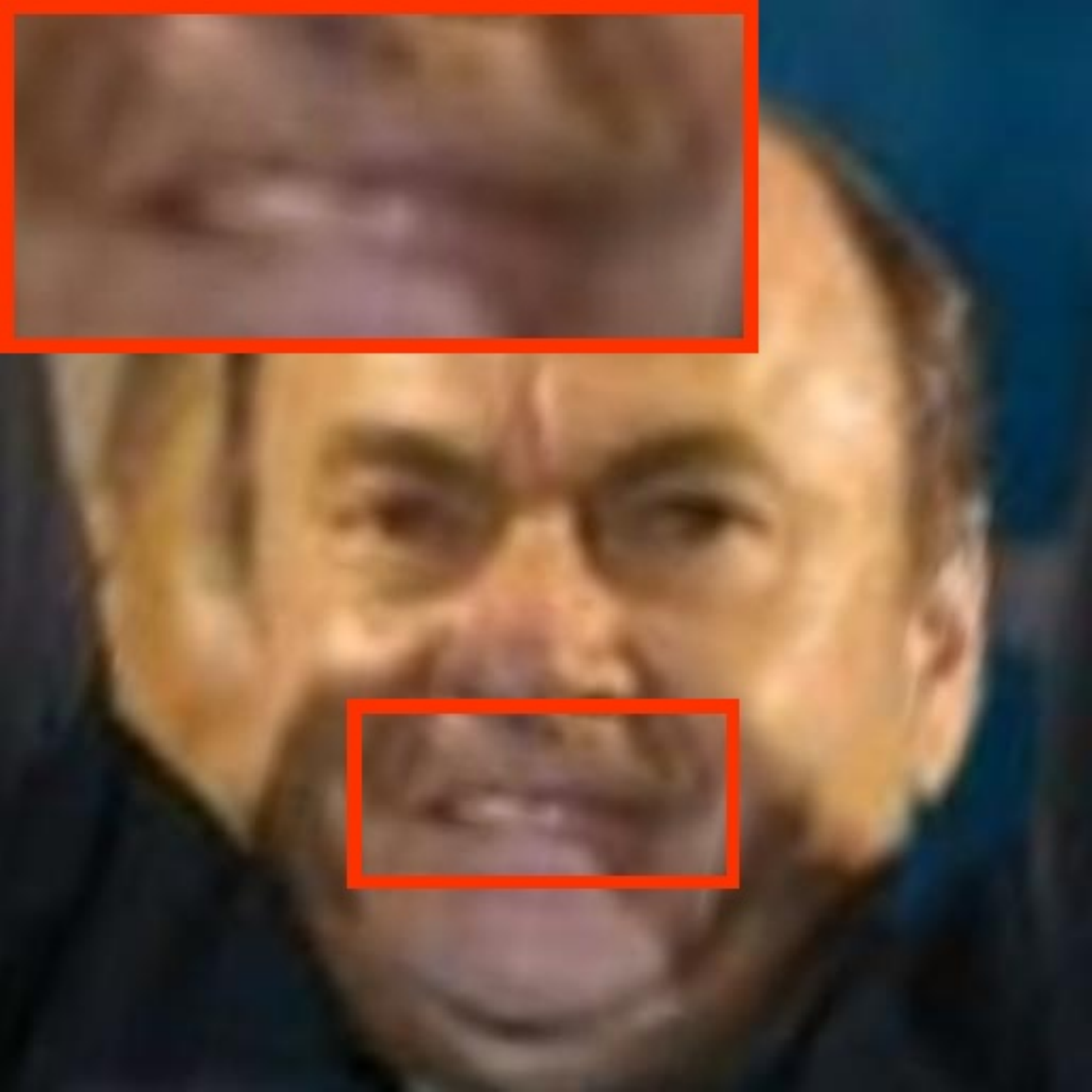}&
   \includegraphics[width=\swseven]{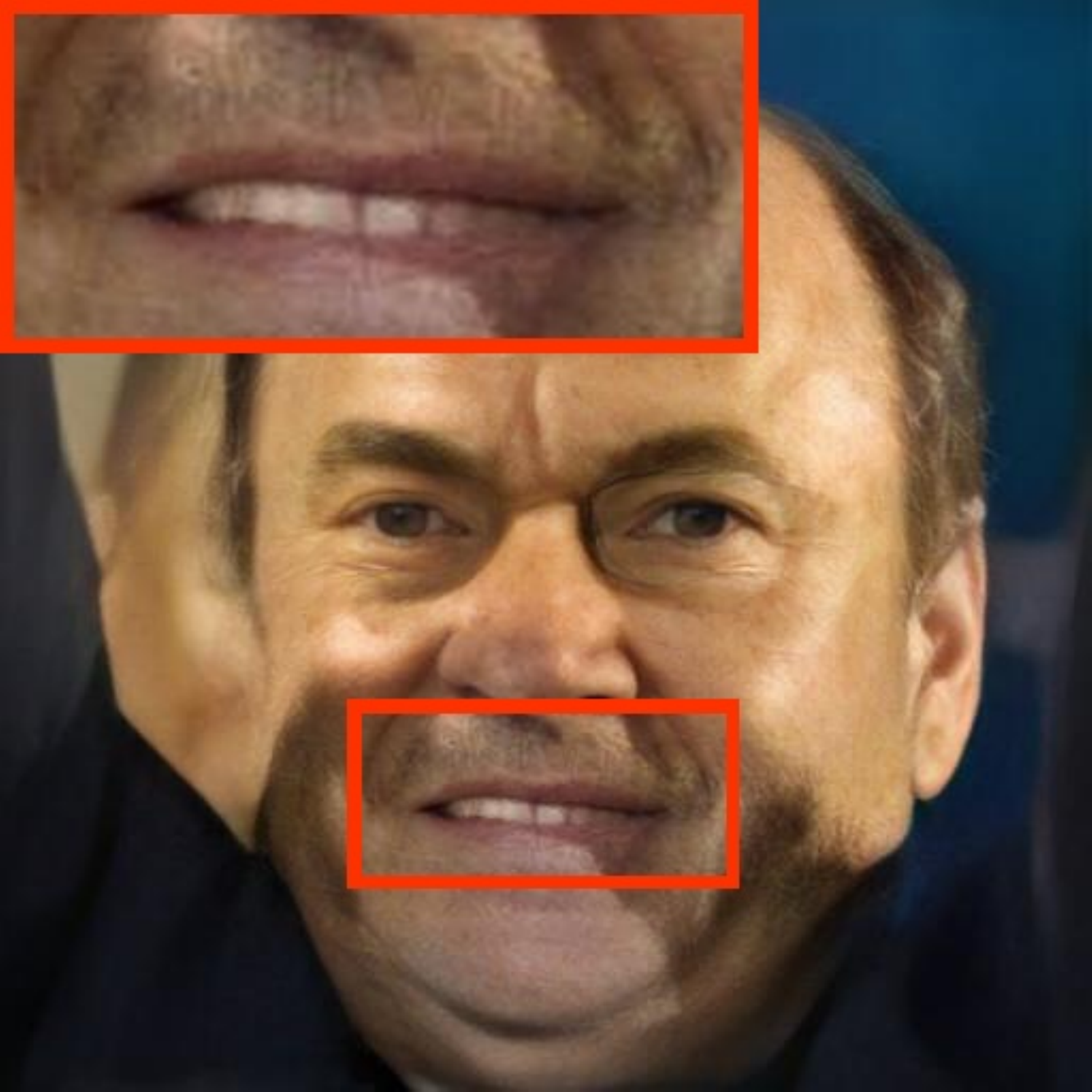}&
   \includegraphics[width=\swseven]{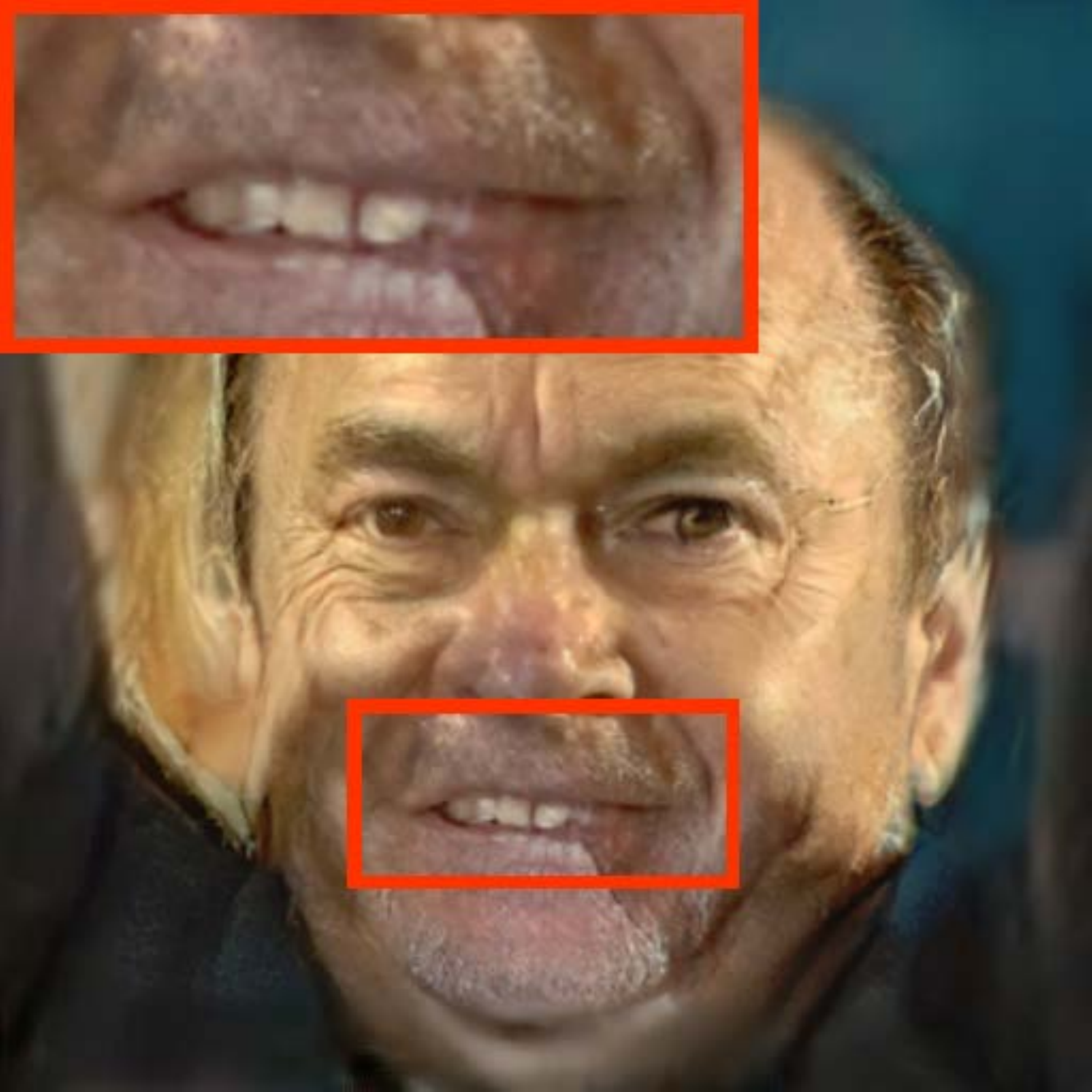}&
   \includegraphics[width=\swseven]{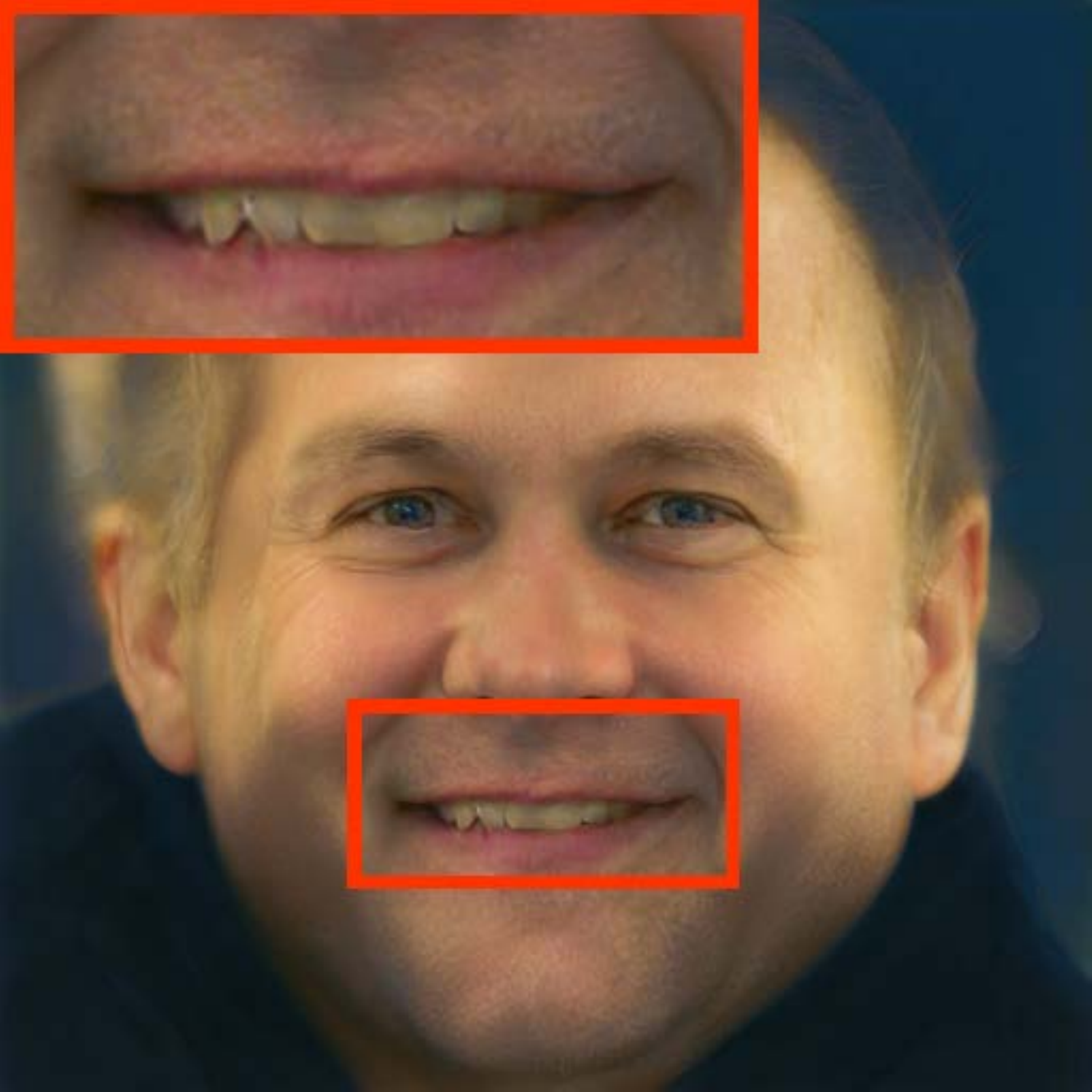}&
   \includegraphics[width=\swseven]{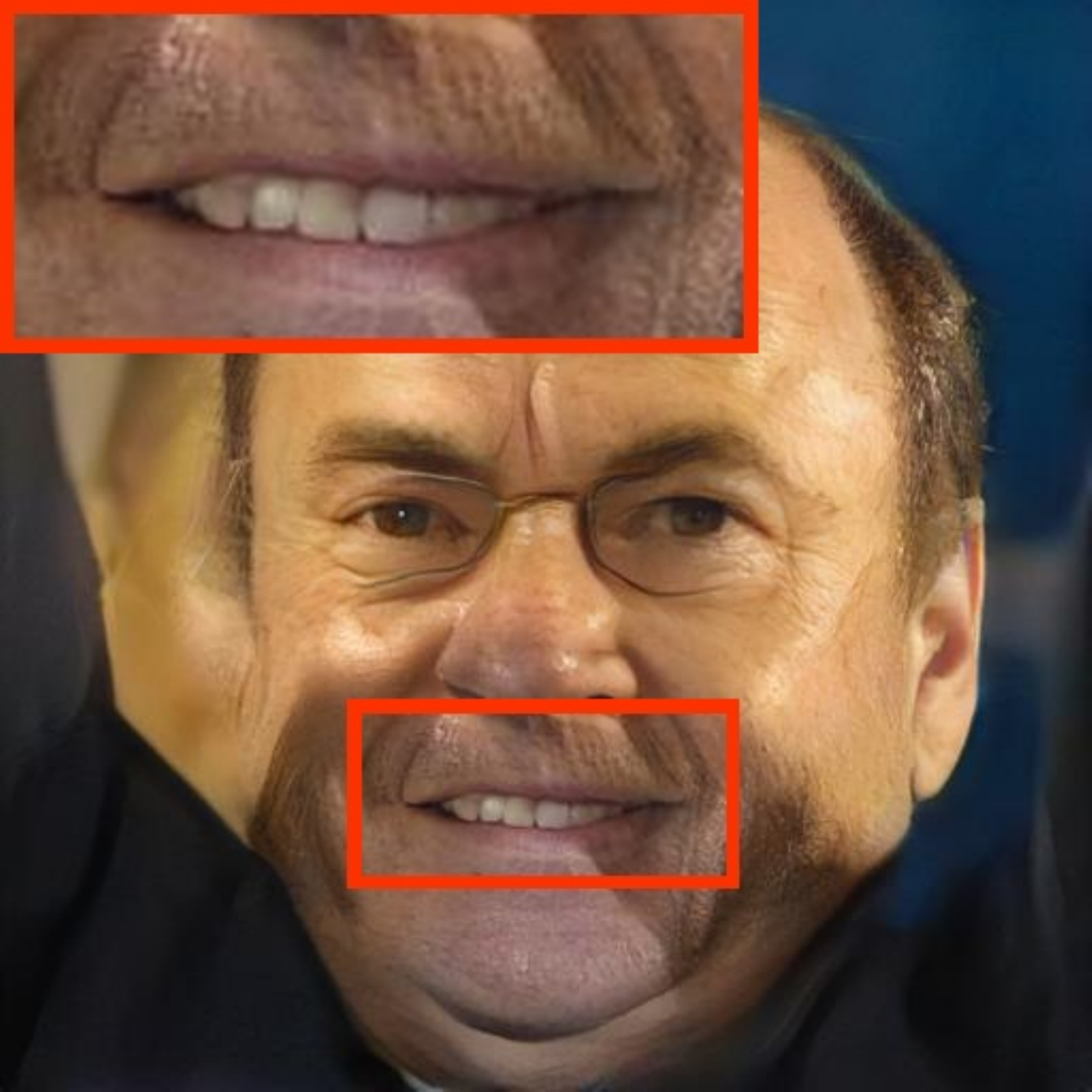}&
   \includegraphics[width=\swseven]{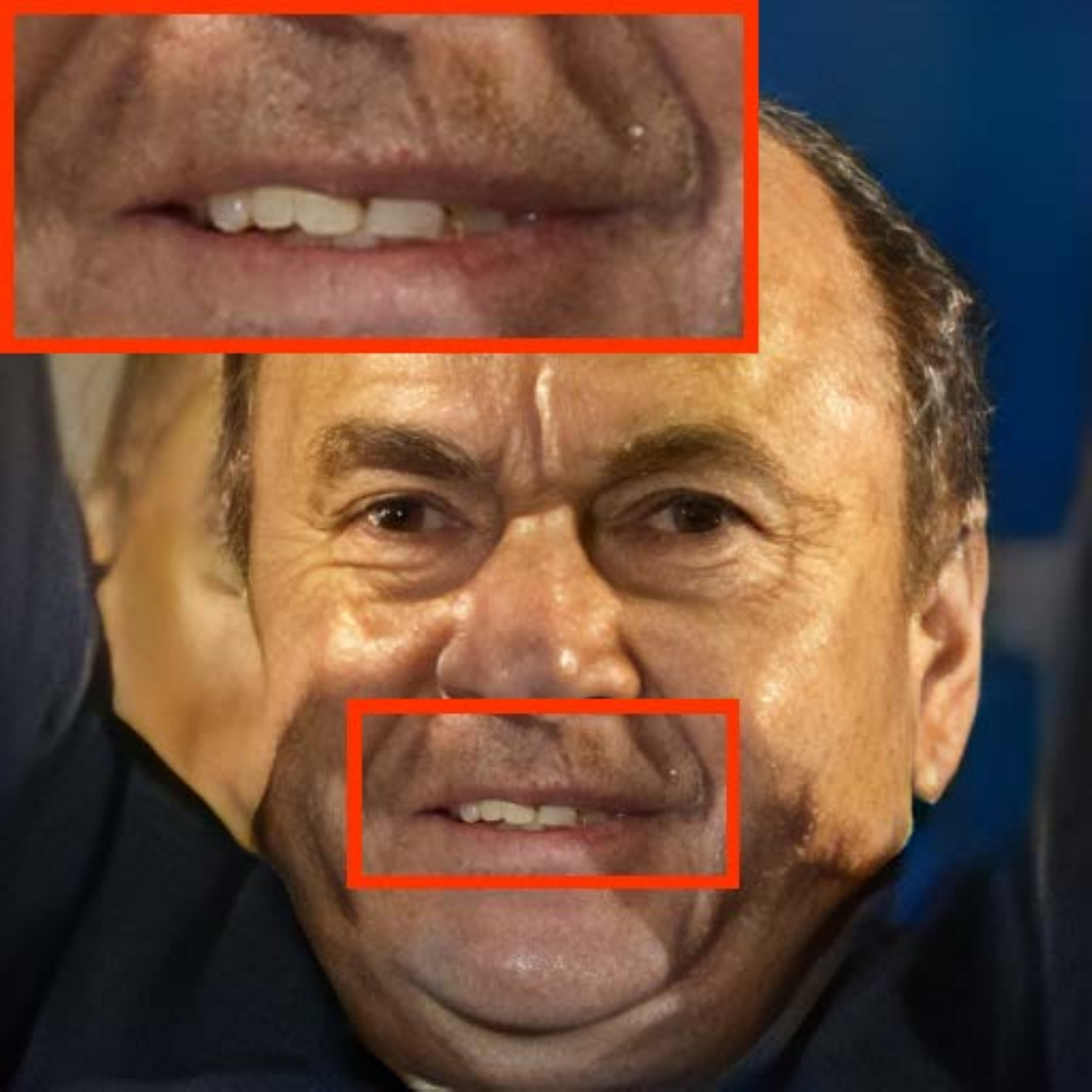}&
   \includegraphics[width=\swseven]{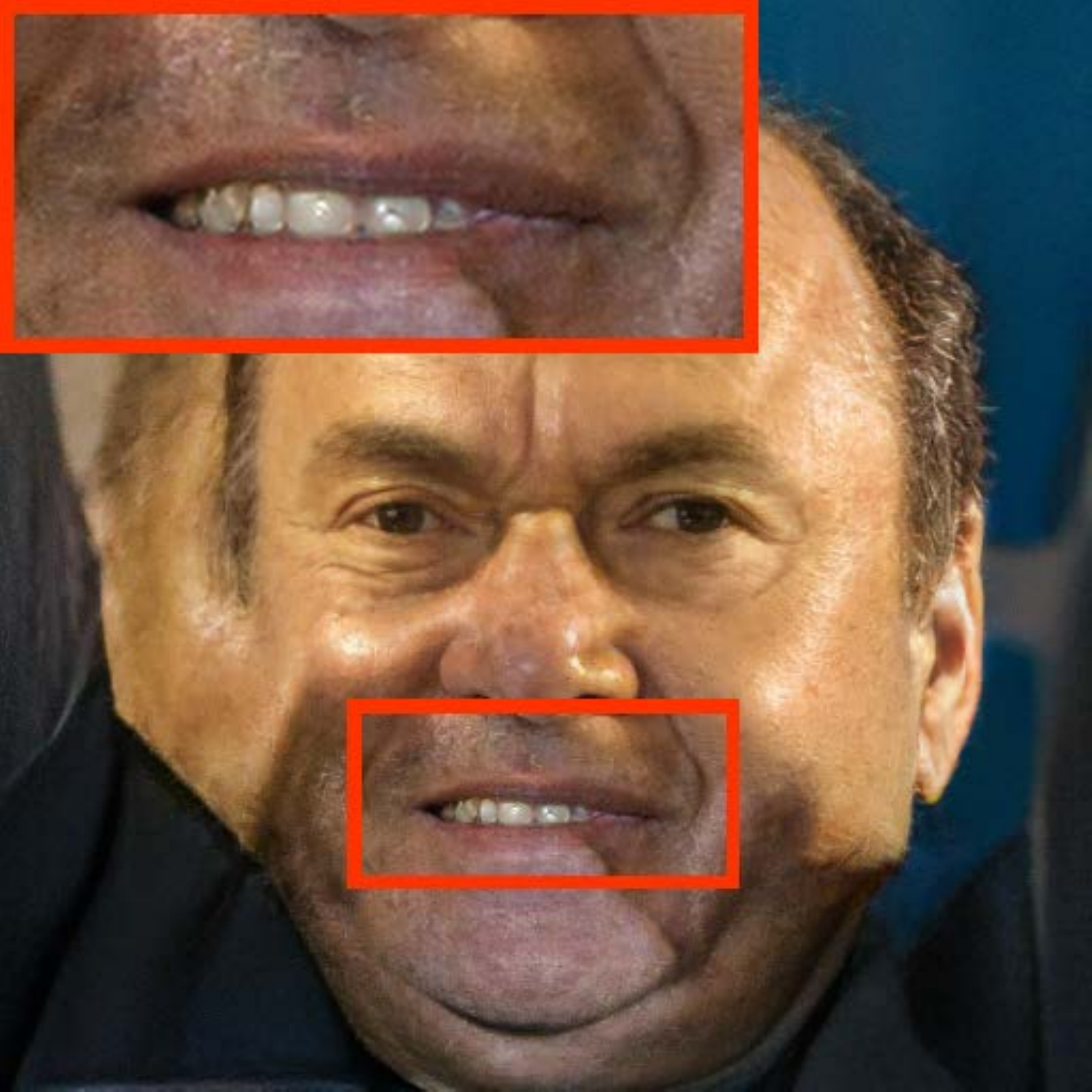} \\
%   \multicolumn{7}{c}{\textbf{\textit{CelebChild-Test}}} \\
   \vspace{-1.0mm}
   \includegraphics[width=\swseven]{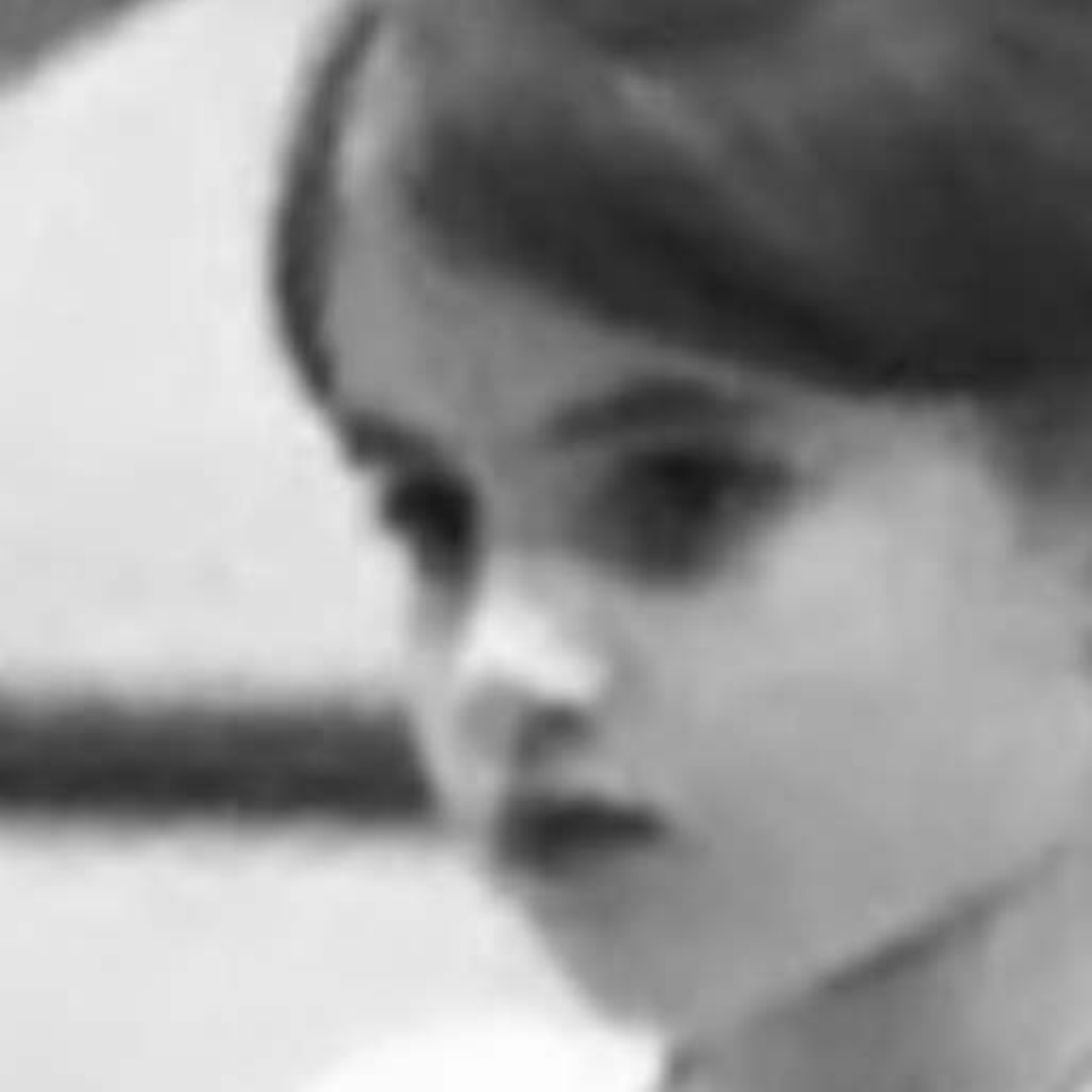}&
   \includegraphics[width=\swseven]{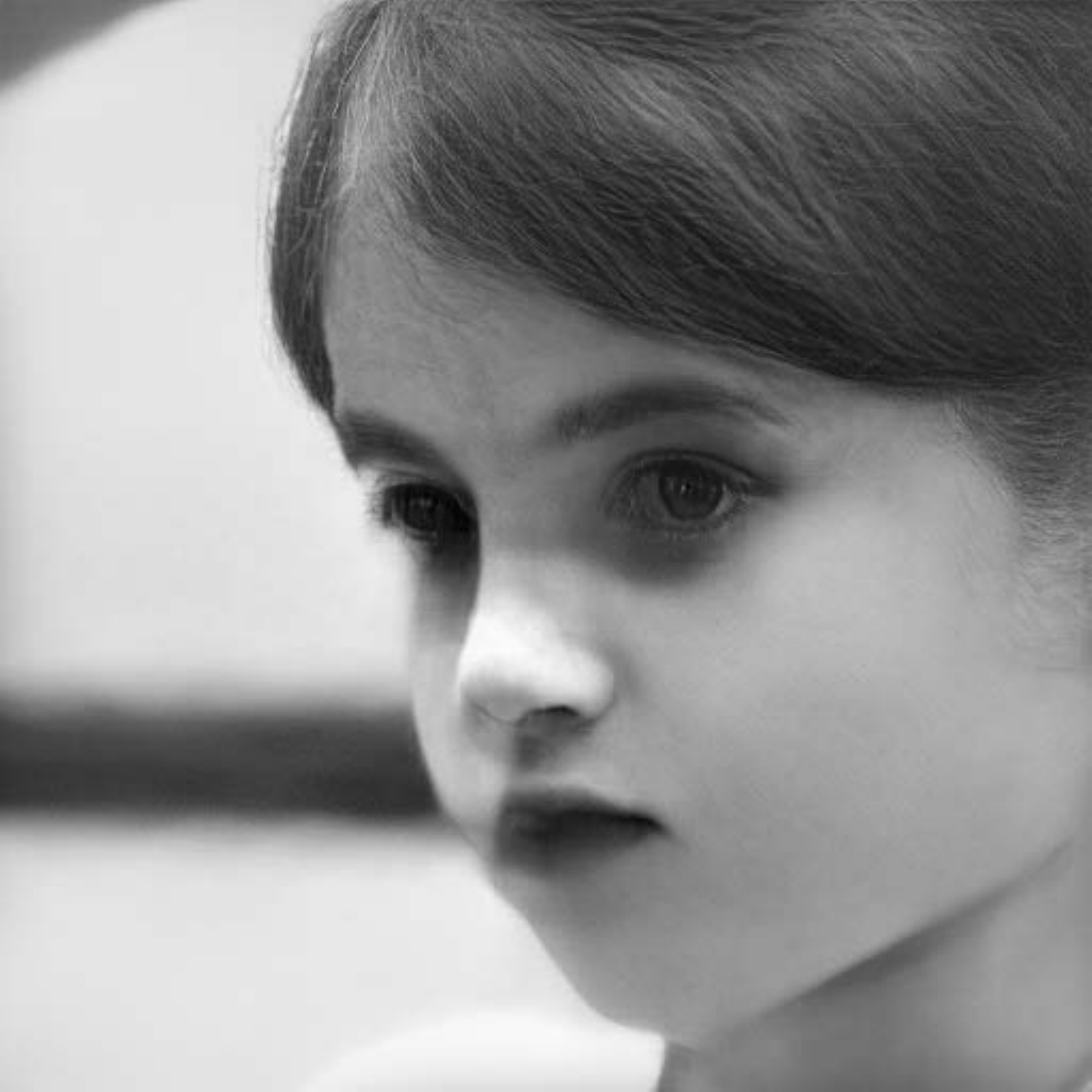}&
   \includegraphics[width=\swseven]{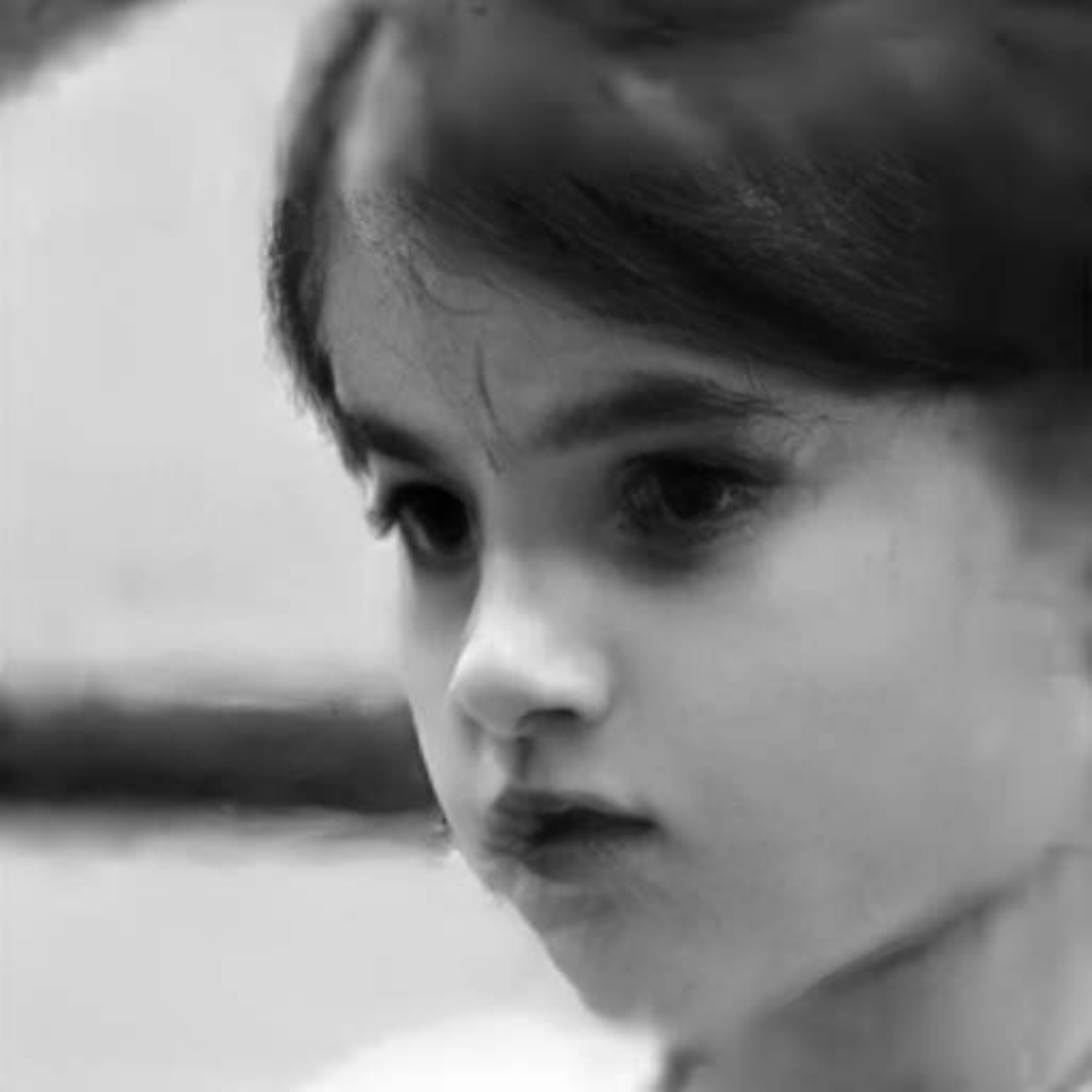}&
   \includegraphics[width=\swseven]{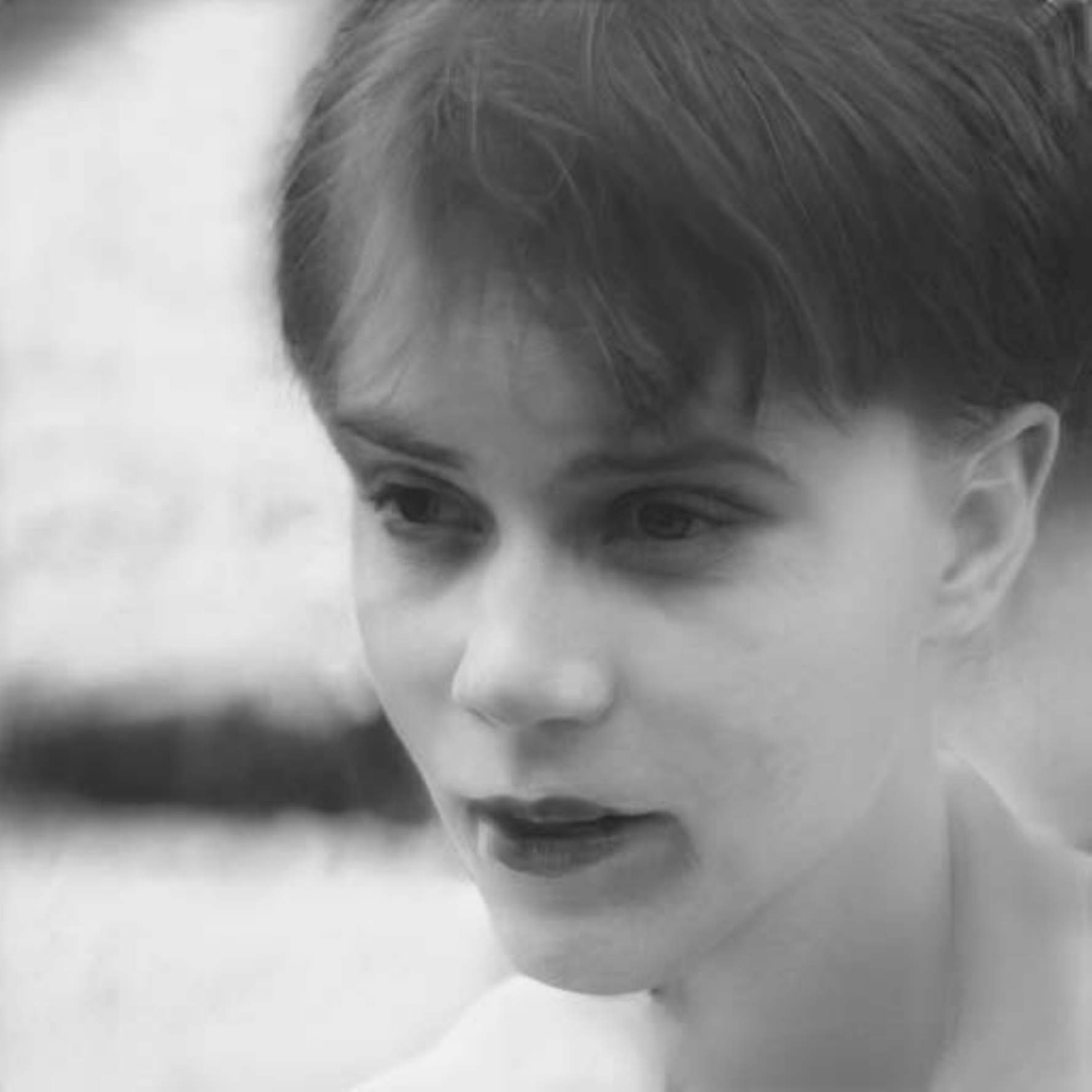}&
   \includegraphics[width=\swseven]{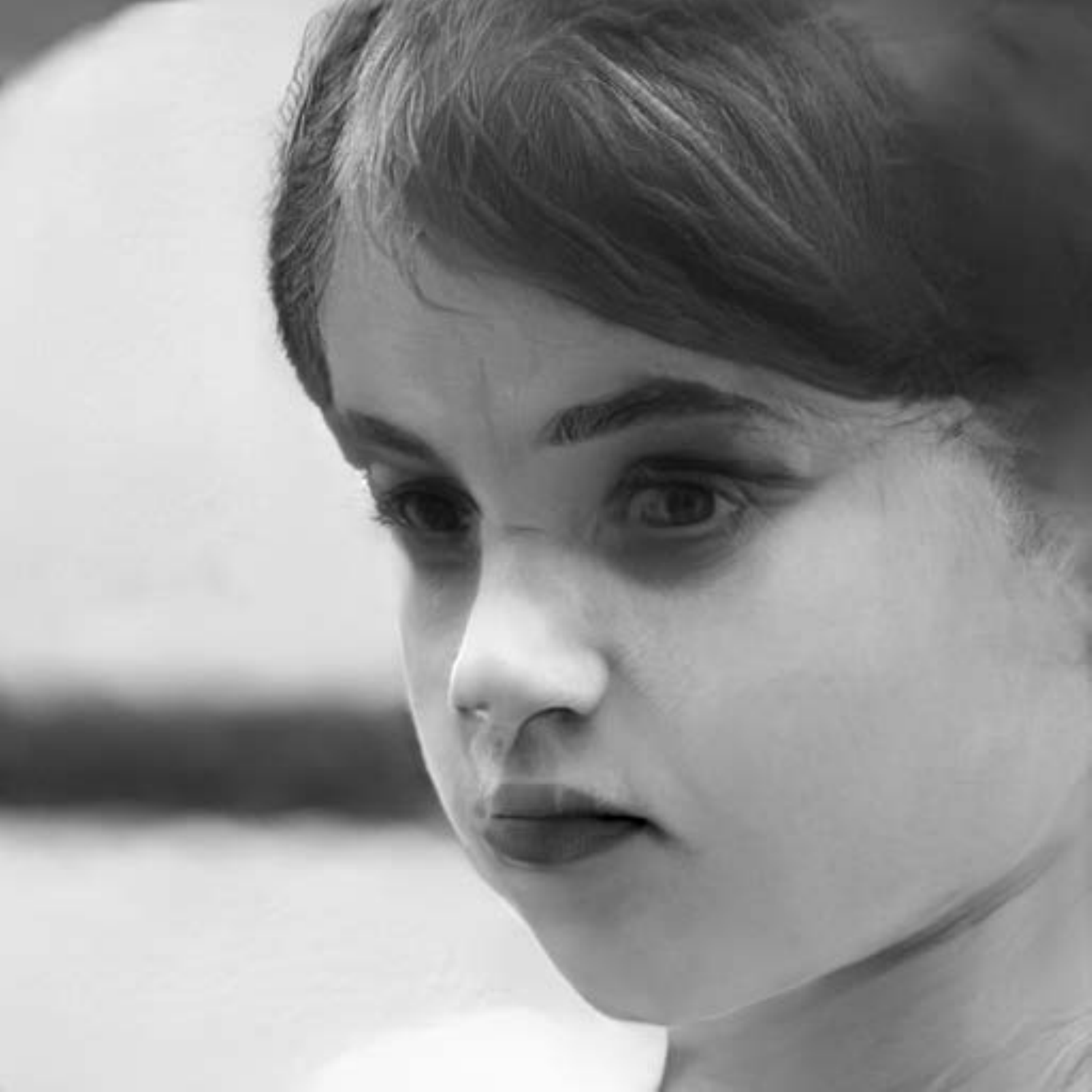}&
   \includegraphics[width=\swseven]{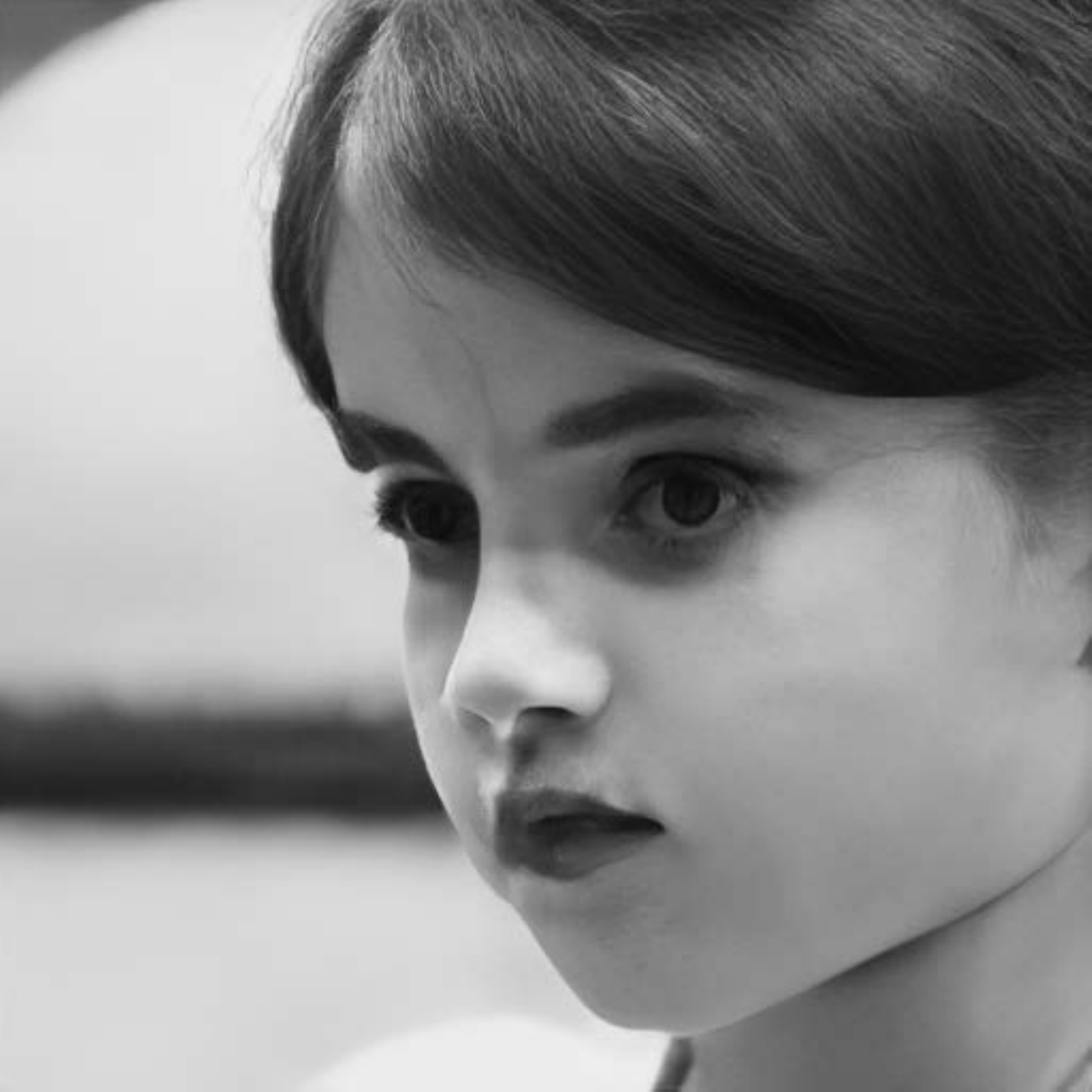}&
   \includegraphics[width=\swseven]{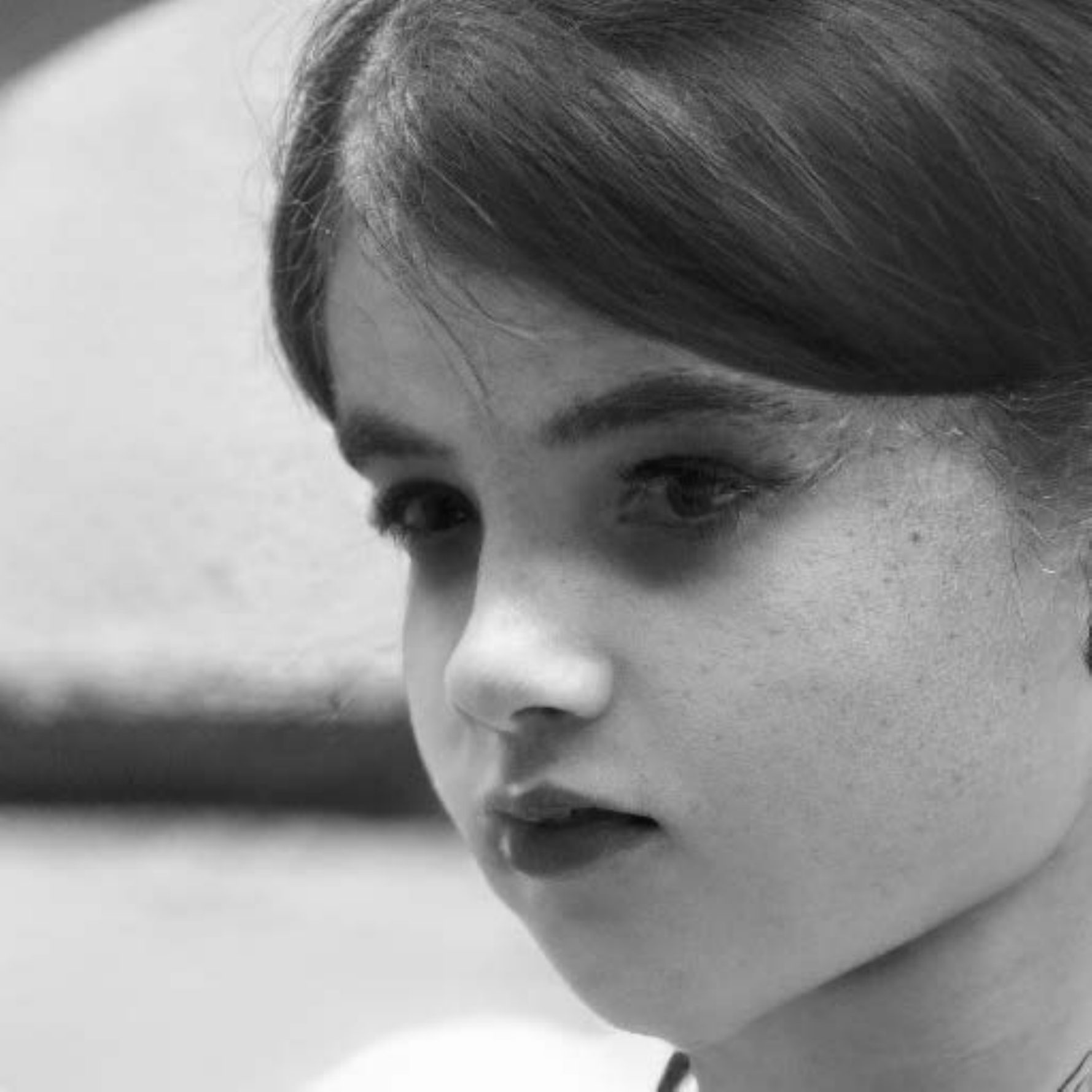} \\
   \vspace{-1.0mm}
   \includegraphics[width=\swseven]{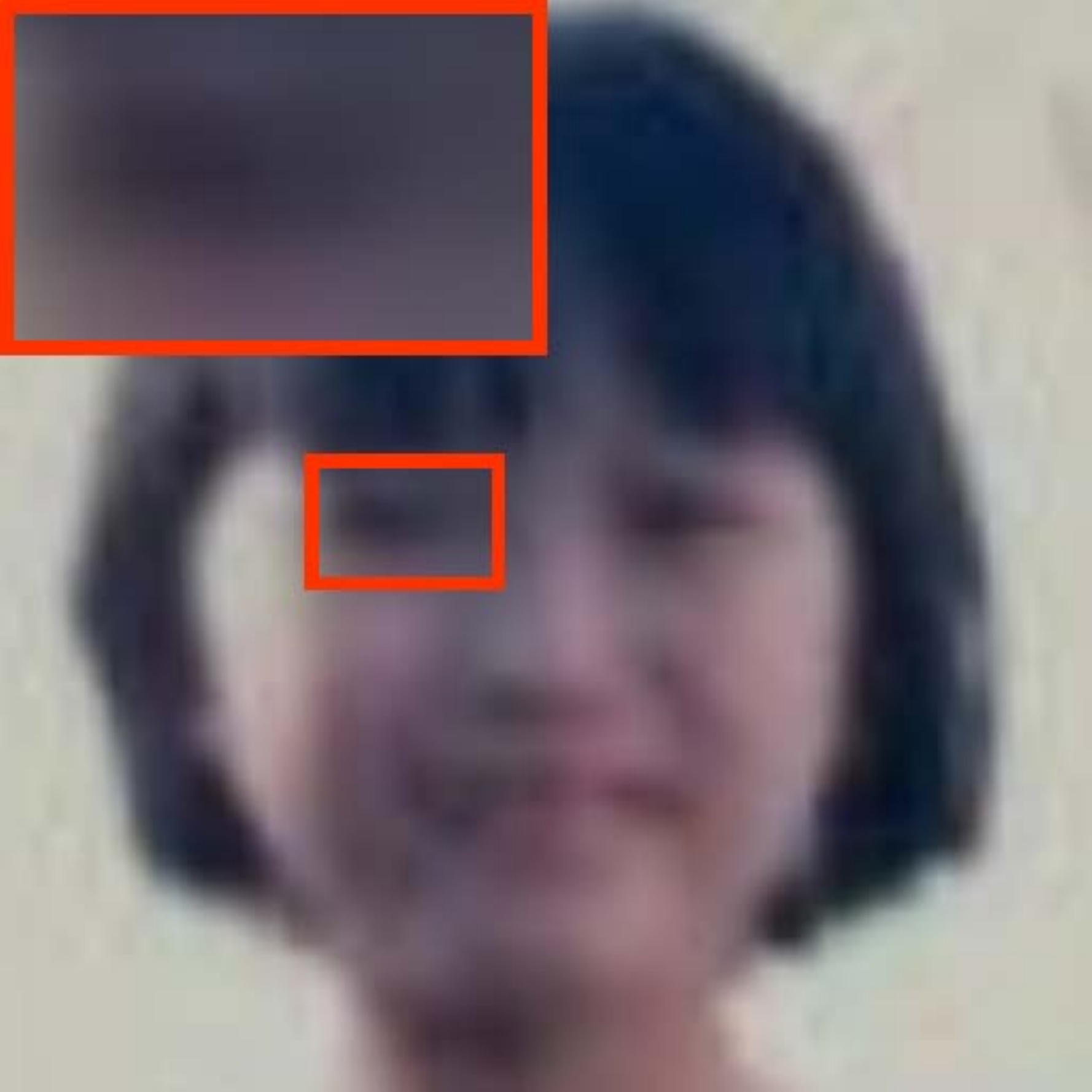}&
   \includegraphics[width=\swseven]{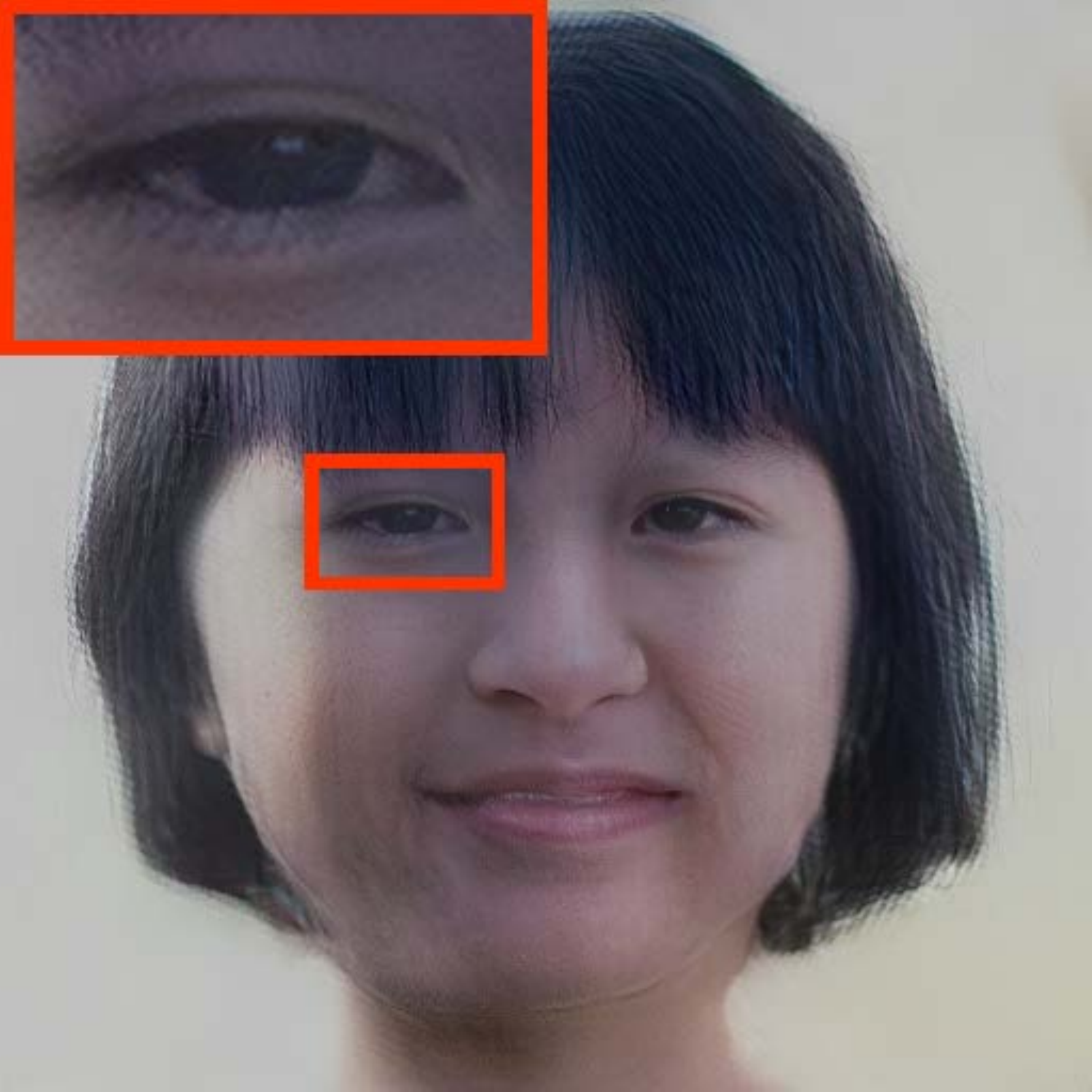}&
   \includegraphics[width=\swseven]{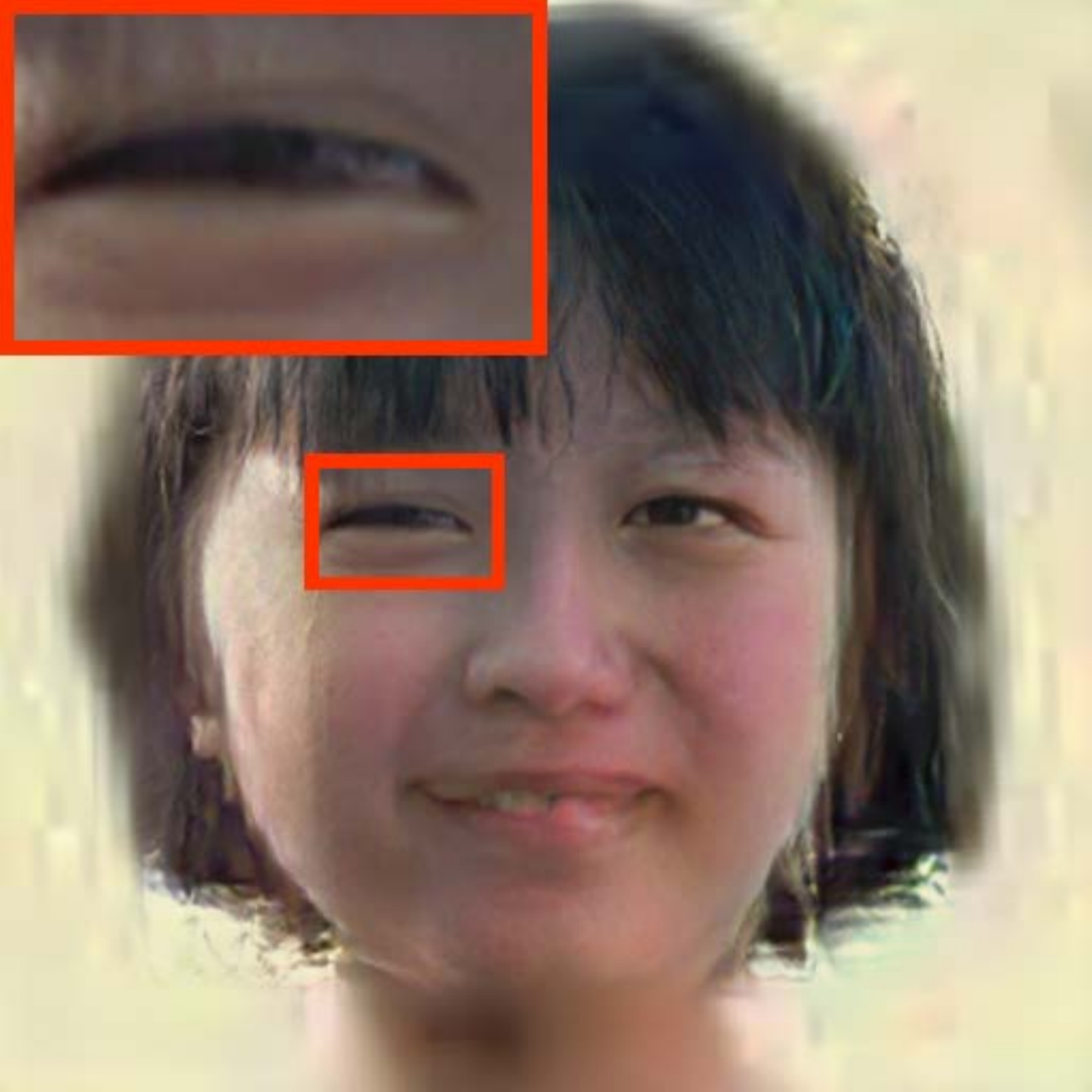}&
   \includegraphics[width=\swseven]{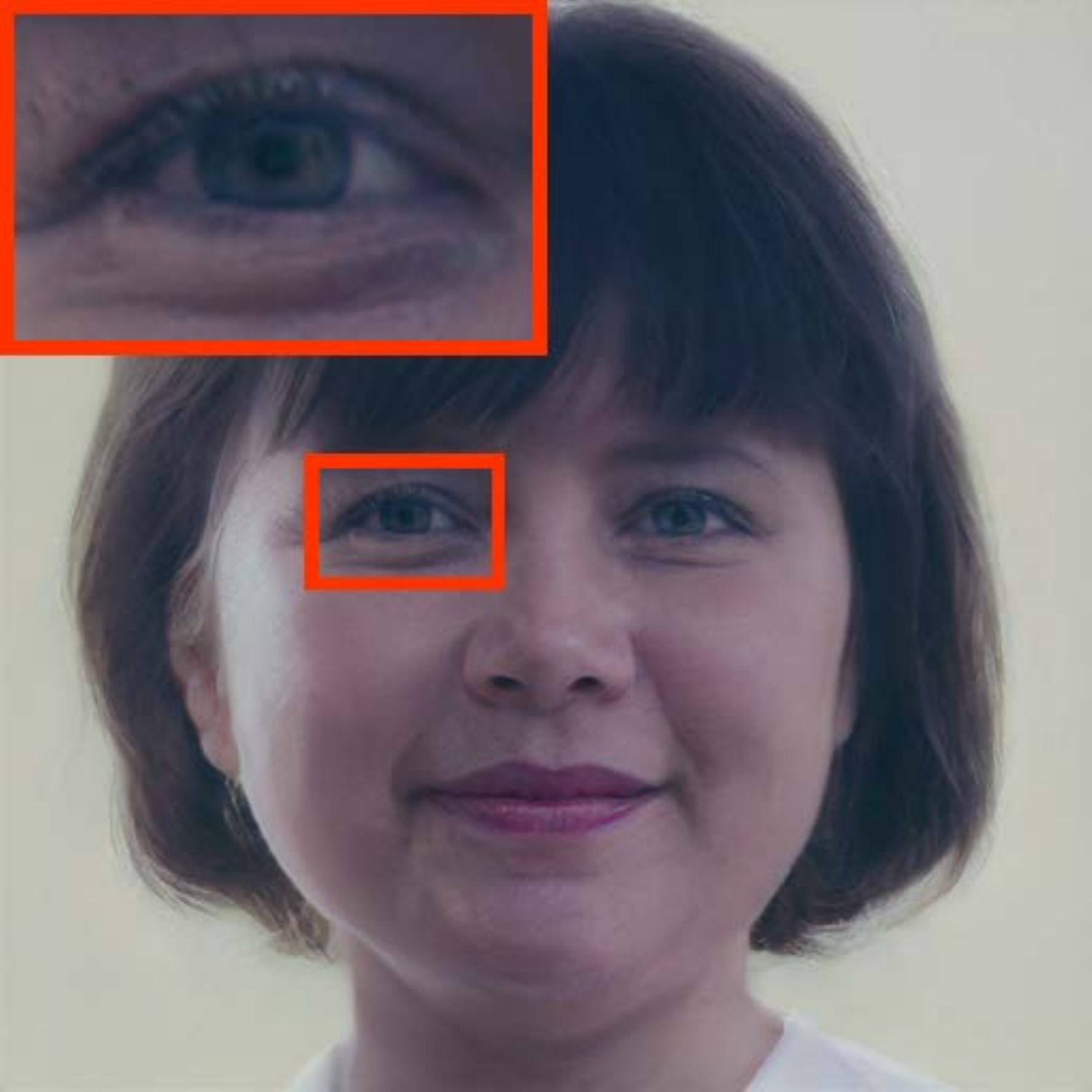}&
   \includegraphics[width=\swseven]{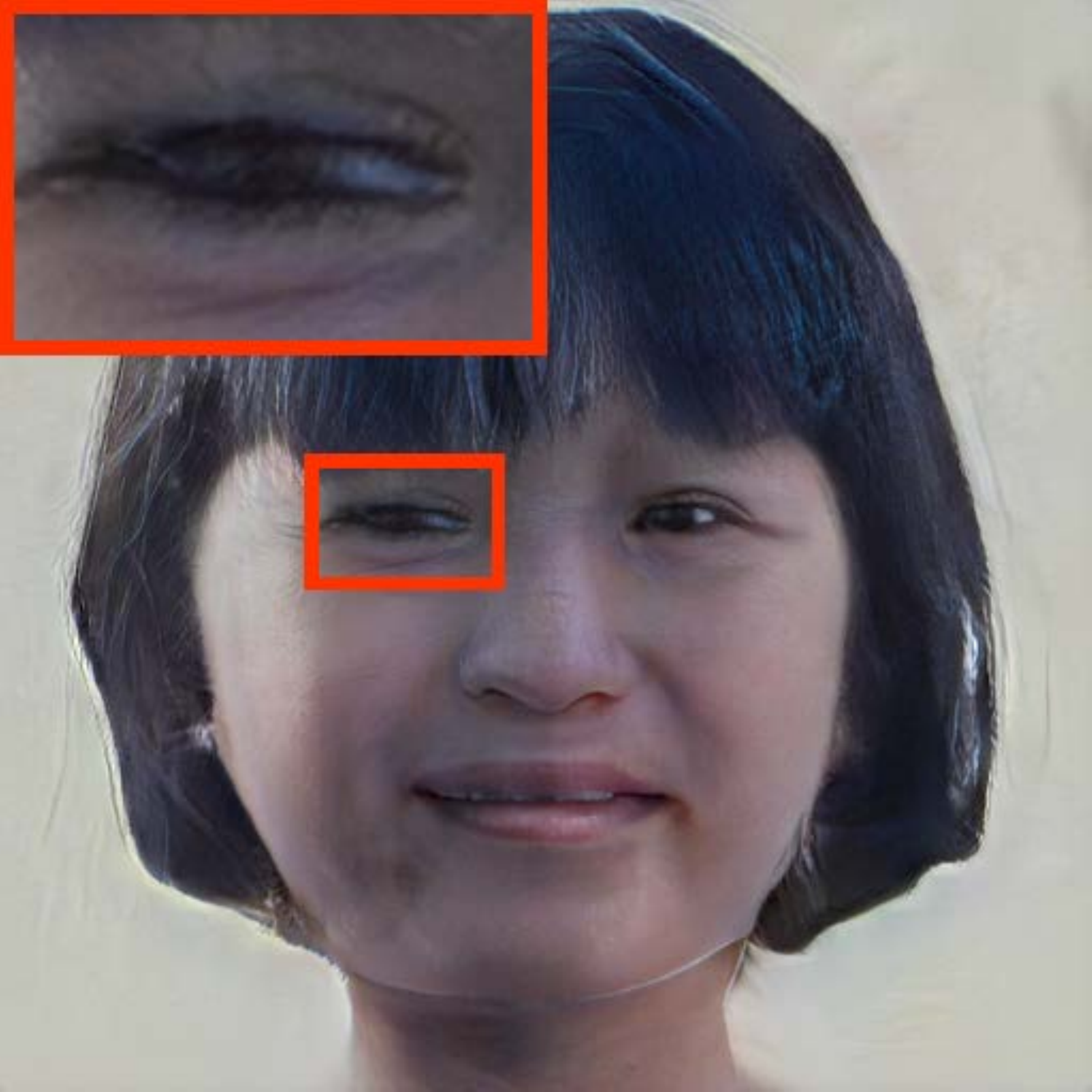}&
   \includegraphics[width=\swseven]{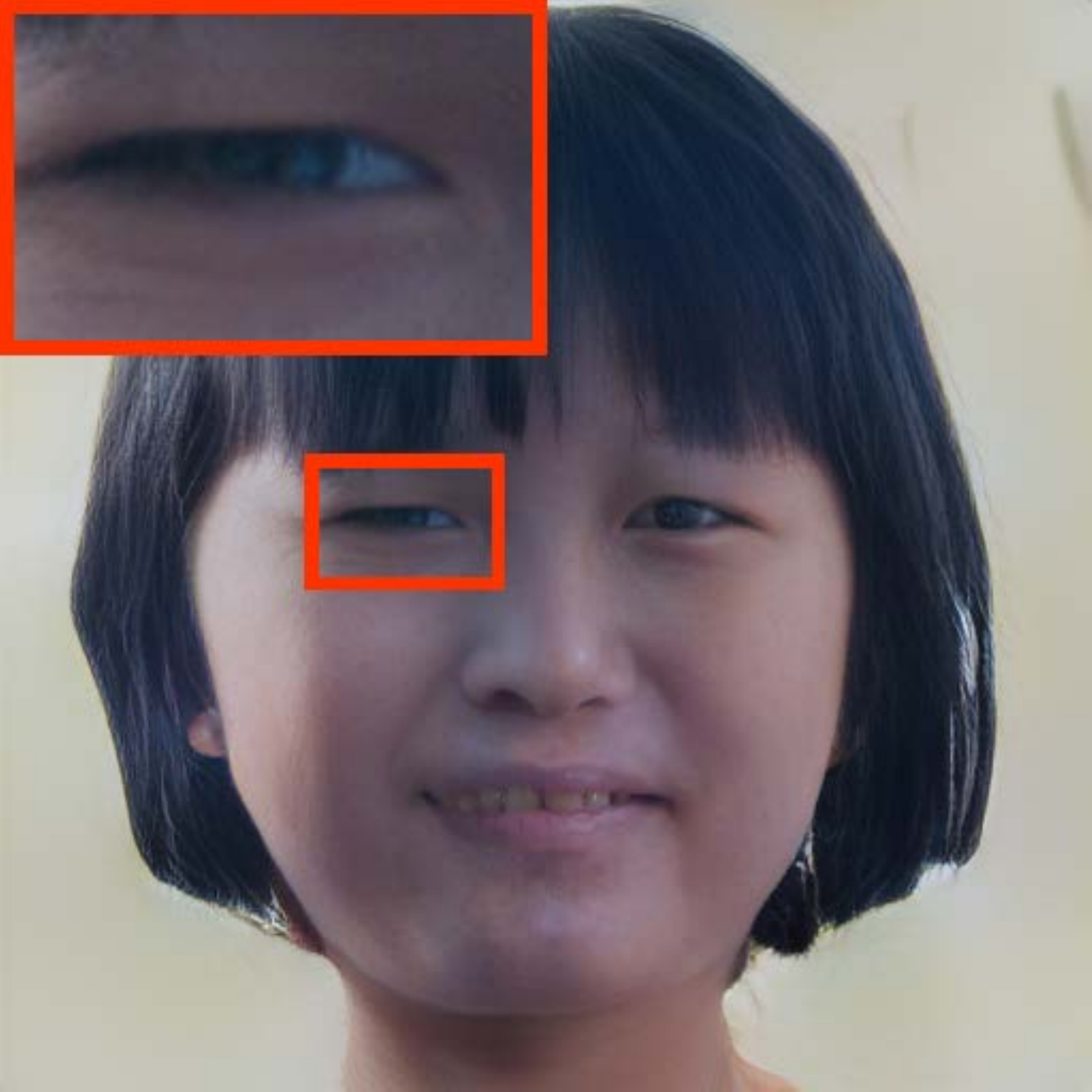}&
   \includegraphics[width=\swseven]{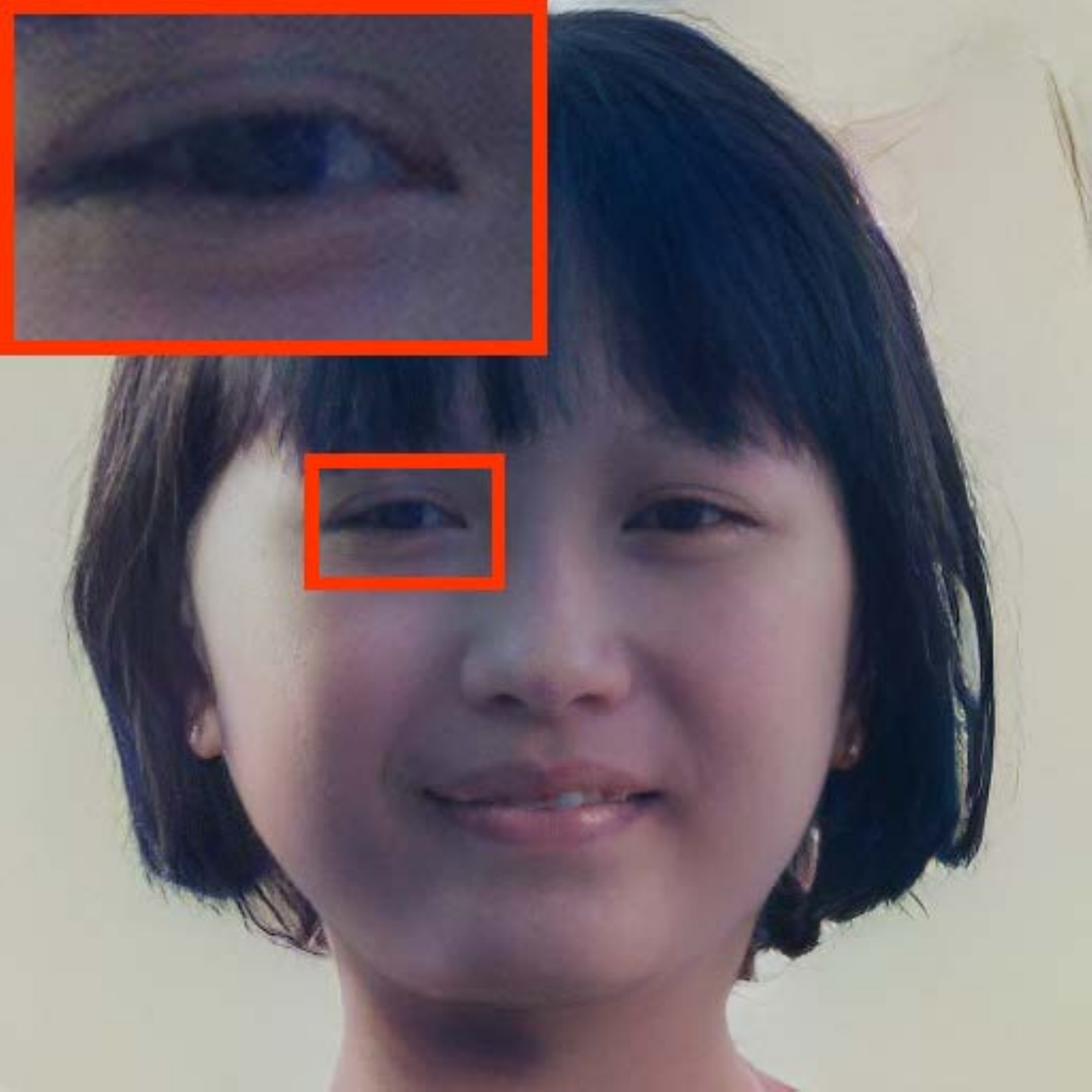} \\
   \label{fig:results}
   Input & DFDNet~\cite{li2020blind}  & Wan~\etal~\cite{wan2020bringing} & PULSE~\cite{menon2020pulse}, & PSFRGAN~\cite{chen2021progressive} & GFP-GAN~\cite{wang2021towards} & \textbf{RestoreFormer} \\
\end{tabular}
\end{center}
\vspace{-6mm}
\caption{Qualitative comparison on the three \textbf{real-world} datasets: \textbf{LFW-Test}, \textbf{CelebChild-Test}, and \textbf{WebPhoto-Test} (from top to down, respectively). \textbf{Zoom in for a better view and more results are shown in supplementary materials}.}
% The results of our RestoreFormer have a more realistic overview and contain more details in eyes, mouth, and hair. Noted that since GFP-GAN will colorize the gray face while restoring, for a fair comparison, we align its color to the color of input while visualizing. This operation will not affect the quality of restored results.
\label{fig:real}
% \end{minipage}
\vspace{-10pt}

\end{figure*}%

\noindent\textbf{Real-world Datasets.}
We also apply our RestoreFormer on three real-world datasets: LFW-Test, CelebChild-Test, and WebPhoto-Test for evaluating the generalization of the proposed method.
Their quantitative results are shown in Table.~\ref{tab:real_fid}.
Due to the reconstruction-oriented HQ Dictionary and powerful MHCA fusion block, our method performs better in all three real-world datasets based on FID.
The visual results of the three real-world datasets shown in Figure~\ref{fig:real} also show that RestoreFormer can also robustly restore faces with more details, fewer artifacts, and keep identity simultaneous relative to existing arts.
Compare to the results of Wan~\etal~\cite{wan2020bringing} and PULSE~\cite{menon2020pulse}, which are based on generative priors without considering the identity information in the degraded faces, the results from RestoreFormer look more similar to the input.
%
%Since the HQ Dictionary deployed in RestoreFormer is reconstruction-oriented, its restored results cover more facial details compared to DFDNet with recognition-oriented component dictionaries (such as the teeth in the first sample in Figure~\ref{fig:real}).
%
Besides, since the MHCA in RestoreFormer can utilize contextual information, the eyes of the third row in Figure~\ref{fig:real} look more visually pleasant than \cite{li2020blind, chen2021progressive, wang2021towards}.
%have a more consistent look compared to the results of GFP-GAN~\cite{wang2021towards} that implemented with a local combination. 

% \noindent\textbf{User Study.} 

To further evaluate the visual quality, we recruit 100 volunteers for a \textbf{user study} on 200 samples randomly selected from  LFW-Test and WebPhoto-Test (each dataset provides 100 samples).
We conduct pair-wise comparisons between RestoreFormer and  three lately state-of-the-art methods: DFDNet~\cite{li2020blind}, PSFRGAN~\cite{chen2021progressive}, and GFP-GAN~\cite{wang2021towards}.
As shown in Table.~\ref{tab:real_us}, our RestoreFormer performs better than other methods with a higher percentage.

\iffalse
\renewcommand{\tabcolsep}{.5pt}  
\begin{figure}
% \vspace{-0.35in}
% \begin{minipage}{\textwidth}
\tiny
\begin{center}
\begin{tabular}{cccc}
   \vspace{-0.5mm}
   \includegraphics[width=\swsix]{figures/ab/figure7/validation_1437_input.pdf}&
   \includegraphics[width=\swsix]{figures/ab/figure7/validation_1437_vqvae.pdf}&
   \includegraphics[width=\swsix]{figures/ab/figure7/validation_1437_vqvae_tp.pdf}&
   \includegraphics[width=\swsix]{figures/ab/figure7/validation_1437_gt.pdf}\\
   Input & prior+MHSA & RestoreFormer & GT \\
\end{tabular}
\end{center}
\vspace{-3mm}
\caption{Ablation studies on no fusion.}
\label{fig:ablation_no_fusion}
% \end{minipage}

\end{figure}%

\renewcommand{\tabcolsep}{.5pt}  
\begin{figure}
% \vspace{-0.35in}
% \begin{minipage}{\textwidth}
\tiny
\begin{center}
\begin{tabular}{cccc}
   \vspace{-0.5mm}
   \includegraphics[width=\swseven]{figures/ab/00084_01_org.pdf}&
   \includegraphics[width=\swseven]{figures/ab/00084_01_sft.pdf}&
   \includegraphics[width=\swseven]{figures/ab/00084_01_td.pdf}&
   \includegraphics[width=\swseven]{figures/ab/00084_01_tp.pdf}\\
   Input & SFT & MHCA-D & RestoreFormer \\
   \includegraphics[width=\swseven]{figures/ab/left_h3.pdf}&
   \includegraphics[width=\swseven]{figures/ab/left_h7.pdf}&
   \includegraphics[width=\swseven]{figures/ab/left_h5.pdf}&
   \includegraphics[width=\swseven]{figures/ab/left_h0.pdf} \\
   head0 & head1 & head2 & head3  \\
\end{tabular}
\end{center}
\vspace{-3mm}
\caption{Ablation studies for fusion methods.}
\label{fig:ablation_fusion}
% \end{minipage}

\end{figure}%

\fi

\renewcommand{\tabcolsep}{.5pt}  
\begin{figure*}
\vspace{-0.2cm}
% \begin{minipage}{\textwidth}
\tiny
\begin{center}
\begin{tabular}{cccccccccc}
   \vspace{-0.5mm}
   \includegraphics[width=\swten]{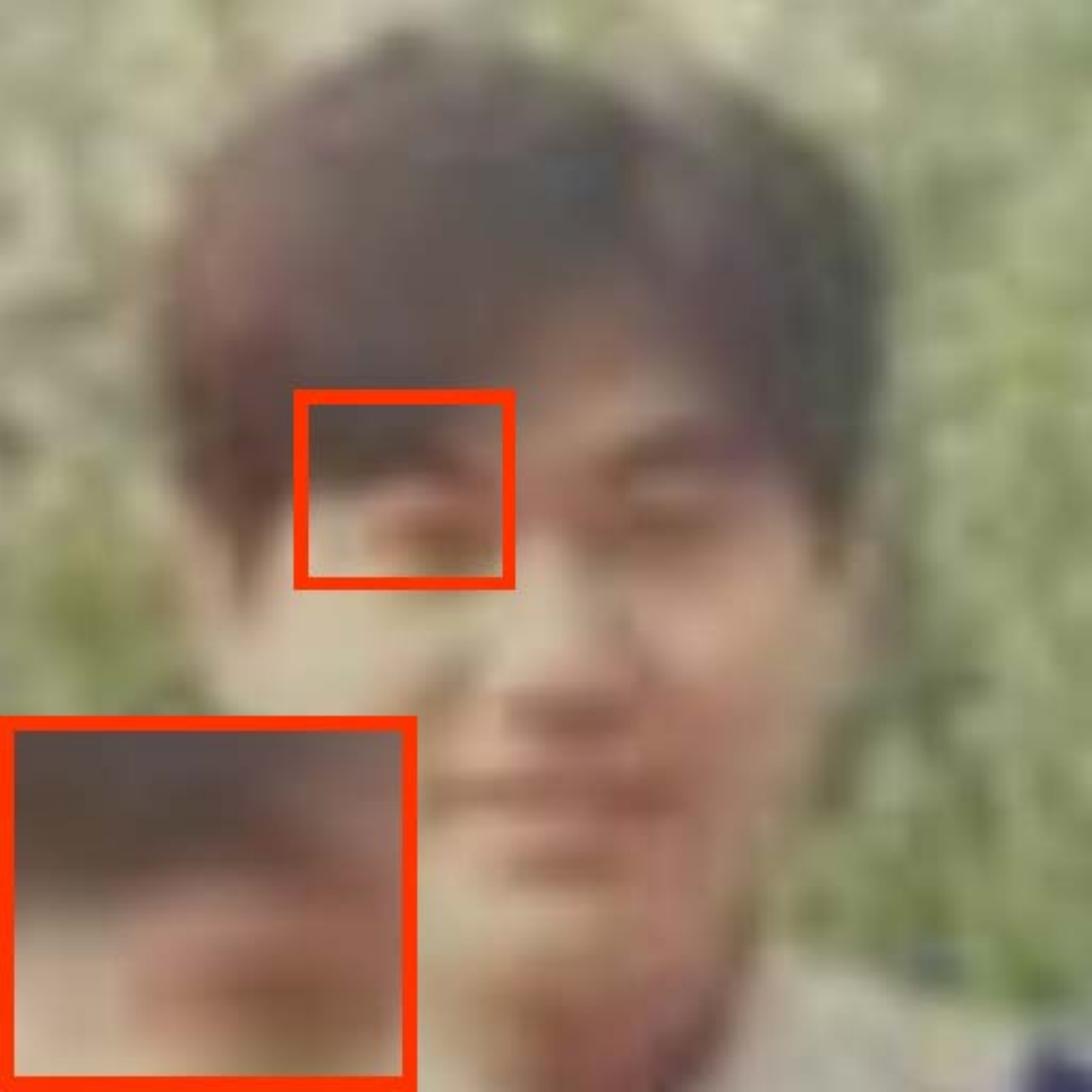}&
   \includegraphics[width=\swten]{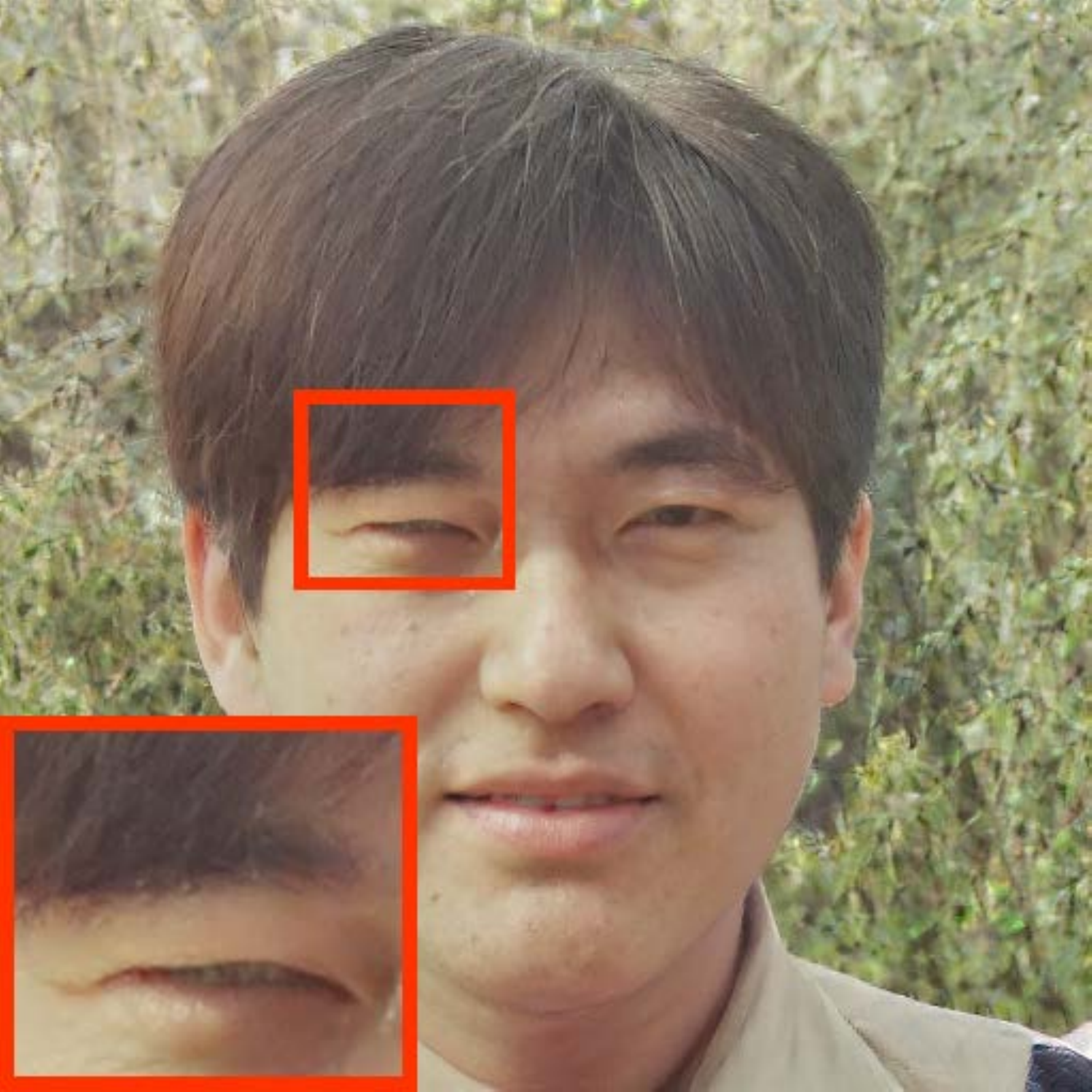}&
   \includegraphics[width=\swten]{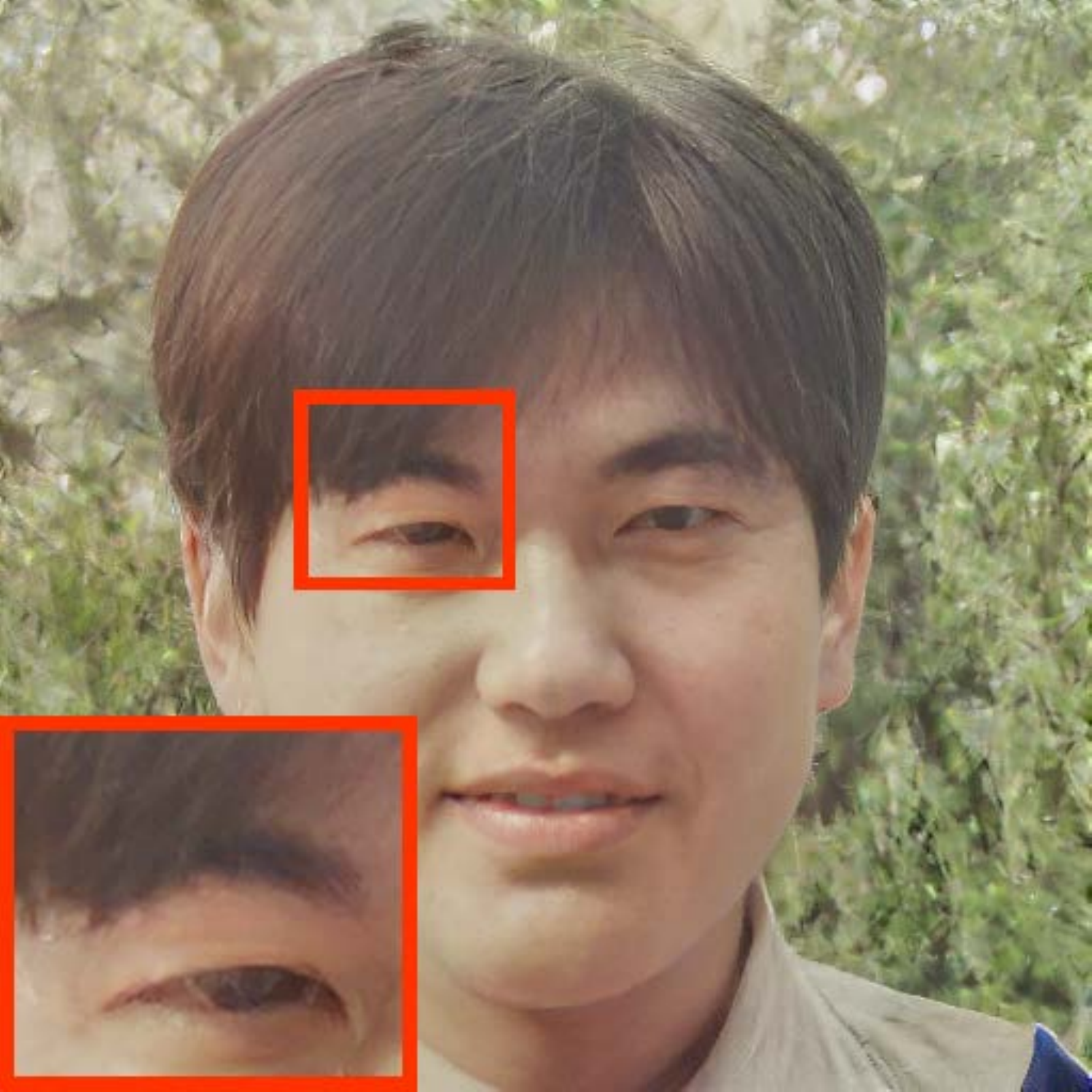}&
   \includegraphics[width=\swten]{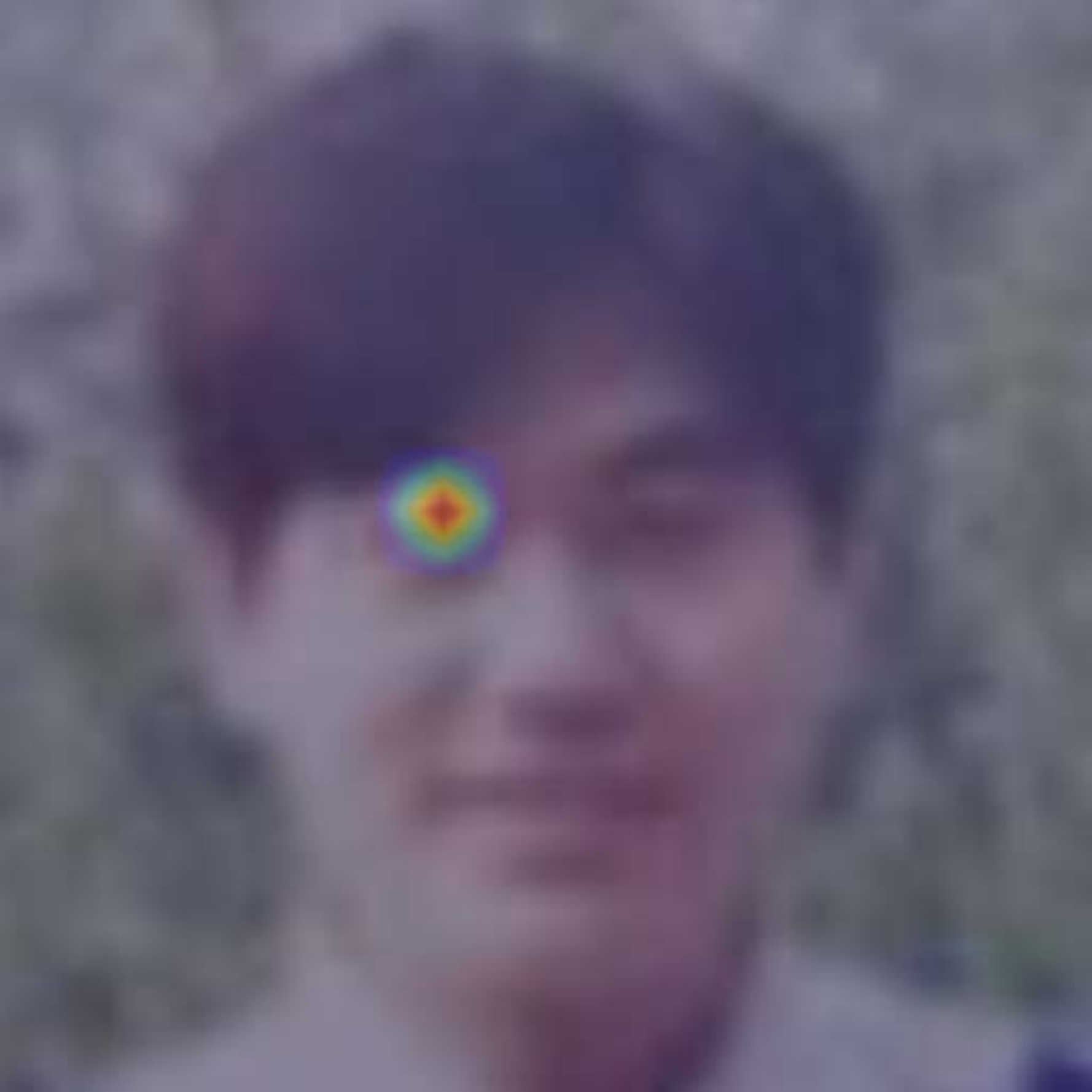}&
   \includegraphics[width=\swten]{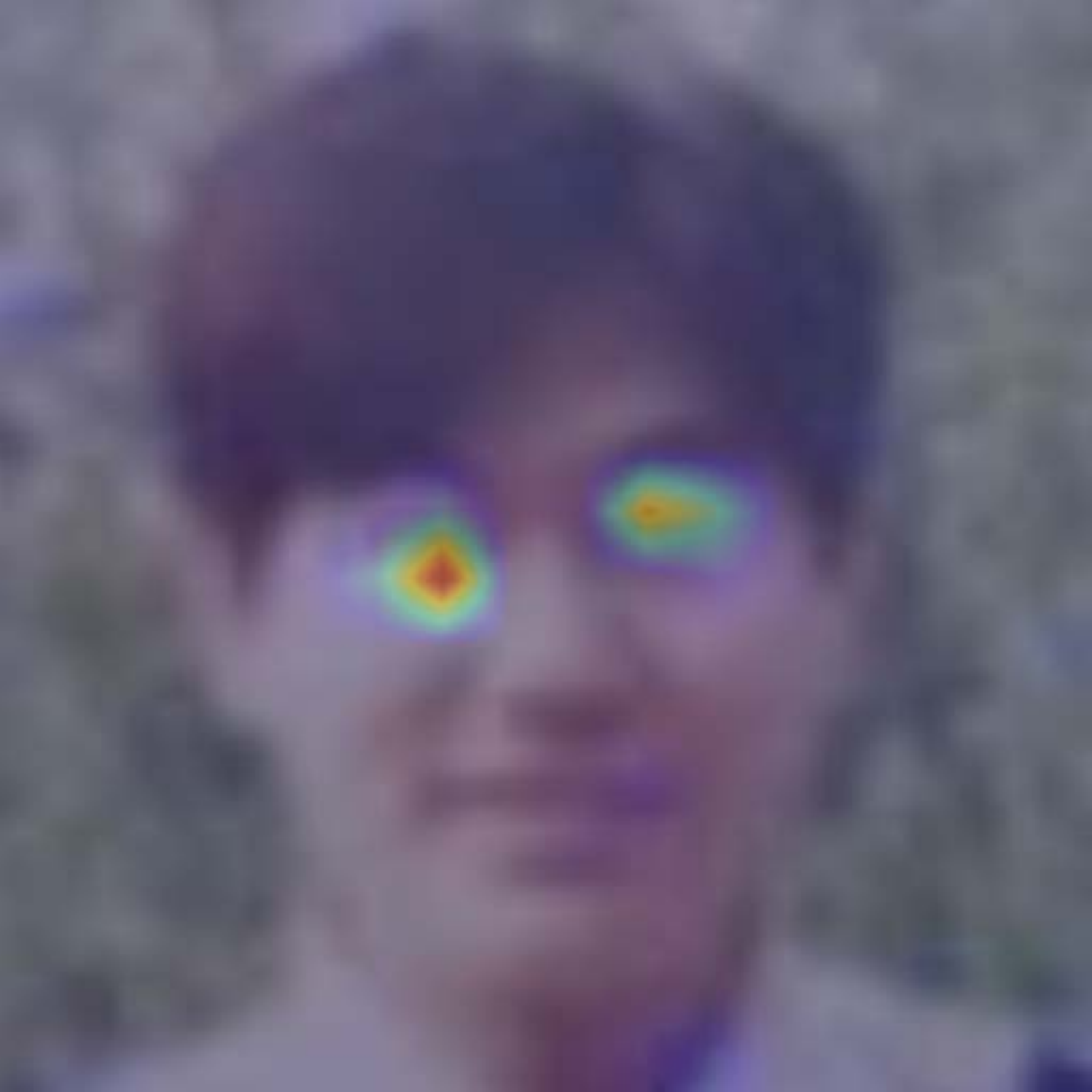} &
   \includegraphics[width=\swten]{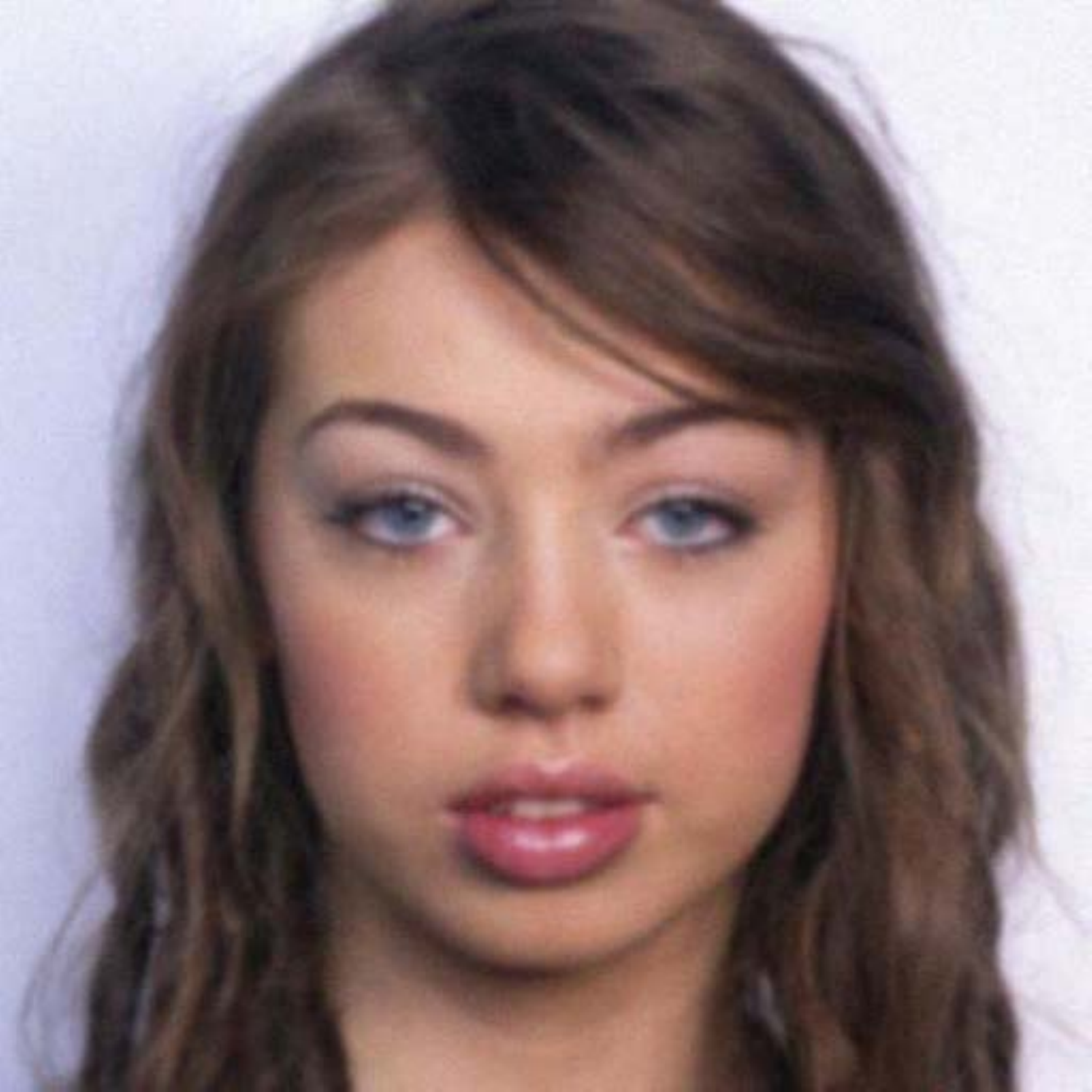}&
   \includegraphics[width=\swten]{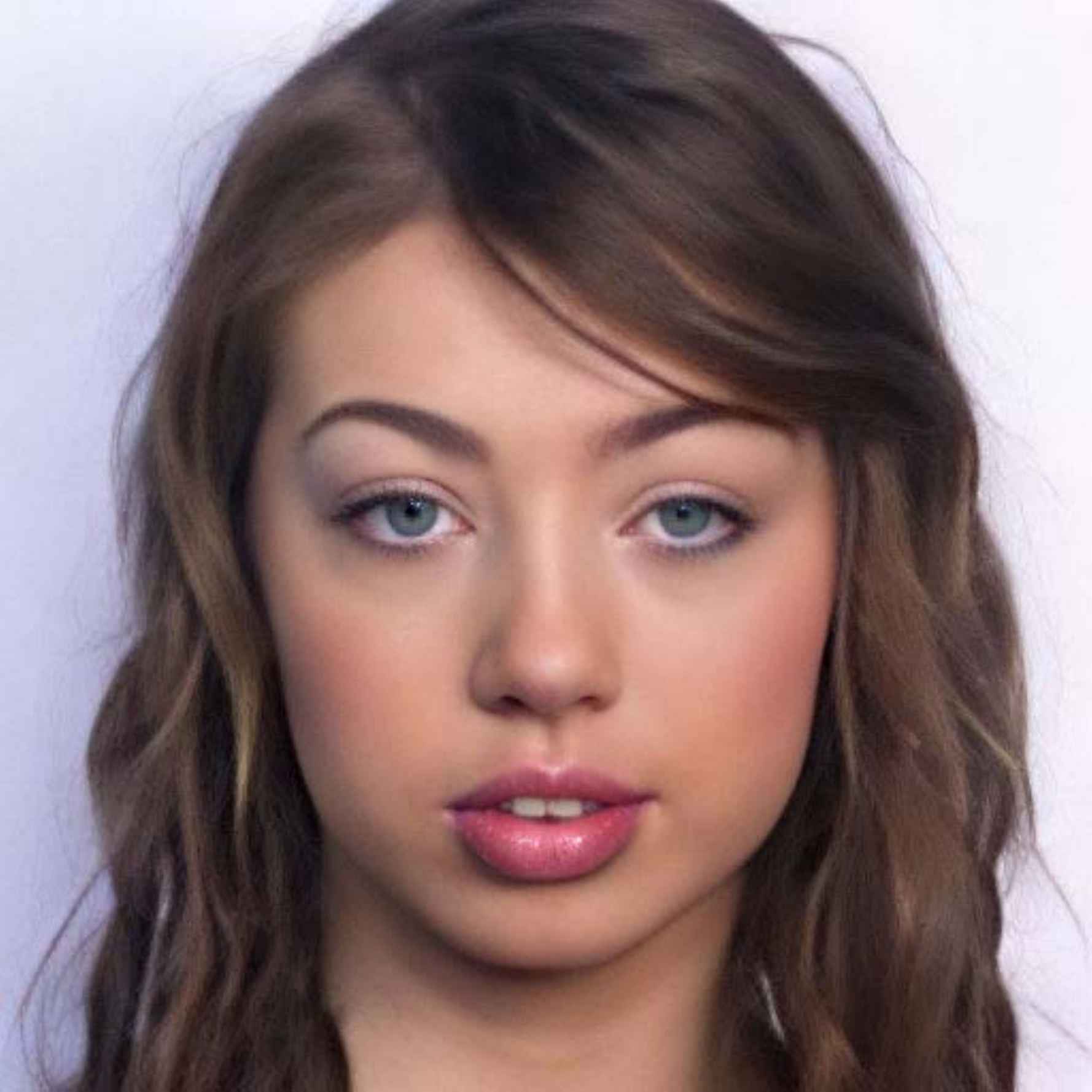}&
   \includegraphics[width=\swten]{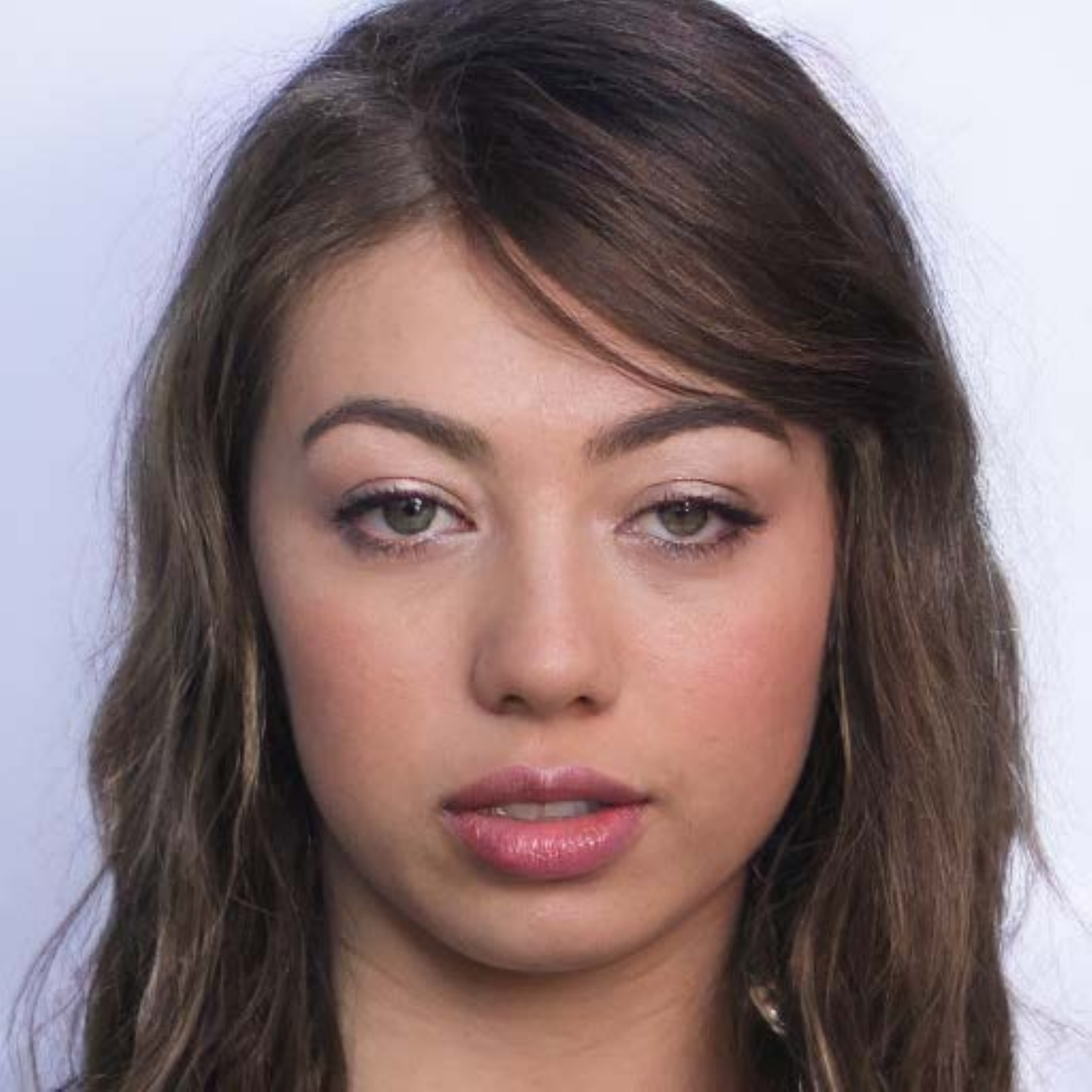}&
   \includegraphics[width=\swten]{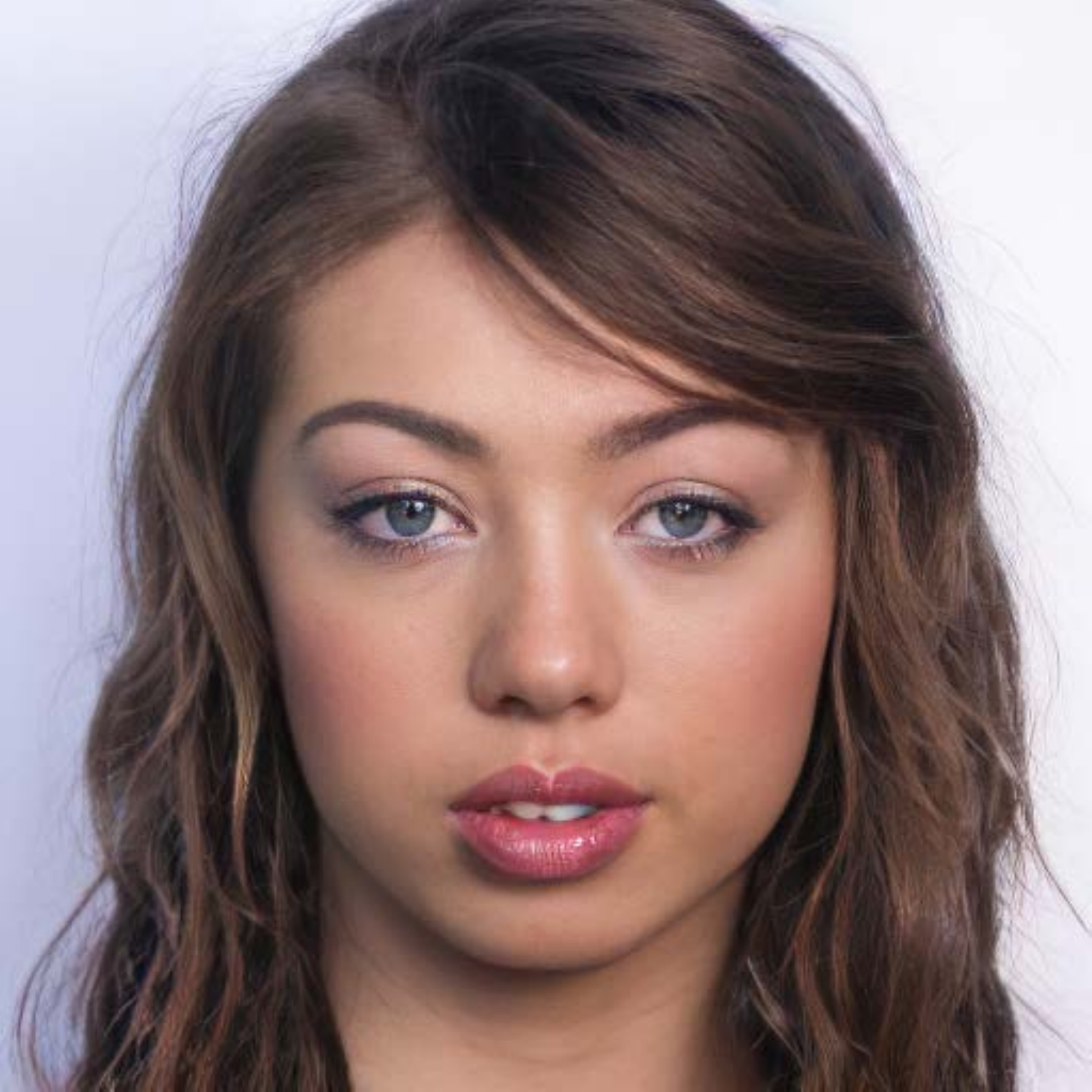}&
   \includegraphics[width=\swten]{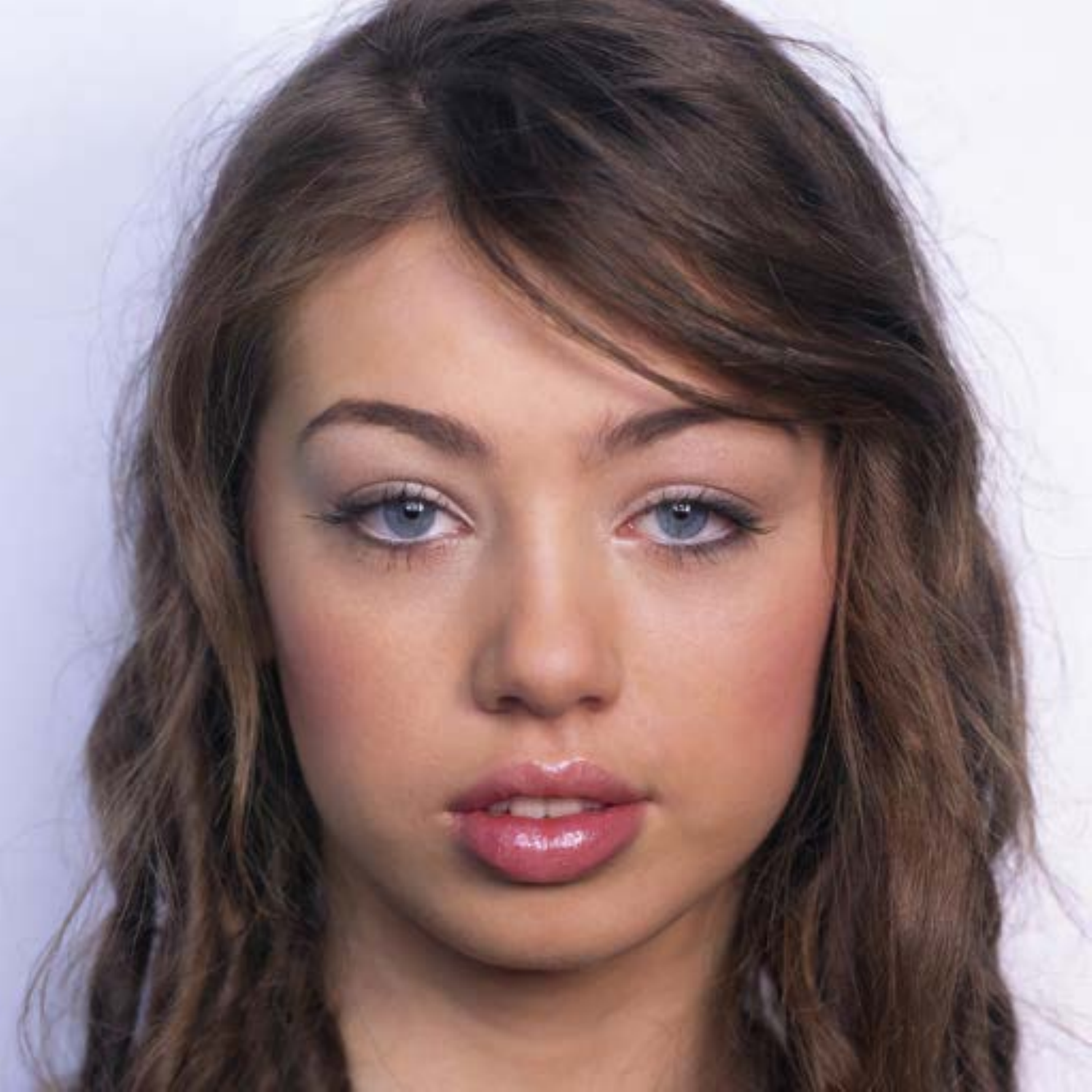}\\
   (a) input & (b) SFT  & (c) RestoreFormer & (d) attention map & (e) attention map & (f)input & (g) degraded+MHSA & (h) prior+MHSA & (i) RestoreFormer & (j) GT \\
\end{tabular}
\end{center}
\vspace{-6mm}
\caption{Ablation studies.
(c) is the restored face of (a) by RestoreFormer.
(b) replaces the MHCA with SFT to validate the effectiveness of the facial contextual information.
(d) and (e) are two attention maps for the left eye in RestoreFormer.
(f) to (j) are to validate the effectiveness of fusing the information from degraded image and prior.
(g) and (h) use self-attention, \emph{i.e}\onedot MHSA, to process either degraded information from the input or prior information from HQ Dictionary.
While our RestoreFormer can utilize these two sources of information to restore a face (i) that looks more visually pleasant than (g) and more similar to the ground truth (j) than (h).
Please see the text for more details.
}
\label{fig:ablation}
% \end{minipage}

\end{figure*}%

\subsection{Ablation Study}

%% No fusion (self-attention and no self-attention?), concat, sft, cross-attention

%% shot-cup with degraded / prior.

%To validate the effectiveness of RestoreFormer, we evaluate the variants of it based on two dimensions, \emph{i.e}\onedot fusion sources and fusion methods.
%To validate the effectiveness of RestoreFormer, we evaluate the variants of it based on different attention and fusion methods.
%
%We also fulfill an experiment to demonstrate the effectiveness of the proposed reconstruction-oriented dictionary relative to the recognition-oriented one.

According to the analysis above, RestoreFormer has several merits.
First of all, the spatial attention mechanism is used to utilize the abundant contextual information in face images for restoration.
In addition, the proposed method can properly utilize the identity information from the degraded face and high-quality facial details from priors.
At last, Dictionary used in RestoreFormer is reconstruction-oriented other than the recognition-oriented one used in \cite{li2020blind}.
The above factors will be discussed in the following subsections and these networks\footnote{Please see the supplemental materials for the detailed structures for these networks.} are trained by exactly the same settings as to RestoreFormer.

\noindent\textbf{Spatial attention.}
In this subsection, variants of RestoreFormer without and with attention mechanism are compared.
Both exp1 and exp2 in Table~\ref{tab:ablation} only use degraded images in the network.
By using self-attention and exploring contextual information, exp2 with MHSA has lower FID and IDD than exp1 which directly uses the features extracted from the degraded image.
This conclusion is also valid when the networks consider information from both degraded image and dictionary prior in exp4 and RestoreFormer in Table~\ref{tab:ablation}.
In exp4, MHCA is replaced by SFT in RestoreFormer to locally fuse the information.
Without considering the global contextual information, the left eye seems strange in Figure~\ref{fig:ablation}~(b).
As shown in Figure~\ref{fig:ablation}~(d) and (e), the multi-head attention maps of the left eye region have more weights for both two eyes in the RestoreFormer with MHCA.
This means RestoreFormer with MHCA utilizes the information from both eyes to restore the left one and generates a more visually pleasant result in Figure~\ref{fig:ablation}~(c).

\noindent\textbf{Degraded information and Prior.}
This subsection analyses the effect of degraded information from input images and priors from the HQ Dictionary.
Similar to existing ViT methods which use self-attention(MHSA), all the query, key and value are either from features of degraded images (exp2) or priors (exp3) in Table~\ref{tab:ablation}.
It shows that exp2 has a better average IDD score for keeping the identity of the faces and exp3 has a better average FID score for the realness of the results.
By utilizing cross-attention (MHCA) in RestoreFormer to fuse these two sources of information, RestoreFormer is better than exp2 and exp3 for both IDD and FID.
As to the visual result, Figure~\ref{fig:ablation}~(g) shows that `degraded+MHSA' (exp2) can restore a face that looks more like the ground truth.
However, its result contains fewer details relative to RestoreFormer in Figure~\ref{fig:ablation}~(i) which makes the face visually less pleasant.
%also contains some unreal details, \emph{e.g}\onedot artifacts in the face and repetitive patterns in hair, which make the face visually unpleasant.
%
Even though the details in `prior+MHSA' (exp3) look more natural in Figure~\ref{fig:ablation}~(h), the generated face looks like a different person relative to the ground truth, especially for the mouth.
By fusing the information from degraded image and prior, RestoreFormer can restore face with more real details as well as maintaining identity shown in Figure~\ref{fig:ablation}~(i).
According to Figure~\ref{fig:framework}~(b) and Eq.~\ref{eq:shortcut}, there is a skip connection between the attended feature $\bm{Z}_{mh}$ and prior $\bm{Z}_p$ in the RestoreFormer.
This is because we experimentally find it performs better than adding $\bm{Z}_{mh}$ with the feature from degraded input $\bm{Z}_{d}$ denoted as exp5 in Table~\ref{tab:ablation}.

\begin{table}
    % \small
%   \tiny
%   \scriptsize
\vspace{-0.6cm}
  \centering
  \resizebox{\linewidth}{!}{
  \begin{tabular}{c|c|c|c|c|c|c|c|c|c} % {@{}lc@{}}
    \toprule
    &\multicolumn{2}{c|}{sources} & \multicolumn{5}{c|}{methods} & \multicolumn{2}{c}{metrics} \\
    \midrule
    ~~No. of exp.~~ & ~~degraded~~ & ~~prior~~ & ~~none~~ & ~~MHSA~~ & ~~SFT~~  & ~~MHCA-D~~ & ~~MHCA-P~~ & ~~FID$\downarrow$~~ & ~~IDD$\downarrow$~~ \\
    \midrule
    exp1 & \checkmark & & \checkmark &&&&&50.68&0.6401 \\
    % \midrule
    exp2 &\checkmark&&&\checkmark&&&&47.39&0.6284 \\
    % \midrule
    exp3 &&\checkmark&&\checkmark&&&&45.83&0.7662 \\
    % \midrule
    exp4 &\checkmark&\checkmark&&&\checkmark&&&41.47&0.6702 \\
    % \midrule
    exp5 &\checkmark&\checkmark&&&&\checkmark&&42.00&0.5938 \\
    % \midrule
    Ours&\checkmark&\checkmark&&&&&\checkmark& \textbf{41.45} & \textbf{0.5650} \\
    \bottomrule
  \end{tabular}
  
  }
  \vspace{-0.3cm}
  \caption{Quantitative results of ablation studies on CelebA-Test.
  `degraded' and `prior' mean fusion information from degraded input and HQ Dictionary, respectively.
  `none' and `MHSA' mean the network uses either `degraded' or `prior' information without or with using self-attention mechanism, respectively.
  `SFT', `MHCA-D' and `MHCA-P' use both `degraded' and `prior' information.
  `SFT' uses SFT to fuse the information while `MHCA-D' and `MHCA-P' use multi-head cross attention.
  The difference between `MHCA-D' and `MHCA-P' is `MHCA-D' fuses $\bm{Z}_{mh}$ with $\bm{Z}_d$ but `MHCA-P' fuses $\bm{Z}_{mh}$ with $\bm{Z}_p$.
  The proposed RestoreFormer integrated with `MHCA-P' performs the best relative to other variants.
  }
  \label{tab:ablation}
\end{table}

\noindent\textbf{Reconstruction-oriented v.s. Recognition-oriented.}
To evaluate the effectiveness of the proposed reconstruction-oriented HQ Dictionary, we replace the encoder $\mathbf{E}_d$ and $\mathbf{E}_h$ with a well-trained VGG~\cite{simonyan2014very} which is used in \cite{li2020blind} for face restoration and get a recognition-oriented HQ Dictionary in Restoreformer.
When training this Restoreformer, the encoder is initialized by VGG and fixed similar to \cite{li2020blind}.
The experimental results in CelebA-Test show that the average FID and IDD of this Restoreformer variant are 61.43 and 1.1401 which are worse than the proposed one according to Table~\ref{tab:ablation}.
And this demonstrates the effectiveness of the reconstruction-oriented dictionary.

% \subsection{Limitaions}

\section{Conclusion}
This paper aims for blind face restoration with a RestoreFormer, which explores fully-spacial attentions to model contextual information with a multi-head cross-attention layer to learn spatial interaction between corrupted queries and high-quality key-value pairs.
Especially, the high-quality key-value pairs are sampled from a reconstruction-oriented dictionary, whose elements are rich in high-quality facial features specifically aimed for face reconstruction.
Extensive comparisons with state-of-the-art methods on several datasets demonstrate the superior capability of the proposed RestoreFormer.

%%%%%%%%% REFERENCES
{\small
\bibliographystyle{ieee_fullname}
\bibliography{RestoreFormer}

\begin{thebibliography}{10}\itemsep=-1pt

\bibitem{blau20182018}
Yochai Blau, Roey Mechrez, Radu Timofte, Tomer Michaeli, and Lihi Zelnik-Manor.
\newblock The 2018 pirm challenge on perceptual image super-resolution.
\newblock In {\em ECCVW}, 2018.

\bibitem{brown2020language}
Tom~B Brown, Benjamin Mann, Nick Ryder, Melanie Subbiah, Jared Kaplan, Prafulla
  Dhariwal, Arvind Neelakantan, Pranav Shyam, Girish Sastry, Amanda Askell,
  et~al.
\newblock Language models are few-shot learners.
\newblock {\em arXiv preprint arXiv:2005.14165}, 2020.

\bibitem{cao2017attention}
Qingxing Cao, Liang Lin, Yukai Shi, Xiaodan Liang, and Guanbin Li.
\newblock Attention-aware face hallucination via deep reinforcement learning.
\newblock In {\em CVPR}, 2017.

\bibitem{carion2020end}
Nicolas Carion, Francisco Massa, Gabriel Synnaeve, Nicolas Usunier, Alexander
  Kirillov, and Sergey Zagoruyko.
\newblock End-to-end object detection with transformers.
\newblock In {\em ECCV}, 2020.

\bibitem{chen2021progressive}
Chaofeng Chen, Xiaoming Li, Lingbo Yang, Xianhui Lin, Lei Zhang, and Kwan-Yee~K
  Wong.
\newblock Progressive semantic-aware style transformation for blind face
  restoration.
\newblock In {\em CVPR}, 2021.

\bibitem{chen2021pre}
Hanting Chen, Yunhe Wang, Tianyu Guo, Chang Xu, Yiping Deng, Zhenhua Liu, Siwei
  Ma, Chunjing Xu, Chao Xu, and Wen Gao.
\newblock Pre-trained image processing transformer.
\newblock In {\em CVPR}, 2021.

\bibitem{chen2018fsrnet}
Yu Chen, Ying Tai, Xiaoming Liu, Chunhua Shen, and Jian Yang.
\newblock Fsrnet: End-to-end learning face super-resolution with facial priors.
\newblock In {\em CVPR}, 2018.

\bibitem{deng2019arcface}
Jiankang Deng, Jia Guo, Niannan Xue, and Stefanos Zafeiriou.
\newblock Arcface: Additive angular margin loss for deep face recognition.
\newblock In {\em CVPR}, 2019.

\bibitem{devlin2018bert}
Jacob Devlin, Ming-Wei Chang, Kenton Lee, and Kristina Toutanova.
\newblock Bert: Pre-training of deep bidirectional transformers for language
  understanding.
\newblock {\em arXiv preprint arXiv:1810.04805}, 2018.

\bibitem{dogan2019exemplar}
Berk Dogan, Shuhang Gu, and Radu Timofte.
\newblock Exemplar guided face image super-resolution without facial landmarks.
\newblock In {\em CVPRW}, 2019.

\bibitem{dosovitskiy2020image}
Alexey Dosovitskiy, Lucas Beyer, Alexander Kolesnikov, Dirk Weissenborn,
  Xiaohua Zhai, Thomas Unterthiner, Mostafa Dehghani, Matthias Minderer, Georg
  Heigold, Sylvain Gelly, et~al.
\newblock An image is worth 16x16 words: Transformers for image recognition at
  scale.
\newblock {\em arXiv preprint arXiv:2010.11929}, 2020.

\bibitem{esser2021taming}
Patrick Esser, Robin Rombach, and Bjorn Ommer.
\newblock Taming transformers for high-resolution image synthesis.
\newblock In {\em CVPR}, 2021.

\bibitem{gatys2016image}
Leon~A Gatys, Alexander~S Ecker, and Matthias Bethge.
\newblock Image style transfer using convolutional neural networks.
\newblock In {\em CVPR}, pages 2414--2423, 2016.

\bibitem{gu2020image}
Jinjin Gu, Yujun Shen, and Bolei Zhou.
\newblock Image processing using multi-code gan prior.
\newblock In {\em CVPR}, 2020.

\bibitem{he2017mask}
Kaiming He, Georgia Gkioxari, Piotr Doll{\'a}r, and Ross Girshick.
\newblock Mask r-cnn.
\newblock In {\em ICCV}, 2017.

\bibitem{heusel2017gans}
Martin Heusel, Hubert Ramsauer, Thomas Unterthiner, Bernhard Nessler, and Sepp
  Hochreiter.
\newblock Gans trained by a two time-scale update rule converge to a local nash
  equilibrium.
\newblock {\em NIPS}, 2017.

\bibitem{huang2008labeled}
Gary~B Huang, Marwan Mattar, Tamara Berg, and Eric Learned-Miller.
\newblock Labeled faces in the wild: A database forstudying face recognition in
  unconstrained environments.
\newblock In {\em Workshop on faces in'Real-Life'Images: detection, alignment,
  and recognition}, 2008.

\bibitem{huang2017wavelet}
Huaibo Huang, Ran He, Zhenan Sun, and Tieniu Tan.
\newblock Wavelet-srnet: A wavelet-based cnn for multi-scale face super
  resolution.
\newblock In {\em ICCV}, 2017.

\bibitem{johnson2016perceptual}
Justin Johnson, Alexandre Alahi, and Li Fei-Fei.
\newblock Perceptual losses for real-time style transfer and super-resolution.
\newblock In {\em ECCV}, 2016.

\bibitem{karras2019style}
Tero Karras, Samuli Laine, and Timo Aila.
\newblock A style-based generator architecture for generative adversarial
  networks.
\newblock In {\em CVPR}, 2019.

\bibitem{kim2019progressive}
Deokyun Kim, Minseon Kim, Gihyun Kwon, and Dae-Shik Kim.
\newblock Progressive face super-resolution via attention to facial landmark.
\newblock {\em arXiv preprint arXiv:1908.08239}, 2019.

\bibitem{kingma2014adam}
Diederik~P Kingma and Jimmy Ba.
\newblock Adam: A method for stochastic optimization.
\newblock {\em arXiv}, 2014.

\bibitem{ledig2017photo}
Christian Ledig, Lucas Theis, Ferenc Husz{\'a}r, Jose Caballero, Andrew
  Cunningham, Alejandro Acosta, Andrew Aitken, Alykhan Tejani, Johannes Totz,
  Zehan Wang, et~al.
\newblock Photo-realistic single image super-resolution using a generative
  adversarial network.
\newblock In {\em CVPR}, 2017.

\bibitem{li2020blind}
Xiaoming Li, Chaofeng Chen, Shangchen Zhou, Xianhui Lin, Wangmeng Zuo, and Lei
  Zhang.
\newblock Blind face restoration via deep multi-scale component dictionaries.
\newblock In {\em ECCV}, 2020.

\bibitem{li2019recovering}
Xiu Li, Guichun Duan, Zhouxia Wang, Jimmy Ren, Yongbing Zhang, Jiawei Zhang,
  and Kaixiang Song.
\newblock Recovering extremely degraded faces by joint super-resolution and
  facial composite.
\newblock In {\em ICTAI}, 2019.

\bibitem{li2020enhanced}
Xiaoming Li, Wenyu Li, Dongwei Ren, Hongzhi Zhang, Meng Wang, and Wangmeng Zuo.
\newblock Enhanced blind face restoration with multi-exemplar images and
  adaptive spatial feature fusion.
\newblock In {\em CVPR}, 2020.

\bibitem{li2018learning}
Xiaoming Li, Ming Liu, Yuting Ye, Wangmeng Zuo, Liang Lin, and Ruigang Yang.
\newblock Learning warped guidance for blind face restoration.
\newblock In {\em ECCV}, 2018.

\bibitem{liu2015deep}
Ziwei Liu, Ping Luo, Xiaogang Wang, and Xiaoou Tang.
\newblock Deep learning face attributes in the wild.
\newblock In {\em ICCV}, 2015.

\bibitem{menon2020pulse}
Sachit Menon, Alexandru Damian, Shijia Hu, Nikhil Ravi, and Cynthia Rudin.
\newblock Pulse: Self-supervised photo upsampling via latent space exploration
  of generative models.
\newblock In {\em CVPR}, 2020.

\bibitem{oord2017neural}
Aaron van~den Oord, Oriol Vinyals, and Koray Kavukcuoglu.
\newblock Neural discrete representation learning.
\newblock {\em arXiv preprint arXiv:1711.00937}, 2017.

\bibitem{parmar2018image}
Niki Parmar, Ashish Vaswani, Jakob Uszkoreit, Lukasz Kaiser, Noam Shazeer,
  Alexander Ku, and Dustin Tran.
\newblock Image transformer.
\newblock In {\em ICML}, 2018.

\bibitem{shen2018deep}
Ziyi Shen, Wei-Sheng Lai, Tingfa Xu, Jan Kautz, and Ming-Hsuan Yang.
\newblock Deep semantic face deblurring.
\newblock In {\em CVPR}, 2018.

\bibitem{simonyan2014very}
Karen Simonyan and Andrew Zisserman.
\newblock Very deep convolutional networks for large-scale image recognition.
\newblock {\em arXiv preprint arXiv:1409.1556}, 2014.

\bibitem{vaswani2017attention}
Ashish Vaswani, Noam Shazeer, Niki Parmar, Jakob Uszkoreit, Llion Jones,
  Aidan~N Gomez, {\L}ukasz Kaiser, and Illia Polosukhin.
\newblock Attention is all you need.
\newblock In {\em NIPS}, 2017.

\bibitem{wan2020bringing}
Ziyu Wan, Bo Zhang, Dongdong Chen, Pan Zhang, Dong Chen, Jing Liao, and Fang
  Wen.
\newblock Bringing old photos back to life.
\newblock In {\em CVPR}, 2020.

\bibitem{wang2021max}
Huiyu Wang, Yukun Zhu, Hartwig Adam, Alan Yuille, and Liang-Chieh Chen.
\newblock Max-deeplab: End-to-end panoptic segmentation with mask transformers.
\newblock In {\em CVPR}, 2021.

\bibitem{wang2021towards}
Xintao Wang, Yu Li, Honglun Zhang, and Ying Shan.
\newblock Towards real-world blind face restoration with generative facial
  prior.
\newblock In {\em CVPR}, 2021.

\bibitem{wang2018recovering}
Xintao Wang, Ke Yu, Chao Dong, and Chen~Change Loy.
\newblock Recovering realistic texture in image super-resolution by deep
  spatial feature transform.
\newblock In {\em CVPR}, 2018.

\bibitem{xu2017learning}
Xiangyu Xu, Deqing Sun, Jinshan Pan, Yujin Zhang, Hanspeter Pfister, and
  Ming-Hsuan Yang.
\newblock Learning to super-resolve blurry face and text images.
\newblock In {\em ICCV}, 2017.

\bibitem{yang2020learning}
Fuzhi Yang, Huan Yang, Jianlong Fu, Hongtao Lu, and Baining Guo.
\newblock Learning texture transformer network for image super-resolution.
\newblock In {\em CVPR}, 2020.

\bibitem{yu2018face}
Xin Yu, Basura Fernando, Bernard Ghanem, Fatih Porikli, and Richard Hartley.
\newblock Face super-resolution guided by facial component heatmaps.
\newblock In {\em ECCV}, 2018.

\bibitem{yu2018super}
Xin Yu, Basura Fernando, Richard Hartley, and Fatih Porikli.
\newblock Super-resolving very low-resolution face images with supplementary
  attributes.
\newblock In {\em CVPR}, 2018.

\bibitem{zhang2018unreasonable}
Richard Zhang, Phillip Isola, Alexei~A Efros, Eli Shechtman, and Oliver Wang.
\newblock The unreasonable effectiveness of deep features as a perceptual
  metric.
\newblock In {\em CVPR}, 2018.

\bibitem{zhao2021improved}
Long Zhao, Zizhao Zhang, Ting Chen, Dimitris~N Metaxas, and Han Zhang.
\newblock Improved transformer for high-resolution gans.
\newblock {\em arXiv preprint arXiv:2106.07631}, 2021.

\bibitem{zhusketch}
Mingrui Zhu, Changcheng Liang, Nannan Wang, Xiaoyu Wang, Zhifeng Li, and Xinbo
  Gao.
\newblock A sketch-transformer network for face photo-sketch synthesis.
\newblock {\em IJCAI}, 2021.

\bibitem{zhu2016deep}
Shizhan Zhu, Sifei Liu, Chen~Change Loy, and Xiaoou Tang.
\newblock Deep cascaded bi-network for face hallucination.
\newblock In {\em ECCV}, 2016.

\bibitem{zhu2020deformable}
Xizhou Zhu, Weijie Su, Lewei Lu, Bin Li, Xiaogang Wang, and Jifeng Dai.
\newblock Deformable detr: Deformable transformers for end-to-end object
  detection.
\newblock {\em arXiv preprint arXiv:2010.04159}, 2020.

\end{thebibliography}
}

\end{document}